%% file: long_context_arxiv.tex
\documentclass[11pt]{article}
\usepackage{natbib}
\usepackage{amsmath}
\usepackage{amsfonts}
\usepackage{amsopn}
\usepackage{epsfig}
\usepackage{ifthen}
\usepackage{algorithm}
\usepackage{algorithmic}
\usepackage{makecell}
\usepackage{multirow}
\usepackage{hyperref}
\usepackage{listings}
\usepackage{graphicx}
\usepackage{subcaption}
\usepackage{booktabs}

\input{template/mymacros.tex} 
\input{template/mymacros2.tex}

\usepackage{makecell}
\usepackage{multirow}

\newcommand{\isarxiv}{yes}

\title{Learning how to Forget: Fine-tuning for Long-Context Sparse Attention}

\author{Matthias Seeger\footnote{Correspondence to \texttt{mseeger@gmail.com}} \\
Amazon Web Services \\
\texttt{mseeger@gmail.com} \\
\and
Zeyu Zhang \\
University of Amsterdam \\
\texttt{z.zhang2@uva.nl} \\
\and
Vihang Patil \\
Amazon \\
\texttt{pvihang@amazon.de}
\and
Konstantinos Benidis \\
Amazon Web Services \\
\texttt{kbenidis@amazon.de}
\and
Sebastian Schelter \\
Technical University Berlin \\
\texttt{schelter@tu-berlin.de}
}

\begin{document}

\maketitle\thispagestyle{empty} 

\begin{abstract}
  \input{longcon_abstract.tex}
\end{abstract}

\input{longcon_introduction.tex}

\input{longcon_relatedwork.tex}

\input{longcon_main.tex}

\input{longcon_experiments.tex}

\input{longcon_conclusions.tex}

\bibliography{template/papers.bib, template/books.bib}
\bibliographystyle{plain}

\appendix

\input{longcon_appendix.tex}

\end{document}

%% file: template/mymacros.tex
\newcommand{\field}[1]{\mathbb{#1}}

\newcommand{\R}{\field{R}}

\newcommand{\vect}[1]{\boldsymbol{#1}} 
\newcommand{\mat}[1]{\boldsymbol{#1}} 
\newcommand{\tvect}[1]{\tilde{\boldsymbol{#1}}}
\newcommand{\tmat}[1]{\tilde{\boldsymbol{#1}}}
\newcommand{\tscal}[1]{\tilde{#1}}
\newcommand{\hvect}[1]{\hat{\boldsymbol{#1}}}
\newcommand{\hmat}[1]{\hat{\boldsymbol{#1}}}
\newcommand{\hscal}[1]{\hat{#1}}
\newcommand{\bvect}[1]{\bar{\boldsymbol{#1}}}
\newcommand{\bmat}[1]{\bar{\boldsymbol{#1}}}
\newcommand{\bscal}[1]{\bar{#1}}
\newcommand{\vzero}{\vect{0}}
\newcommand{\vone}{\vect{1}}
\newcommand{\mxzero}{\mat{0}}

\newcommand{\dummystring}{QWERTYU}
\newcommand{\vci}[3][\dummystr]{\ifthenelse{\equal{#1}{\dummystring}}{\vect{#2}_{#3}}{\vect{#2}_{#3}^{(#1)}}}
\newcommand{\mx}[3][\dummystr]{\ifthenelse{\equal{#1}{\dummystring}}{\mat{#2}_{#3}}{\mat{#2}_{#3}^{(#1)}}}
\newcommand{\tvci}[3][\dummystr]{\ifthenelse{\equal{#1}{\dummystring}}{\tvect{#2}_{#3}}{\tvect{#2}_{#3}^{(#1)}}}
\newcommand{\tmx}[3][\dummystr]{\ifthenelse{\equal{#1}{\dummystring}}{\tmat{#2}_{#3}}{\tmat{#2}_{#3}^{(#1)}}}
\newcommand{\tscl}[3][\dummystr]{\ifthenelse{\equal{#1}{\dummystring}}{\tscal{#2}_{#3}}{\tscal{#2}_{#3}^{(#1)}}}
\newcommand{\hvci}[3][\dummystr]{\ifthenelse{\equal{#1}{\dummystring}}{\hvect{#2}_{#3}}{\hvect{#2}_{#3}^{(#1)}}}
\newcommand{\hmx}[3][\dummystr]{\ifthenelse{\equal{#1}{\dummystring}}{\hmat{#2}_{#3}}{\hmat{#2}_{#3}^{(#1)}}}
\newcommand{\hscl}[3][\dummystr]{\ifthenelse{\equal{#1}{\dummystring}}{\hscal{#2}_{#3}}{\hscal{#2}_{#3}^{(#1)}}}
\newcommand{\bvci}[3][\dummystr]{\ifthenelse{\equal{#1}{\dummystring}}{\bvect{#2}_{#3}}{\bvect{#2}_{#3}^{(#1)}}}
\newcommand{\bmx}[3][\dummystr]{\ifthenelse{\equal{#1}{\dummystring}}{\bmat{#2}_{#3}}{\bmat{#2}_{#3}^{(#1)}}}
\newcommand{\bscl}[3][\dummystr]{\ifthenelse{\equal{#1}{\dummystring}}{\bscal{#2}_{#3}}{\bscal{#2}_{#3}^{(#1)}}}

\DeclareMathOperator{\diag}{diag}

\newcommand{\Ind}[1]{\mathrm{I}_{\{#1\}}}

\newcommand{\figref}[1]{Figure~\ref{fig:#1}}
\newcommand{\tabref}[1]{Table~\ref{tab:#1}}
\newcommand{\secref}[1]{Section~\ref{sec:#1}}
\renewcommand{\eqref}[1]{Eq.~\ref{eq:#1}}
\newcommand{\eqp}[1]{(\ref{eq:#1})}


%% file: template/mymacros2.tex
\newcommand{\vp}[2][\dummystring]{\vci[#1]{p}{#2}}

\newcommand{\vw}[2][\dummystring]{\vci[#1]{w}{#2}}
\newcommand{\vx}[2][\dummystring]{\vci[#1]{x}{#2}}

\newcommand{\vlam}[2][\dummystring]{\vci[#1]{\lambda}{#2}}

\newcommand{\tvv}[2][\dummystring]{\tvci[#1]{v}{#2}}

\newcommand{\tvy}[2][\dummystring]{\tvci[#1]{y}{#2}}

\newcommand{\tvlam}[2][\dummystring]{\tvci[#1]{\lambda}{#2}}

\newcommand{\mxa}[2][\dummystring]{\mx[#1]{A}{#2}}

\newcommand{\mxf}[2][\dummystring]{\mx[#1]{F}{#2}}

\newcommand{\mxk}[2][\dummystring]{\mx[#1]{K}{#2}}

\newcommand{\mxm}[2][\dummystring]{\mx[#1]{M}{#2}}

\newcommand{\mxq}[2][\dummystring]{\mx[#1]{Q}{#2}}

\newcommand{\mxv}[2][\dummystring]{\mx[#1]{V}{#2}}

\newcommand{\mxx}[2][\dummystring]{\mx[#1]{X}{#2}}
\newcommand{\mxy}[2][\dummystring]{\mx[#1]{Y}{#2}}

\newcommand{\tmxf}[2][\dummystring]{\tmx[#1]{F}{#2}}

\newcommand{\rng}[2][1]{{#1},\dots,{#2}}

\newcommand{\srng}[2][1]{\{{#1},\dots,{#2}\}}

%% file: longcon_abstract.tex
A lot of prior work addressed key-value (KV) cache selection and compression by sparse attention to enable long-context inference for transformer language models without excessive hardware budgets. We provide a new method for fine-tuning models with sparse attention. It works for any KV cache policy, runs on a moderate hardware budget (e.g., a single Nvidia A100 GPU with 40 GB RAM), and allows the model to co-adapt with the policy, often outperforming models trained with exact attention (sequence parallelism). We also provide an efficient implementation of H2O sparse attention (the leading policy in our experiments) with dedicated scaled dot product attention kernel support. \ifthenelse{\equal{\isarxiv}{no}}{A new open source library for long-context inference and fine-tuning will be released with the publication.}{\href{https://github.com/awslabs/keys_values}{$\mathtt{KeysAndValues}$}, a new open source library for long-context inference and fine-tuning, provides easy-to-use and performant code for all methods discussed here.}

%% file: longcon_introduction.tex
\section{Introduction}\label{sec:intro}

Modern large language models need to process very long contexts (i.e., number of tokens) for calling many tools with sizable outputs~\citep{Schick:23,Feng:25}, running chain of thought reasoning~\citep{Wei:22}, or sustaining multi-turn conversations. While naive transformer implementations scale quadratically in compute and linearly in memory with context width, a lot of progress has been made on approximations with essentially linear time and constant memory scaling. A particularly fruitful direction is {\em sparse attention}, where key-value (KV) information is stored in a fixed-size KV cache, slots of which are evicted once it is full.

In this paper, we address the open problem of how to {\em post-train} a transformer language model with sparse attention on a moderate hardware budget.
Our novel method works for any KV cache policy and requires no further approximations beyond sparse attention. As we demonstrate in experiments on a range of long-context benchmarks, our training algorithm allows the model to co-adapt with the KV cache policy, often outperforming models trained with exact attention (sequence parallelism). Moreover, our fine-tuning method runs on resources comparable to sparse attention {\em inference}.

The heavy-hitter oracle (H2O)~\citep{Zhang:23} is one of the most prominent sparse attention policies. We provide a much more efficient H2O implementation with dedicated scaled dot product attention (SDPA) kernel support. Variants of H2O outperform other KV cache policies in our experiments, and our fast implementation takes a big step towards latencies competitive with SotA inference libraries such as vLLM~\citep{Kwon:23}.
In summary, our contributions are:
\begin{itemize}
\item
  New method for fine-tuning transformer language models with sparse attention and arbitrary KV cache policy in place, which runs on resources comparable to sparse attention inference. It combines nested activation checkpointing and CPU off\-loading with exploiting a linear KV cache buffer recurrence. Our method can process sequences of arbitrary length with constant GPU resources.
\item
  Improvements of heavy-hitter oracle (H2O) cache policy~\citep{Zhang:23}. In particular, we provide {\em Triton} code to return summed attention weights alongside a FlashInfer SDPA kernel~\citep{Ye:25} (see also \secref{app-sdpa-kernels}).
\item
  A comprehensive evaluation on a range of long-context benchmarks. Our training algorithm strongly outperforms models trained with sequence parallelism~\citep{Li:23} when sparse attention inference is used.
\item
  \ifthenelse{\equal{\isarxiv}{no}}{A novel open source library for long-context inference and fine-tuning, which will be released with the publication of this paper.}{$\mathtt{KeysAndValues}$, a novel open source library for long-context inference and fine-tuning (\url{https://github.com/awslabs/keys_values}).}
\end{itemize}

%% file: longcon_relatedwork.tex
\section{Related Work}\label{sec:relwork}

\vspace{-0.5mm}
There is a large body of work on long-context inference by way of KV cache compression. Grouping heads requires fewer key and value vectors to be stored~\citep{Shazeer:19,Ainslie:23,Brandon:24,Deepseek:24}. Cache buffers can be quantized to 8 or 4 bits~\citep{Hooper:24,Liu:24b,Zhang:24b,Li:25b,Shutova:25,Lancucki:25,Staniszewski:26,Zandieh:26}. Sparse attention comes in many variants. Big Bird~\citep{Zaheer:20} uses fixed attention sparsity patterns. The heavy-hitter oracle (H2O)~\citep{Zhang:23} is discussed in \secref{main-h2o}, extensions include Q-Hitter~\citep{Zhang:24}, SnapKV~\citep{Li:25}, and CAKE~\citep{Qin:25}. FlexGen~\citep{Sheng:23} maximizes throughput by using a cache hierarchy. FastGen~\citep{Ge:24} shows how to vote between different cache policies. Other sparse attention techniques include~\citep{Han:24,Xiao:24,Tang:24,Cai:25,Xiao:25,Tang:25,Feng:25b,Yang:25b,Wang:25,Devoto:25}. KVPop~\citep{Hauzenberger:26} learns a cache policy against a future-attention target.
qTTT~\citep{Bansal:26} uses a few gradient updates at test time in order to improve inference results. MInference~\citep{Jiang:24}, KVPress~(\url{https://github.com/NVIDIA/kvpress}) and ShadowKV~\citep{Sun:25} are open source libraries providing sparse attention methods. SCBench~\citep{Li:25c} provides an empirical analysis.

Optimized scaled dot product attention (SDPA) kernels are essential for fast inference and training, pioneered by FlashAttention~\citep{Dao:23,Shah:24}. FlashInfer~\citep{Ye:25} is optimized for the inference case $1\ll N_q\ll N_k$ (notation from \secref{main-kv-cache}). FlexAttention~\citep{Dong:25} allows users to specify mask and score modification code.

Our main contribution is on long-context {\em fine-tuning}. Prior work can be ordered into two groups. Proposals in the first group modify multi-head attention in ways which remedy the difficulties detailed in \secref{main-finetune}.
Native sparse attention (NSA)~\citep{Yuan:25} bakes a mixture of some sparse attention kernels with different fixed policies into the model architecture. DeepSeek sparse attention (DSA)~\citep{Deepseek:25b} is a refined variant of this idea. Different to {\em selective} sparse attention, this needs to be used during pre-training already, whose cost is significantly increased. IndexCache~\citep{Bai:26} speeds up DSA somewhat by sharing the indexer between several layers. Also, by fixing the selection policy as part of the model choice, DSA offers considerable less flexibility during post-training. LSA and DSA still require storing the complete KV cache (even though each attention call uses only a part of it). LongGen~\citep{Ge:25} uses static sparsity patterns for the top and bottom third of layers and full attention for the middle third, which speeds up training a bit, but does not reduce its memory requirements. DMC~\citep{Nawrot:24} uses a form of KV cache compression where new information is either appended or accumulated with the last recent slot. They propose heuristics to train a model with this mechanism in place. This approach seems restricted to KV cache updates during generation, it is not clear what is done when a large prompt needs to be processed. Starting from Mamba~\citep{Gu:23}, there have been several attempts to resurrect LSTMs~\citep{Hochreiter:97}, e.g.~\citep{Dao:24,Beck:25}. When compared against the long-context transformer SotA, none of them have been competitive enough in order to warrant costly pre-training efforts. YOCO~\citep{Sun:24} changes the transformer architecture, so that a single KV cache block serves all layers. KV buffer scaling with layers is not a major problem in practice, since we can use CPU offloading (see \secref{app-chunks-cells}).

In the second group, KV cache buffers are not compressed, but both storage and computation are distributed across several devices. This can be done with RingAttention~\citep{Liu:24} or more recent variants~\citep{Liu:25}, in what is called context parallelism (CP), or with sequence parallelism (SP)~\citep{Li:23}. While \cite{Li:25d} observe that long sequences can be split into chunks, and that the computation graph factorizes along the chunk axis, the factor for the final chunk depends on {\em all} KV cache buffers of all layers, so cannot be represented on a single device\footnote{
  In fact, RingAttention's way to compute gradients makes use of this graph structure implicitly.} (attempts to sparsify this computation are heuristics, and experiments are done only on rather short sequences of up to 16k tokens). OOMB~\citep{Li:26} shares properties with our work, such as chunk-level processing, activation checkpointing, and efforts to compress KV cache buffers for {\em autograd}. While for exact KV caches, they run into the same issues as \citep{Li:25d}, their implementation supports NSA and DSA as well. They subscribe to the paged memory model of vLLM~\citep{Kwon:23}, for which modern sparse attention policies like H2O cannot be implemented. Their way of hiding KV cache buffers from {\em autograd} requires a number of complex dedicated CUDA kernels, which need to be hardcoded for every sparse attention variant. We manage KV buffer size in {\em autograd} via delta encoding, which renders our implementation agnostic to the KV cache policy. Details are given in \secref{app-relwork}. LongRoPE~\citep{Ding:24,Shang:25} combines SP with a search for non-uniform RoPE and several lifting stages. Other work on tuning position encoding, data mix and fine-tuning recipes, but not compressing KV cache, includes~\citep{Wu:24,Zhang:24c,Gao:25}. LongStraw~\citep{Zhou:26b} is a system designed for long context reinforcement learning, with a specific emphasis on sharing prompt graphs and KV caches between different roll-outs.
Highly optimized implementations of CP/SP are the state of the art for long-context inference, e.g.\ vLLM~\citep{Kwon:23}, SGLang~\citep{Zheng:24}, and fine-tuning, e.g.\ Nvidia NeMo RL (\url{https://github.com/NVIDIA-NeMo/RL}), MS-SWIFT~\citep{Zhao:25}. In \secref{main-satt-vs-cp}, we comment on reasons why sparse attention methods are less frequently used in practice, and how this could be changed.

%% file: longcon_main.tex
\section{Long-Context Fine-Tuning}\label{sec:main}

In this section, we introduce sparse attention and key-value caching, providing several improvements to the heavy-hitter oracle (H2O) cache policy~\citep{Zhang:23}, such as an efficient implementation with dedicated SDPA kernel support. We detail our main contribution: a novel method to fine-tune models with sparse attention in place, using resources comparable to sparse attention inference. While in SotA CP/SP techniques, GPUs need to be used for sharding along the context, they can be used to increase throughput (i.e., larger batch sizes) or to handle larger models in our method.

\subsection{Sparse Attention. Key-Value Caching}\label{sec:main-kv-cache}

{\em Multi-head attention (MHA)} is the most important mechanism in modern transformer architectures~\citep{Vaswani:18}. At its core lies {\em scaled dot product attention (SDPA)}:
\begin{equation}\label{eq:main-sdpa}
  \mxy{} = \mathtt{SDPA}(\mxq{}, \mxk{}, \mxv{}),\quad
  \mxy{}, \mxq{}\in\R^{(B, H_q, N_q, d_h)},\;
  \mxk{}, \mxv{}\in \R^{(B, H_k, N_k, d_h)}.
\end{equation}
Here, $\mxq{}$ (queries), $\mxk{}$ (keys), $\mxv{}$ (values) are 4D arrays, $B$ is the batch size, $d_h$ the {\em per-head embedding dimension}, $H_q, H_k$ are numbers of heads, and $N_q, N_k$ are sequence lengths (i.e., the third dimension maps to token positions in the model context). The {\em model embedding dimension} is $d = H_q\cdot d_h$. Let us first assume that $H_q = H_k$ and drop the first two dimensions. Then:
\begin{equation}\label{eq:main-sdpa-comp}
   \mxy{} = \mxm{}\mxv{},\quad \mxm{} = \mathtt{softmax}\left( \mathtt{mask}\left( d_h^{-1/2} \mxq{}\mxk{}^T \right), \mathtt{dim=1}\right).
\end{equation}
$\mxy{}$ are weighted combinations of values $\mxv{}$ with {\em attention weights} $\mxm{}\in\R^{(B, H_q, N_q, N_k)}$, $\mathtt{softmax}$ applies $\vx{}\mapsto \exp(\vx{}) / (\vone^T\exp(\vx{}))$ along rows. $\mathtt{mask}$ implements causal masking: $(\mathtt{mask}(\mxx{}))_{i, j} = x_{i, j} - \infty \Ind{P + i < t(j)}$, where $P$ and $t(j)$ are token positions (see \secref{app-sdpa-kernels} for details). If arrays are indexed by $(b, h, j, k)$, SDPA operates on $(j, k)$ in the same way for all $(b, h)$, computations are parallelized over batch and head positions.

Inference in transformers switches between prompt processing and token generation. Generating a token after a prompt of size $N$ requires SDPA with $N_k = N, N_q = 1$, with keys and values of size $(B, H_k, N, d_h)$ in memory, for each of $L$ model layers. Exact transformer inference therefore requires $\mathcal{O}(L\cdot N\cdot B H_k d_h)$ memory for the full {\em key-value (KV) cache}. Even for moderate context lengths $N$ of several hundred thousands, the KV cache far surpasses the model weights in size and cannot be stored in GPU memory as is.

A large amount of prior work confronts this problem (see also \secref{relwork}). In grouped query attention (GQA)~\citep{Ainslie:23}, we set $H_q = H_k\cdot q_g$, $q_g > 1$, so that $q_g$ heads map to the same query group, which reduces KV cache size by a factor of $q_g$. Most relevant to our work is {\em sparse attention} (or {\em selective KV caching}; e.g.~\cite{Zhang:23}), where the KV cache is represented by fixed-size buffers {\em independent} of the context length of the model. Once all slots are filled, new information overwrites (or {\em evicts}) existing ones. Formally, the cache (for one model layer) is represented by arrays $\mathtt{keys}, \mathtt{values}: (B, H_k, N_C, d_h)$ and $\mathtt{token\_pos}: (B, H_k, N_C)$.
Here, $N_C$ is the {\em cache length}, which is chosen as large as GPU memory permits. For up to $N_C$ tokens, the cache is filled from left to right. After that, the {\em KV cache policy} $\pi_l(b, h, t)\in\srng[0]{N_C - 1}$ dictates where additional key-value information is written for token position $t\ge N_C$, batch position $b\in\srng[0]{B-1}$ and head (or query group) $h\in\srng[0]{H_k - 1}$. Importantly, $\pi_l$ can depend on $b, h$: a token may be in the cache for some batch positions and heads, and not for others. $t(b, h, j) = \mathtt{token\_pos}[b,h,j]$ lists the token position of what is stored in $(b, h, j)$. We do not require complex memory layouts and dedicated kernels for this setup, as for example PagedAttention~\citep{Kwon:23} needs. In fact, \eqp{main-sdpa-comp} depends on absolute token positions only via $\mathtt{mask}$. This is as sparse as the conventional triangular one, but depends on token positions $t(\cdot) = \mathtt{token\_pos}$, which through cache evictions becomes non-monotonic in general. Moreover, $\mathtt{torch.gather}$ and $\mathtt{torch.scatter}$ provide fast read and write access for these buffers (see \secref{app-scatter-gather}).

With sparse attention, we can run inference for any context width $N$. First, we process up to the first $N_C$ tokens with a single SDPA call ($N_q = N_k = N_C$), this is known as {\em prefilling}. The remaining $N - N_C$ tokens are processed in {\em chunks} of size $S < N_C$, using SDPA calls with $N_q = S, N_k = N_C$ (see \secref{app-chunks-cells} for details). Token generation uses $N_q = 1, N_k = N_C$. Memory requirements are independent of $N$. For each chunk, the policy $[\pi_l]$ is used to determine $B\cdot H_k\cdot S$ positions $(b, h, j)$ which are overwritten by the new keys and values. While $N_C$ is chosen as large as memory permits, the choice of $S$ is more subtle. The larger $S$, the fewer chunks, and less sequential computation results in faster processing. But if $S$ is too large, the cache policy is used in a coarse-grained way only, and performance may degrade (see also \secref{main-satt-vs-cp}).

\subsubsection{Variants of Heavy-Hitter Oracle}\label{sec:main-h2o}

The key idea behind the heavy-hitter oracle (H2O)~\citep{Zhang:23} is to make use of the attention weights $\mxm{} = [m_{i, j}]$, a by-product of SDPA \eqp{main-sdpa-comp}. Dropping $(b, h)$ for the moment, we have that $\mxy{i, :} = \sum_j m_{i, j} \mxv{j, :}$ and $\sum_j m_{i, j} = 1$. $m_{i, j}$ quantifies how much values $\mxv{j, :}$ are used to create $\mxy{i, :}$. The cumulative sum $\sum_i m_{i, j}$ can be used to score the usefulness of vectors $(\mxk{j, :}, \mxv{j, :})$ in the KV cache. Define the H2O score after having processed $t$ tokens as
\begin{equation}\label{eq:h2o-score}
  \phi^t_{\text{h2o}}(b, h, j) = \sum_{ t(b,h,j)\le s < t} m_{b, h, s, j},
\end{equation}
where $t(b, h, j) = \mathtt{token\_pos}[b, h, j]$ is the position represented at $(b, h, j)$ right now. We sum over $s\in\srng[t(b,h,j)]{t}$ because the slot holds KV information for token position $t(b,h,j)$, which entered the cache only then. The larger $\phi^t_{\text{h2o}}(b, h, j) $, the more valuable this information has been so far. When asked to insert new content for $S$ tokens, for each $(b, h)$, we overwrite these $S$ slots $j$ for which $\phi^t_{\text{h2o}}(b, h, j)$ is smallest.

In this paper, we modify the original H2O policy~\citep{Zhang:23} (as provided by their implementation) in several ways. First, their code selects the same cache slots for each batch position $b$, using the score $\phi^t_{\text{h2o-orig}}(h, j) = \sum_b \phi^t_{\text{h2o}}(b, h, j)$. The rationale for this restriction is unclear, and we implement H2O without it as well. Second, the cumulative H2O score seems to favour entries $(b, h, j)$ which have been in the cache for longer, since more terms in $[0, 1]$ are summed then. We introduce the {\em normalized H2O score}: $\phi^t_{\text{h2o-norm}}(b, h, j) = (t - t(b,h,j))^{-1} \phi^t_{\text{h2o}}(b, h, j)$.

Despite convincing empirical results of H2O, both in \citep{Zhang:23} and \secref{experiments}, it is not widely used. This is mostly because current implementations of H2O are much slower than the state of the art. We need summed attention weights $\sum_i m_{b, h, i, j}$ for each $(b, h, j)$, as by-product of SDPA \eqp{main-sdpa-comp}, but none of the fast SDPA kernels derived from FlashAttention~\citep{Dao:23} provide them. Current H2O implementations use naive SDPA implementations, which are much too slow in practice. Our implementation contains {\em Triton} code to return summed attention weights alongside a FlashInfer SDPA kernel~\citep{Ye:25}. In \secref{app-attn-weights}, we show how FlexAttention~\citep{Dong:25} can be used to this end as well.

\subsection{Fine-Tuning for Sparse Attention}\label{sec:main-finetune}

How should we train a model which uses sparse attention with some KV cache policy such as H2O? As noted in \secref{relwork}, all prior long-context fine-tuning methods either restrict the MHA approximation to a particular form, or use exact MHA with KV cache buffers distributed across several devices (i.e., sequence or context parallelism). However, the choice of KV cache policy, which dictates how the model's short term memory is organized, should influence how the model is best trained. Our results in \secref{experiments} validate this hypothesis. {\em Sparse attention inference for a model trained with exact MHA and sequence parallelism (which is the SotA) often performs significantly worse than for a model trained with sparse attention and the desired policy in place}.

Fine-tuning for models with sparse attention is difficult, because a huge amount of memory is required. Gradients for loss functions on sequences of length $N\gg N_C$ are computed by reverse mode automatic differentiation (or error backpropagation, {\em autograd}). This creates a computation graph during the forward pass, storing arrays needed during the backward pass. With $L$ layers and chunk size $S$, the training sequence is split into $1 + \lceil(N - N_C)/S\rceil$ chunks, the first (prefill) chunk of length $N_C$ and subsequent chunks of length $S$. SDPA is called for each layer and chunk, creating at least one node of the size of the KV cache, so we need at least $\mathcal{O}(L\cdot S^{-1} (N - N_C)\cdot N_C\cdot \mathcal{D})$ of memory, where $\mathcal{D} = B H_k d_h$. Assuming $S = a N_C$ for some constant $a$, this is $\mathcal{O}(L\cdot N\cdot \mathcal{D})$: more than the {\em full} KV cache would need, and far beyond what is tractable.

\begin{figure}[ht!]
\begin{center}
  \includegraphics[width=\textwidth]{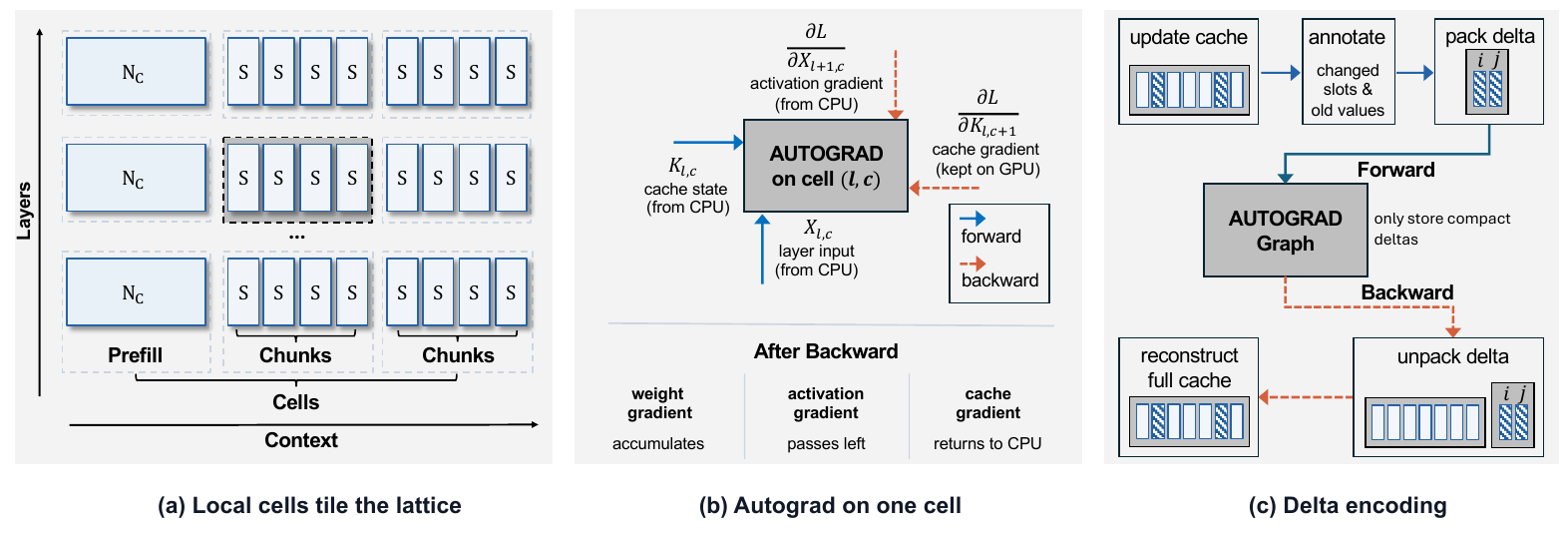}
\end{center}
\caption{\label{fig:method-diagram}
  (a) Structure of computation graph as lattice of chunks. The first (prefill) chunk in a row has full length $N_C$, subsequent chunks have length $S < N_C$. Chunks are grouped into cells (see \secref{app-chunks-cells}).
  (b) {\tt autograd} is run separately on each cell $(l, c)$. This needs checkpointed inputs from bottom and left, and head gradients from top (checkpointed) and right (on GPU). The direction is right to left, and head gradients for layer inputs are written back to CPU. See \secref{app-grad-summary} for details and notation.
  (c) Delta encoding of KV cache buffers during {\tt autograd} on a cell. During the forward pass, deltas of the buffer sequences are extracted, matched in {\tt pack\_hook} calls, and written to the computation graph. During the backward pass, the buffer sequences are reconstructed from the deltas and supplied to {\tt unpack\_hook}. See \secref{app-exploit-recurrence} for details.}
\end{figure}

Our goal is to bring GPU memory requirements down to levels comparable to what inference needs. First, we avoid differentiation through the KV cache policy (which is often not even possible, and in general not tractable). Along the forward pass, we store all KV cache policy decisions in a {\em replay log}, containing for each chunk (and each layer $l$) the tokens processed, and the decisions $\{\pi_l(b, h, t)\}$. Later on, we use {\em replay caches}, which act like normal KV caches, except that eviction decisions are replayed from the log.

Next, we use activation checkpointing~\citep{Herrmann:19}. Different from autograd, there is no computation graph built during the forward pass, only the input tensors for each transformer layer are stored to CPU memory.\ifthenelse{\equal{\isarxiv}{no}}{}{\footnote{
  Our implementation allows for cells spanning several layers in the vertical direction. For notational simplicity, we restrict ourselves to one layer per cell here.}}
The backward pass is split into $L$ autograd calls, starting from the top. Head gradients are supplied from the previous layer, inputs are loaded from CPU. Computation graphs on GPU are $L$ times smaller, while forward computations have to be run twice.

While standard activation checkpointing tackles large $L$, in long-context situations the context width $N$ is the more serious problem, in that not even complete inputs or head gradients for a single layer can be kept in GPU memory (see \secref{app-chunks-cells}). We use activation checkpointing twice, in a nested fashion. To this end, we partition chunks into {\em cells}: $\{(B, S, d)\}\to (B, k S, d)$, where $k = \lfloor \alpha N_C / S \rfloor$, and $\alpha > 0$ is a hyperparameter which defaults to $\alpha = 1$. The first (prefill) chunk becomes the first cell. As a rule of thumb, we group chunks into cells which occupy about the size of KV cache buffers. The computation graph becomes a {\em lattice of cells}: rows are layers, columns are cells (i.e., groups of chunks) along the context (\figref{method-diagram}, left). Our backward pass runs an outer loop over rows, then inner loops over cells in each row. Each inner loop starts with a (non-autograd) forward pass along the context, storing KV cache buffers going into each cell to CPU\footnote{
  Our implementation writes checkpoints to disk once CPU memory is occupied.}.
Next, autograd is run separately on each cell, starting from the right. Inputs to a cell (layer inputs from the bottom and KV cache buffers from the left) and head gradients from the top are read from CPU, while head gradients from the right stay in GPU memory (\figref{method-diagram}, middle). Gradients are accumulated. We reuse the same CPU and GPU buffers for all inner loops.
A detailed summary of our method is given in \secref{app-grad-summary}.

However, the autograd calls still need too much GPU memory. Recall that a cell consists of $k = \lfloor \alpha N_C / S \rfloor$ chunks. Autograd stores KV cache buffers for each chunk, so needs at least $\mathcal{O}(k\cdot N_C\cdot \mathcal{D})$ memory, where $\mathcal{D} = B H_k d_h$. As detailed in \secref{main-satt-vs-cp}, $k$ must be sizable to allow cache policies to make good decisions. We now show how to cut GPU memory by a factor of $k$, to $\mathcal{O}(N_C\cdot \mathcal{D})$ per autograd call, comparable to what is needed during inference alone. Consider KV cache buffers $(\mathtt{keys}, \mathtt{values})$, $(\mathtt{keys}', \mathtt{values}')$ for neighboring chunks. Their size is $(B, H_k, N_C, d_h)$, but they only differ in $S\cdot \mathcal{D}$ values, because a chunk consists of $S$ tokens only. The relationship is simple:
\[
  \mathtt{keys}' = \mathtt{scatter}(\mathtt{keys}, \mathtt{index}, \mathtt{key\_new}),\;
  \mathtt{values}' = \mathtt{scatter}(\mathtt{values}, \mathtt{index}, \mathtt{value\_new}).
\]
Here, $\mathtt{key\_new}, \mathtt{value\_new}$ are KV vectors for new tokens with sizes $(B, H_q, S, d_h)$, $\mathtt{index}$ is based on the cache policy $\pi_l$, determining which slots are overwritten, and $\mathtt{scatter}, \mathtt{gather}$ are linear $\mathtt{torch}$ operators defined in \secref{app-scatter-gather}. This is a linear recurrence, which is easily inverted:
\begin{equation}\label{eq:recurrence-reverse}
  \mathtt{keys} = \mathtt{scatter}(\mathtt{keys}', \mathtt{index}, \mathtt{delta\_key}),\quad
  \mathtt{delta\_key} = \mathtt{gather}(\mathtt{keys}, \mathtt{index}),
\end{equation}
and the same for $\mathtt{values}$. Instead of storing $\mathtt{keys}, \mathtt{values}$ for each chunk in the compute graph, it suffices to store\footnote{
  This is a form of {\em delta encoding} (\url{https://en.wikipedia.org/wiki/Delta_encoding}).
} $\mathtt{delta\_key}, \mathtt{delta\_value}$ (\figref{method-diagram}, right). Since memory requirements of autograd calls are dominated by the KV cache buffers, they are reduced by a factor of $k$.

While the linear recurrence relation between neighboring cache buffers is simple, implementing it in the context of {\em PyTorch autograd} is not. We use a mechanism called {\em autograd saved tensors hooks}\footnote{
  \small\url{https://docs.pytorch.org/tutorials/intermediate/autograd_saved_tensors_hooks_tutorial.html}},
originally intended to implement activation checkpointing by CPU offloading, which can be adapted to ours needs. In a nutshell, we use the {\em PyTorch} mechanism to store $(\mathtt{delta\_key}, \mathtt{delta\_value})$ in the autograd graph in place of $(\mathtt{keys}, \mathtt{values})$ (called "packing"), reconstructing the latter from the former and subsequent $(\mathtt{keys}', \mathtt{values}')$ during the backward pass over chunks (called "unpacking"). The key difficulty is the non-selectiveness of the mechanism: it provides a $\mathtt{pack\_hook}$ function called for all arrays {\em PyTorch autograd} decides to place into its graph. There is no way to tag tensors in the forward code, so they can be recognized as $\mathtt{pack\_hook}$ arguments. Our solution is to create {\em annotations} alongside the forward pass code, storing $(\mathtt{index}, \mathtt{delta\_key})$ for $\mathtt{keys}$, $(\mathtt{index}, \mathtt{delta\_value})$ for $\mathtt{values}$. In a $\mathtt{pack\_hook}(\vx{})$ call, we relate $\vx{}$ to current annotations: $\vx{}$ matches $(\mathtt{index}, \mathtt{delta\_key})$ (say) if $\mathtt{gather}(\vx{}, \mathtt{index}) = \mathtt{delta\_key}$. A match leads to $\mathtt{pack\_hook}(\vx{})$ returning a reference to $(\mathtt{index}, \mathtt{delta\_key})$, which is removed from the annotation list. Note that failing to match an annotation does not lead to errors, but at most to a bit more memory being used. More details are given in \secref{app-exploit-recurrence}.

\subsection{Sparse Attention and Sequence Parallelism}
\label{sec:main-satt-vs-cp}

While with context or sequence parallelism, the context width is strictly limited by the number of GPUs available, sparse attention inference can be run for any context width on moderate GPU resources. Redundancies, which exist in multi-head attention, can be exploited by way of cache compression.
Even if many GPUs are available, sparse attention allows us to increase batch size by way of distributed data parallel (DDP). A large number of sparse attention variants have been proposed (see \secref{relwork}). {\em Why is it then that sparse attention methods are hardly used in SotA inference libraries such as vLLM}~\citep{Kwon:23}? The short answer is that latency is significantly higher with existing sparse attention implementations. Further comments are in \secref{app-sparse-attn-libraries}.

Some of the gap is due to less low level implementation support for sparse attention. Highly optimized SDPA kernels are vital for fast inference~\citep{Dao:23, Dong:25, Ye:25}. However, as noted in \secref{main-h2o}, existing kernels do not cater for sparse attention inference (see details in \secref{app-sdpa-kernels}). Other reasons for the gap are more difficult to address. A long sequence is split into chunks, the first (prefill) chunk of length $N_C$ (cache length), subsequent chunks of length $S$. Larger $S$ means fewer chunks and lower latency. But KV cache policies can make useful eviction decisions only if $S$ is much smaller than $N_C$ (in our experiments in \secref{experiments}, we use $N_C = 32768$ and $S\in\{1024, 2048, 8192\}$). For the extreme choice $S = N_C$, the whole cache is overwritten by new content for every chunk, and the KV cache policy plays no role at all! Information cannot be kept in the cache beyond a chunk if we do not allow for substantial overlap.

For example, suppose we use 8 devices, each supporting a cache length $N_C$, and the sequence length is $N = 8 N_C$. With RingAttention, each device holds $N_C$ slots, and the sequence is processed in 8 sequential chunks. But for sparse attention, we need $1 + 7 N_C / S$ chunks, which can be substantially larger. Despite sparse attention supporting an 8 times larger batch size via DDP, inference tends to still be slower than with RingAttention. In future work, we plan to improve sparse attention latency by appropriate kernel fusion. However, the sequential nature of decision making in sparse attention may be an inherent disadvantage over sequence or context parallelism.

Apart from large hardware requirements to even work on long sequences, RingAttention requires $\mathcal{O}(L\cdot D)$ synchronizations between all devices per gradient update, while sparse attention DDP only needs a single gradient averaging reduction. While memory transfer between devices can be run in parallel with computations, this needs double buffering\footnote{
  On each device, one set of $\mathtt{keys}, \mathtt{values}$ are read for computation and transfer, another set is written to~\citep{Liu:24}.}
in RingAttention, doubling the GPU memory needed. The peer-to-peer memory transfer is more brittle than DDP used with sparse attention, and robust training code is more difficult to implement.

Moreover, most implementations of sequence or context parallelism, as well as sparse variants such as DSA~\citep{Deepseek:25b}, subscribe to the paged memory model of vLLM~\citep{Kwon:23}, where CPU--GPU transfer is done in pages. Unfortunately, in vLLM a page must span over all heads and even across a contiguous range of tokens, which makes it impossible to implement modern sparse attention policies such as H2O. We would argue that {\em for efficient sparse attention, paging is detrimental}. If heads specialize to different ways of encoding the past, why force them to attend to the same tokens? And why should the most relevant KV information come in contiguous ranges of 64 or even 128 tokens? This argument is discussed in more detail in \secref{app-sparse-attn-libraries}. Finally, the "waste by pre-allocation" issues which motivate PagedAttention have a simpler solution with sparse inference: KV cache buffers are of a fixed length, there is no need to split\footnote{
  If we frequently encounter batches shorter than the cache length, we can build up buffers in chunks. But most modern agentic AI applications come with long prompts anyway.}
them into pages along the sequence axis, and no custom SDPA code is needed.

A surprising outcome from our experiments is that a smaller chunk size $S$ does not necessarily imply better performance. Results for $S=128$ in \tabref{results-ablations} are not significantly different from results for $S=2048$ in \tabref{helmet-results}, despite the former requiring up to 150\% more training time (see \tabref{time-train-64k}). Moreover, if the chunk size is increased to $S=8192$, results are only slightly worse for some policies and not significantly different for the best performing ones (see \tabref{results-ablations}), while training times are much less (see \tabref{time-train-64k} and \tabref{time-train-128k}). We recommend to {\em choose the chunk size $S$ as large as possible without witnessing a performance degradation}.

%% file: longcon_experiments.tex
\setlength{\tabcolsep}{4.25pt}
\begin{table}[ht!]
\centering
\begin{tabular}{|l|rr|rr|rr|rr|rr|rr|rr|rr|}
\hline
 & \multicolumn{8}{c|}{64k datasets} & \multicolumn{8}{c|}{128k datasets} \\
 \hline
 & \multicolumn{2}{c|}{nq} & \multicolumn{2}{c|}{tri\_qa} & \multicolumn{2}{c|}{hot\_qa} & \multicolumn{2}{c|}{pop\_qa}
 & \multicolumn{2}{c|}{nq} & \multicolumn{2}{c|}{tri\_qa} & \multicolumn{2}{c|}{hot\_qa} & \multicolumn{2}{c|}{pop\_qa} \\
 & us & sp & us & sp & us & sp & us & sp
 & us & sp & us & sp & us & sp & us & sp \\
\hline\hline
\rule{0pt}{13pt} $\mathrm{exact}$ &
  - & {\small\!48.8} &
  - & {\small\!78.7} &
  - & {\small\!53.3} &
  - & {\small\!59.3} &
  - & {\small\!52.5} &
  - & {\small\!55.2} &
  - & {\small\!47.7} &
  - & {\small\!45.7} \\
\hline
\rule{0pt}{13pt} $\mathrm{lr}_{2k}$ &
  {\small\!32.7} & {\small\!14.7} &
  {\small\!56.8} & {\small\!48.5} &
  {\small\!43.3} &  {\small\!5.7} &
  {\small\!44.8} & {\small\!28.5} &
  {\small\!21.2} &  {\small\!9.0} &
  {\small\!55.2} & {\small\!23.5} &
  {\small\!27.7} &  {\small\!5.0} &
  {\small\!31.0} & {\small\!37.3} \\
\rule{0pt}{11pt} $\mathrm{slr}_{2k}$ &
  {\small\!32.3} & {\small\!15.7} &
  {\small\!58.2} & {\small\!43.7} &
  {\small\!43.3} &  {\small\!6.3} &
  {\small\!41.7} & {\small\!24.7} &
  {\small\!19.5} &  {\small\!9.0} &
  {\small\!50.7} & {\small\!20.2} &
  {\small\!23.3} &  {\small\!7.3} &
  {\small\!30.8} & {\small\!37.3} \\
\rule{0pt}{11pt} $\mathrm{h2o}_{2k}$ &
  {\small\!41.0} & {\small\!18.2} &
  {\small\!68.2} & {\small\!54.2} &
  {\small\!35.7} &  {\small\!8.7} &
  {\small\!47.3} & {\small\!41.2} &
  {\small\!16.3} &  {\small\!7.2} &
  {\small\!39.5} & {\small\!26.5} &
  {\small\!17.0} &  {\small\!6.0} &
  {\small\!35.8} & {\small\!35.0} \\
\rule{0pt}{11pt} $\mathrm{h2o}_{2k}^{\text{no}}$ &
  {\small\!45.2} & {\small\!16.2} &
  {\small\!72.8} & {\small\!56.7} &
  {\small\!46.0} &  {\small\!7.7} &
  {\small\!48.0} & {\small\!41.3} &
  {\small\!38.5} & {\small\!14.5} &
  {\small\!67.2} & {\small\!29.0} &
  {\small\!30.7} & {\small\!10.0} &
  {\small\!45.0} & {\small\!41.2} \\
\rule{0pt}{11pt} $\mathrm{h2o}_{2k}^{\text{or}}$ &
  {\small\!40.8} & {\small\!18.5} &
  {\small\!70.7} & {\small\!54.7} &
  {\small\!42.7} &  {\small\!9.0} &
  {\small\!49.2} & {\small\!42.0} &
  {\small\!37.0} & {\small\!14.7} &
  {\small\!70.2} & {\small\!39.0} &
  {\small\!22.0} &  {\small\!6.3} &
  {\small\!48.0} & {\small\!46.7} \\
\rule{0pt}{11pt} $\mathrm{lr}_{1k}$ &
  {\small\!32.3} & {\small\!15.5} &
  {\small\!64.5} & {\small\!45.8} &
  {\small\!41.0} &  {\small\!6.0} &
  {\small\!43.5} & {\small\!26.2} &
  {\small\!20.8} &  {\small\!8.5} &
  {\small\!56.7} & {\small\!20.0} &
  {\small\!26.7} &  {\small\!5.7} &
  {\small\!32.5} & {\small\!38.0} \\
\rule{0pt}{11pt} $\mathrm{slr}_{1k}$ &
  {\small\!30.5} & {\small\!14.0} &
  {\small\!66.3} & {\small\!49.5} &
  {\small\!45.0} &  {\small\!6.0} &
  {\small\!39.5} & {\small\!25.2} &
  {\small\!22.0} &  {\small\!9.7} &
  {\small\!45.3} & {\small\!19.8} &
  {\small\!22.0} &  {\small\!5.0} &
  {\small\!27.8} & {\small\!36.8} \\
\rule{0pt}{11pt} $\mathrm{h2o}_{1k}$ &
  {\small\!43.5} & {\small\!18.0} &
  {\small\!67.2} & {\small\!53.5} &
  {\small\!47.3} &  {\small\!6.3} &
  {\small\!43.2} & {\small\!39.7} &
  {\small\!14.8} &  {\small\!7.2} &
  {\small\!41.2} & {\small\!29.7} &
  {\small\!13.7} &  {\small\!7.7} &
  {\small\!31.7} & {\small\!39.0} \\
\rule{0pt}{11pt} $\mathrm{h2o}_{1k}^{\text{no}}$ &
  {\small\!42.8} & {\small\!16.2} &
  {\small\!63.2} & {\small\!53.5} &
  {\small\!47.3} &  {\small\!9.3} &
  {\small\!48.8} & {\small\!37.8} &
  {\small\!35.5} & {\small\!12.3} &
  {\small\!56.2} & {\small\!30.2} &
  {\small\!30.6} & {\small\!11.3} &
  {\small\!42.5} & {\small\!39.7} \\
\rule[-5pt]{0pt}{16pt} $\mathrm{h2o}_{1k}^{\text{or}}$ &
  {\small\!43.2} & {\small\!17.3} &
  {\small\!67.7} & {\small\!56.7} &
  {\small\!41.3} &  {\small\!6.7} &
  {\small\!47.2} & {\small\!41.2} &
  {\small\!38.0} & {\small\!16.5} &
  {\small\!61.5} & {\small\!40.8} &
  {\small\!24.7} &  {\small\!5.7} &
  {\small\!47.2} & {\small\!44.0} \\
\hline\hline
   & \multicolumn{16}{c|}{128k datasets} \\
 \hline
 & \multicolumn{2}{c|}{trec\_c} & \multicolumn{2}{c|}{trec\_f} & \multicolumn{2}{c|}{nlu} & \multicolumn{2}{c|}{bank77}
 & \multicolumn{2}{c|}{clc150} & \multicolumn{2}{c|}{inf\_qa} & \multicolumn{2}{c|}{inf\_mc} & \multicolumn{2}{c|}{json\_kv} \\
 & us & sp & us & sp & us & sp & us & sp
 & us & sp & us & sp & us & sp & us & sp \\
\hline\hline
\rule{0pt}{13pt} $\mathrm{exact}$ &
  - & {\small\!94.6} &
  - & {\small\!85.4} &
  - & {\small\!90.6} &
  - & {\small\!91.4} &
  - & {\small\!97.4} &
  - & {\small\!44.2} &
  - & {\small\!64.0} &
  - & {\small\!100} \\
\hline
\rule{0pt}{13pt} $\mathrm{slr}_{1k}$ &
  {\small\!95.8} & {\small\!40.0} &
  {\small\!84.6} & {\small\!19.0} &
  {\small\!89.6} & {\small\!20.4} &
  {\small\!91.6} & {\small\!17.8} &
  {\small\!97.2} & {\small\!12.0} &
  {\small\!25.3} &  {\small\!2.2} &
  {\small\!41.0} & {\small\!23.0} &
  {\small\!48.0} & {\small\!13.0} \\
\rule{0pt}{11pt} $\mathrm{h2o}_{1k}^{\text{no}}$ &
  {\small\!96.0} & {\small\!33.4} &
  {\small\!90.6} & {\small\!13.2} &
  {\small\!86.6} & {\small\!18.0} &
  {\small\!91.0} & {\small\!19.8} &
  {\small\!97.8} & {\small\!19.8} &
  {\small\!32.1} &  {\small\!2.6} &
  {\small\!53.0} & {\small\!26.0} &
  {\small\!52.0} &  {\small\!1.0} \\
\rule{0pt}{11pt} $\mathrm{h2o}_{1k}^{\text{or}}$ &
  {\small\!96.2} & {\small\!51.6} &
  {\small\!77.2} & {\small\!76.8} &
  {\small\!80.6} & {\small\!11.4} &
  {\small\!90.2} & {\small\!65.2} &
  {\small\!95.6} & {\small\!59.8} &
  {\small\!34.7} &  {\small\!4.9} &
  {\small\!50.0} & {\small\!33.0} &
   {\small\!1.0} &  {\small\!2.0} \\
\rule{0pt}{11pt} $\mathrm{slr}_{8k}$ &
  {\small\!96.0} & {\small\!41.4} &
  {\small\!87.6} & {\small\!18.4} &
  {\small\!87.2} & {\small\!17.2} &
  {\small\!92.2} & {\small\!15.2} &
  {\small\!96.6} & {\small\!12.0} &
  {\small\!23.0} &  {\small\!2.3} &
  {\small\!43.0} & {\small\!21.0} &
  {\small\!48.0} & {\small\!15.0} \\
\rule{0pt}{11pt} $\mathrm{h2o}_{8k}^{\text{no}}$ &
  {\small\!94.2} & {\small\!36.4} &
  {\small\!90.8} & {\small\!17.3} &
  {\small\!87.2} & {\small\!13.6} &
  {\small\!91.8} & {\small\!17.8} &
  {\small\!98.0} & {\small\!19.6} &
  {\small\!31.9} &  {\small\!2.7} &
  {\small\!53.0} & {\small\!22.0} &
  {\small\!44.0} &  {\small\!2.0} \\
\rule{0pt}{11pt} $\mathrm{h2o}_{8k}^{\text{or}}$ &
  {\small\!96.2} & {\small\!44.2} &
  {\small\!88.8} & {\small\!65.2} &
  {\small\!85.0} &  {\small\!6.8} &
  {\small\!87.6} & {\small\!59.4} &
  {\small\!95.2} & {\small\!41.7} &
  {\small\!33.3} &  {\small\!3.3} &
  {\small\!49.0} & {\small\!21.0} &
   {\small\!0.0} &  {\small\!1.0} \\
\hline
\end{tabular}
\caption{\label{tab:helmet-results}
  Results for long-context inference on Helmet datasets. In the upper part, we use 4 datasets of sequence lengths 64k, 128k each (cols), run 5 KV cache policies and 2 chunk sizes $2048 = 2k, 1024 = 1k$ (rows). The {\em MatchFirstWordOrPhrase} metric is used with these QA datasets. In the lower part, we use 8 additional datasets and run 6 KV cache policies (chunk sizes 1024, 8192). Different metrics are used, depending on the dataset (see \tabref{dataset-summary-helmet}). The {\em exact} rows are for exact inference (sequence parallelism). Columns {\em us} are for models trained using our novel method with the same cache policy in place, columns {\em sp} are for models trained with sequence parallelism.}
\end{table}

\section{Experiments}\label{sec:experiments}

The key question addressed in our experiments is: {\em if long context inference uses sparse attention with a particular KV cache policy, how much is gained by {\em fine-tuning} the model with the same policy in place (using our novel method) over training it with state of the art libraries using sequence or context parallelism}? We run comparisons on the Helmet~\citep{Yen:25} benchmarks, with context widths of 64k and 128k, covering a range of cache policies:
\begin{itemize}
\item
  $\mathtt{lastrec}$ (lr): $\pi(b, h, t) = t \Ind{t < N_C} + (\mathrm{mod}(t - N_C, N_C - \beta) + \beta) \Ind{t \ge N_C}$, where $\beta\in[0, N_C)$. Keeps the last recent $N_C - \beta$ and first $\beta$ tokens in the cache. It is important to choose $\beta>0$ as a default "attention sink"~\citep{Xiao:24}. Our default is $\beta = \min(16, \lceil N_C / 8 \rceil)$.
\item
  $\mathtt{smart\_lastrec}$ (slr): Variant of $\mathtt{lastrec}$, where the number $\beta$ of initial tokens is chosen dependent on content (see \secref{app-smart-lastrec} for details). A simple version of this heuristic appeared in \citep{Han:24}.
\item
  $\mathtt{h2o}$ ($\mathrm{h2o}$), $\mathtt{h2o\_norm}$ ($\mathrm{h2o}^{\text{no}}$), $\mathtt{h2o\_orig}$ ($\mathrm{h2o}^{\text{or}}$): Variants of H2O~\citep{Zhang:23} (see \secref{main-h2o} for details). $\mathtt{h2o\_orig}$ is equivalent to their code released, where $\pi(b, h, t)$ does not depend on batch position $b$.
\end{itemize}
We train a {\tt Qwen3-4B-Instruct-2507}\footnote{
  This checkpoint has been post-trained to process long contexts. We still need to adjust the RoPE position encoding, which we do with YaRN.}
model~\citep{Yang:25}, using {\tt AdamW}~\citep{Loshchilov:17} with base learning rate $0.0005$ for up to 5 epochs. We train LoRA weights only~\citep{Hu:21} (rank $r=16$, $\alpha=16$, on all linear blocks). We run on four Nvidia A100 40 GB devices with a per-device batch size of 2, using RoPE~\citep{Su:24} and YaRN~\citep{Peng:24} for position encoding. Fine-tuning is done in different ways:
\begin{itemize}
\item
  Sequence parallelism (sp): We use MS-SWIFT~\citep{Zhao:25} for fine-tuning with DeepSpeed ZeRO-3 offload, FlashAttention, and Liger kernels enabled, running on four Nvidia A100 40GB GPUs. We use a per-device batch size of 2 and sequential gradient accumulation (effective batch size 8)\footnote{
    For 64k (128k), 2 (4) devices cover the context, so we need 2 (4) sequential gradient accumulation steps.}.
\item
  Our method (us): We use a cache length $N_c = 32768$, chunk size $S\in\{1024, 2048\}$, and chunks per cell multiplier $\alpha=1$ for $S=2048$, $\alpha=0.75$ for $S=1024$ (see also \secref{app-chunks-cells}). KV cache buffers are quantized to 8 bits using $\mathtt{torchao}$. We run distributed data parallel optimization on 4 devices to obtain an effective batch size of 8. We evaluate the model on a heldout validation set every 10 gradient steps (5 for {\tt pop\_qa}), and choose the checkpoint with the lowest validation loss for testing.
\end{itemize}

\tabref{helmet-results} provides results on 4 Helmet datasets ({\tt nq, trivia\_qa, hotpot\_qa, pop\_qa}; length 64k and 128k)~\citep{Yen:25} and 10 setups (5 policies, 2 chunk sizes) in the upper part, and results for further 8 Helmet datasets ({\tt trec\_coarse, trec\_fine, nlu, banking77, clinc150, inf\_qa, inf\_mc, json\_kv}; length 128k) and 6 setups (3 policies, 2 chunk sizes) in the lower part. Respective results for the base checkpoint (no fine-tuning) are given in \tabref{results-basemodel} in the Appendix. Details about Helmet datasets and metrics are given in \secref{app-data-helmet}.

These results confirm our hypothesis: {\em us} outperforms {\em sp} across the board, with the possible exception of {\tt pop\_qa}. If inference is to be done with sparse attention, it is a poor idea to fine-tune a model with SotA sequence parallelism, even if ample GPU resources are available. A closer look at generated samples reveals a major failure mode of {\em sp} (see \secref{app-error-analysis}): {\em its outputs are far too long and contain mostly random nonsense}. For several Helmet datasets in the lower part, targets are single numerical values, and the {\em Accuracy} metric (see \secref{app-data-helmet}) compares this to the most frequently occurring number in the output. Poor results are due to the output for {\em sp} often containing many numbers. In contrast, {\em us} learns how to stop properly and usually outputs a single number. The same failure mode dominates outcomes on the QA datasets (upper part, {\em MatchFirstWordOrPhrase} metric). {\em sp} fails for {\tt json\_kv} as well, despite this using the more lenient {\em SubEM} metric. Targets are UUIDs of length 32 tokens. While variants of {\em us} identify them about half the time, they are hardly ever contained in the outputs of {\em sp}. In \secref{app-error-analysis}, we quantify the failure mode in \tabref{stats-token-lengths-128k}. While outputs for {\em us} are close in length to true targets, they are longer by large factors for {\em sp}, which often (but not always) span the full 128 tokens (despite true targets being much shorter). We also provide randomly chosen examples for outputs there.

Target conciseness is a clear signal in the training data, expressed also by the {\em sp} checkpoints if exact inference is used with them (see \tabref{stats-token-lengths-exact}). We should not be surprised by {\em sp} failing. When training with sequence parallelism (SP), each token can attend to any earlier one. This property is cut during inference, when most KV information is evicted at some point according to a logic which SP was never aware of. {\em Models should be trained under the same conditions and restrictions which govern inference later on}. Our new method allows practitioners to do that even on a low budget, no matter what KV cache policy they like to use during inference.

Results in \tabref{helmet-results} (lower) indicate that accuracy does not significantly differ between chunk sizes 1024 and 8192, despite the latter requiring 25\% less training time (see \tabref{time-train-128k}). Results for chunk size 128 (\tabref{results-ablations}) are not significantly better either, despite training running much longer. At least on the benchmarks considered here, chunk size does not play a major role and should be chosen as large as possible before accuracy degrades. In terms of differences between cache policies, H2O variants outperform $\mathtt{lastrec}$ and $\mathtt{smart\_lastrec}$, while $\mathtt{h2o}$ tends to do worse than the variants $\mathtt{h2o\_norm}$, $\mathtt{h2o\_orig}$. Despite its independence from batch position, $\mathtt{h2o\_orig}$ does as well as $\mathtt{h2o\_norm}$, except it fails in a manner similar to {\em sp} on {\tt json\_kv} (see also \tabref{stats-token-lengths-128k}).

{\bf Choice of Metric for Helmet QA Datasets}. Results of {\em sp} on the Helmet QA datasets (\tabref{helmet-results}, upper part) are sensitive to the metric being used. In \secref{app-metrics-helmet-qa}, we provide results for {\em MatchFirstWordOrPhrase}, {\em SubEM}, and 6 other metrics. For {\em SubEM}, {\em Prefix} (not realizable), and {\em TypedFirst} (realizable, but complex), differences between {\em us} and {\em sp} are not significant, as they seem to tolerate the failure modes we identified. However, {\em sp} fails for Helmet {\tt json\_kv} and the RULER datasets, despite {\em SubEM} being used for them.

\setlength{\tabcolsep}{4.25pt}
\begin{table}[ht!]
\begin{subtable}[c]{0.5\textwidth}
\centering
\begin{tabular}{|l|rr|rr|rr|rr|rr|}
\hline
 & \multicolumn{2}{c|}{niah\_s\_1} & \multicolumn{2}{c|}{vt} & \multicolumn{2}{c|}{cwe} & \multicolumn{2}{c|}{fwe} \\
 & us & sp & us & sp & us & sp & us & sp \\
\hline\hline
\rule{0pt}{13pt} $\mathrm{exact}$ &
  - & {\small\!91.0} &
  - & - &
  - & {\small\!70.2} &
  - & {\small\!87.0} \\
\hline
\rule{0pt}{11pt} $\mathrm{slr}_{8k}$ &
  {\small\!13.0} & {\small\!13.0} &
  {\small\!33.2} & - &
  {\small\!94.1} & {\small\!14.0} &
   {\small\!100} & {\small\!85.0} \\
\rule{0pt}{11pt} $\mathrm{h2o}_{8k}^{\text{no}}$ &
  {\small\!96.0} & {\small\!31.0} &
  {\small\!99.4} & - &
   {\small\!100} & {\small\!24.2} &
   {\small\!100} & {\small\!89.3} \\
\rule{0pt}{11pt} $\mathrm{h2o}_{8k}^{\text{or}}$ &
  {\small\!43.0} &  {\small\!7.0} &
  {\small\!79.6} & - &
  {\small\!83.4} & {\small\!25.8} &
  {\small\!95.3} & {\small\!80.7} \\
\hline
\end{tabular}
\end{subtable}
\begin{subtable}[c]{0.5\textwidth}
\centering
\begin{tabular}{|l|r|r|r|}
\hline
 & trec\_f & inf\_qa & inf\_mc \\
\hline\hline
\rule{0pt}{13pt} exact &
  {\small\!85.4} & {\small\!44.2} & {\small\!64.0} \\
\rule{0pt}{11pt} us, $\mathrm{h2o}_{1k}^{\text{no}}$ &
  {\small\!90.6} & {\small\!32.1} & {\small\!53.0} \\
\rule{0pt}{11pt} OOMB &
  {\small\!88.3} & {\small\!35.0} & {\small\!36.0} \\
\rule{0pt}{11pt} bs, $\mathrm{h2o}_{1k}^{\text{no}}$ &
  {\small\!26.0} &  {\small\!2.9} & {\small\!41.0} \\
\rule{0pt}{11pt} OOMB(bs) &
  {\small\!22.8} &  {\small\!2.8} & {\small\!43.0} \\
\hline
\end{tabular}
\end{subtable}
\caption{\label{tab:ruler-oomb-results}
  {\bf Left}:
  Results for long-context inference on RULER datasets (256k sequence lengths). The {\em StringMatchAll} metric is used with these datasets. We compare exact inference with 3 KV cache policies, using chunk size 8192 = 8k. Columns {\em us} are for models trained using our novel method with the same cache policy in place, columns {\em sp} are for models trained with sequence parallelism. \\
  {\bf Right}:
  Results for long-context inference on 3 Helmet datasets (128k sequence lengths). We compare {\em exact} (dense attention training and inference), {\em us} {\tt h2o\_norm} (our method), OOMB~\citep{Li:26}, {\em bs} {\tt h2o\_norm} (our method, no fine-tuning), and OOMB(bs) (no fine-tuning).
}
\end{table}

{\bf Results for RULER(256k) Datasets}. We also run comparisons on the RULER~\citep{Hsieh:24} benchmarks, with context width 256k (see details in \secref{app-data-ruler}. Results for 4 datasets and 3 policies (chunk size 8192) are given in \tabref{ruler-oomb-results} (left). The StringMatchAll metric used here is closely related to SubEM and just as non-realizable. Despite the RULER tasks being solved to high accuracy by {\em us} (for the $\mathtt{h2o\_norm}$ policy), results for {\em sp} are much worse across the board. Interestingly, {\em us} does significantly better than {\em exact} on the {\tt cwe} and {\tt fwe} tasks. Note that SP training for {\em exact} and {\em sp} requires eight GPUs to cover the 256k sized context, while our method runs on a single GPU as above (using 4 GPUs to attain batch size 8 with DDP). For RULER {\tt niah\_s\_1} and {\tt vt}, we see a clear difference in accuracy between different KV cache policies: $\mathtt{h2o\_norm}$ does best by far, followed by $\mathtt{h2o\_orig}$, and $\mathtt{smart\_lastrec}$ performs worst. In order for a KV cache 8 times smaller than the full range to result in the same performance as exact attention, it does matter how slots are selected for eviction.

{\bf Results for Comparison to OOMB}. We also run comparisons to OOMB~\citep{Li:26}, a losely related long context fine-tuning method (\secref{relwork}, \secref{app-relwork}). Details about the comparative setup are given in \secref{app-compare-oomb}. Results are given in \tabref{ruler-oomb-results} (right). For {\tt inf\_mc}, OOMB is much worse than our method ({\em us}) and even performs worse than OOMB(bs), which does not run fine-tuning. For {\tt inf\_qa} and {\tt trec\_fine}, their results are comparable to ours. On the same hardware setup and batch size, their code runs about three times longer than ours. As detailed in \secref{app-relwork}, OOMB requires complex and intransparent hardcoding for any new KV cache policy, SDPA kernel, or other external dependency, while our method is entirely agnostic to all these choices, and our open source library imports SotA dependencies instead of rewriting them.

%% file: longcon_conclusions.tex
\section{Conclusions}\label{sec:conclusions}

We showed how transformer language models with sparse attention can be fine-tuned on a moderate hardware budget (e.g., a single Nvidia A100 GPU with 40 GB RAM). Our method works for any KV cache selection or compression policy and allows the model to co-adapt with the policy, often outperforming models trained with exact attention (sequence parallelism). We also provide a much more efficient implementation of H2O sparse attention (the leading policy in our experiments) with dedicated scaled dot product attention (SDPA) kernel support. By simplifying KV cache structure and clarifying the requirements on SDPA, we hope to draw the fast inference community’s attention to sparse attention (see \secref{app-sdpa-kernels}), which despite its major potential for post-training specialization via cache selection or compression policy design does not currently play a significant role in long-context inference or fine-tuning practice.

In future work, we will combine context parallelism with sparse attention. We are also considering kernel fusion ideas in order to narrow the latency gap further. An important direction will be multi-stream asynchronous implementations which allow for on-the-fly CPU offloading~\citep{Yuan:26}. We believe that once the host memory of a system can be used without much synchronization overhead and less double buffering, many current difficulties with KV caching will be much diminished. \ifthenelse{\equal{\isarxiv}{no}}{Finally, we hope that the open source library to be released with this publication will make it easier to use, compare and extend sparse attention policies, work on which so far is somewhat cluttered when it comes to implementations.}{Finally, we hope that $\mathtt{KeysAndValues}$ (\url{https://github.com/awslabs/keys_values}), the open source library with which most experiments were run here (see \secref{app-os-library}), will make it easier to use, compare and extend sparse attention policies, work on which so far is somewhat cluttered when it comes to implementations.}

%% file: longcon_appendix.tex
\section{Appendix}\label{sec:appendix}

\ifthenelse{\equal{\isarxiv}{no}}{
\subsection{AI Use Statement}

In this work, we used generative AI tools for writing boilerplate code for experimentation (data preprocessing and loading, scripts to collect results and create result tables), debugging and profiling code (with mixed success), cleaning and reformatting datasets, supporting qualitative and thematic data analysis. We have not used generative AI tools for proposing or refining hypotheses, designing or providing feedback on research methodology or experiments, assisting with translation, interpreting results, suggesting experimental parameters, creation of artifacts, drafting parts of a research paper, transcribing recordings of research material, discovering research topics or identifying gaps, brainstorming, editing a research paper to improve readability, formatting references, suggesting a structure for a research paper, or proposing a title or keywords for a research paper. Moreover, generating synthetic data sets, helping to develop theoretical models or conceptual frameworks, formulating mathematical claims, providing critical ingredients for proving mathematical claims, or assisting in the writing of proofs are not applicable to this work (and we would not use AI for them anyway, except the first).

Additionally, we used generative AI tools for creating a figure and identifying relevant literature. We have reviewed all AI-assisted work (mostly code). We take responsibility for the final content of this work, including text, claims or artifacts produced with the aid of generative AI.

\subsection{Reproducibility Statement}

Experiments done for this paper are in principle fully reproducible. The code we used is either already open source, or will be open sourced along with this publication (it is held back for reasons of author anonymity only). An anonymous snapshot of our code can be accessed here: \url{https://anonymous.4open.science/r/review-artifact-2027-5850/README.md}.

The reader interested in reproduction will need at least one AWS P4 instance (or equivalent setup, containing 8 Nvidia A100 GPUs with 40 GB of RAM) as well as some patience, since our experiments require fine-tuning a 4B parameters checkpoint under hundreds of different conditions, on a range of different datasets with token lengths between 64k and 256k. All scripts we used for collecting results, creating tables, or processing data, are included with the code. Experimental protocols can be obtained on request. We are not authorized to publish the fine-tuned checkpoints.
}{}

\subsection{SDPA Kernels for Sparse Attention}\label{sec:app-sdpa-kernels}

Here, we list some ideas for SDPA kernel developers to better support sparse attention. First, we consider causal masking for sparse attention. In standard MHA (the "training case"), $(\mathtt{mask}(\mxx{}))_{i, j} = x_{i, j} - \infty \Ind{i < j}$. For sparse attention, KV information is stored in cache buffers in an ordering given by token positions $t(b, h, j)$, and the causal mask is given by $(b, h, i, j)\mapsto (-\infty) \Ind{P + i < t(b, h, j)}$, where $P$ is the number of tokens processed before the current MHA call (so the new information is for tokens $\srng[P]{P + N_q - 1}$). The new key-value information has been written into the cache already, so that $\srng[P]{P + N_q - 1}$ is part of $\{t(b, h, j)\}$ for each $(b, h)$.

Unfortunately, none of the fast SDPA kernel implementations we know of support such variants of causal masking in an implicitly\footnote{
  We cannot pass an explicit mask matrix, which would be huge and defy the purpose of fast SDPA.}
defined way. In our implementation, we sort the token positions and reorder keys and values according to this index, separate for each $(b, h)$, after which we can use standard causal masking, where queries are right-aligned with keys and values. This needs extra computation and memory which could be saved with better SDPA kernel support. Based on our experience, the following simple extensions of fast SDPA kernel libraries could make a major difference for sparse attention:
\begin{itemize}
\item
  Return summed attention weights $\sum_i m_{b, h, i, j}$ (see \secref{main-h2o}), an array of size $(B, H_q, N_k)$, based on attention weights which are computed anyway. This allows for H2O and related scores to be computed, driving advanced KV cache policies. It should be a straightforward add-on, see \secref{app-attn-weights}.
\item
  Allow for implicitly defined causal masks of the form $(b, h, i, j)\mapsto (-\infty) \Ind{P + i < t(b, h, j)}$, where $t(\cdot)$ is an $(B, H_k, N_k)$ integer array as additional input to the operator. While FlexAttention~\citep{Dong:25} supports custom mask patterns, they need to be written in terms of scalar index variables, so cannot have a 3D array as input to the compute graph.
\end{itemize}

\subsection{Notation. Definitions}

Here, we add some details missing in the main text.

\subsubsection{Buffer Read/Write Access by $\mathtt{torch.scatter}, \mathtt{torch.gather}$}\label{sec:app-scatter-gather}

These linear operations are defined in \url{https://docs.pytorch.org/docs/2.12/generated/torch.Tensor.scatter_.html}. $\mathtt{scatter\_}$ assigns values to entries in certain positions, $\mathtt{gather}$ extracts values of entries at certain positions. For reference, $\mathtt{arr.scatter\_(dim, index, src)}$ requires $\mathtt{arr.ndim == index.ndim == src.ndim}$ and $\mathtt{index.shape == src.shape}$, whereas $\mathtt{arr.shape}$ can differ on position $\mathtt{dim}$. Say that $\mathtt{arr.ndim == 3}$. Then:
\[
  \begin{split}
    \mathtt{arr[index[i, j, k], j, k] = src[i, j, k]}\quad & |\; dim == 0 \\
    \mathtt{arr[i, index[i, j, k], k] = src[i, j, k]}\quad & |\; dim == 1 \\
    \mathtt{arr[i, j, index[i, j, k]] = src[i, j, k]}\quad & |\; dim == 2 \\
  \end{split}
\]
Also, if $\mathtt{res = arr.gather(dim, index)}$, then:
\[
  \begin{split}
    \mathtt{res[i, j, k] = arr[index[i, j, k], j, k]}\quad & |\; dim == 0 \\
    \mathtt{res[i, j, k] = arr[i, index[i, j, k], k]}\quad & |\; dim == 1 \\
    \mathtt{res[i, j, k] = arr[i, j, index[i, j, k]]}\quad & |\; dim == 2 \\
  \end{split}
\]
In our use case, we apply these operations to 4D arrays with $\mathtt{dim = 2}$, so that $\mathtt{arr, src}$ are 4D, but $\mathtt{index}$ is 3D. This is done by broadcasting $\mathtt{index}$ along the final axis: $\mathtt{index.unsqueeze(-1).extend(-1, -1, -1, d_h)}$.

\subsection{Key-Value Cache Policies}

In this section, we present additional details about KV cache policies used in our experiments.

\subsubsection{Policy $\mathtt{smart\_lastrec}$ (slr)}
\label{sec:app-smart-lastrec}

Recall the simple $\mathtt{lastrec}$ (lr) policy from \secref{experiments}, which keeps the last recent $N_C - \beta$ and first $\beta$ tokens in the cache. One drawback of this policy is that $\beta$ is fixed, while prompts often start with initial important control information of {\em variable} length. Our $\mathtt{smart\_lastrec}$ policy is defined in terms of a regular expression for the end of this task control prefix, as well as some maximum prefix length $M_{\text{prefix}} < N_C$. When processing the first (prefill) chunk, we search for the first match (separately for each batch position $b$). If this results in a prefix length $M(b)\le M_{\text{prefix}}$, this is used, otherwise $M(b) = M_{\text{prefix}}$. For subsequent chunks and token positions $t\ge N_C$, the policy is $\pi(b, h, t) = M(b) + \mathrm{mod}(t - N_C, N_C - M(b))$.

We also implemented a generalization where a {\em range} $[M_0(b), M_1(b))$ is protected from eviction.\footnote{
  We allow for $M_1(b) < M_0(b)$, in which case the protected area is $[0, M_1(b))\cup [M_0(b), N_C)$.
}
Here, $M_0(b)$ is chosen as the position of the first non-padding token, and $M_1(b)$ is chosen as above. The idea is that initial padding tokens do not carry information and should not be attended to, so they can be evicted as soon as the cache is full. In this variant, if all tokens in the prefill chunk are padding for some $b$, the search for $[M_0(b), M_1(b))$ is shifted to subsequent chunks. The prefix case above is obtained with $M_0(b) = 0$. Surprisingly, in our experiments, the general variant did not improve over the prefix variant,\footnote{
  This needs further analysis. Maybe some initial padding tokens are used as so-called "attention sinks"~\citep{Xiao:24}.}
so that $\mathtt{smart\_lastrec}$ in \secref{experiments} is the prefix variant throughout.

\subsection{Long-Context Benchmarks}

\subsubsection{Helmet}\label{sec:app-data-helmet}

Our training and evaluation suite is derived from Helmet~\citep{Yen:25}, a benchmark designed for inference-time evaluation of long-context language models. Helmet covers five capability categories across five context-length scales (8k to 128k tokens). Helmet provides only a small number of instances per task. For supervised fine-tuning, we adapted it as follows.

\paragraph{Instance Construction and Task Scope.} We reconstruct each task from its upstream source data, following the original Helmet logic for forming contexts and controlling sequence length. We focus on the 64k and 128k context-length settings and include 12 tasks spanning five capability categories. Table~\ref{tab:dataset-summary-helmet} provides a summary of each task. 

\paragraph{Split Separation.} For each task we produce two non-overlapping partitions. The instances used in the original Helmet evaluation are reserved as a held-out \emph{evaluation (test) set}. All remaining instances are collected into a \emph{development set} used for training. For RAG tasks, we sample a single depth (difficulty) variant per query in the development set to prevent the model from memorising the same question paired with multiple distractor configurations. For InfiniteBench QA/MC, we remove in-context demonstrations from development instances because the demonstrations are drawn from the same small pool and would otherwise create data leakage during training.

\paragraph{Evaluation Metrics.} Our evaluation metrics are mostly taken from the code coming with Helmet (\url{https://github.com/princeton-nlp/HELMET/blob/main/utils.py}). Here, the output is a string generated by the model, the target is a string or a list of strings:
\begin{itemize}
\item
  MatchFirstWordOrPhrase: The {\tt target} is a list of strings. Strings (both {\tt output} and {\tt target} entries) are normalized by stripping off whitespace and converting letters to lower-case. We build the set {\tt extracts} by appending the first word in {\tt output} if at least one {\tt target} entry is a single word, and appending the first sentence in {\tt output} if at least one {\tt target} entry consists of more than one word. The metric is 1 if $\mathtt{t} = \mathtt{e}$ for some $\mathtt{t}\in\mathtt{target}$, $\mathtt{e}\in\mathtt{extracts}$, and 0 otherwise.
\item
  Accuracy: The target is a numerical value. We extract all numerical values from the output and find the value which occurs most often (with ties, the value is chosen which appears first). We then use exact match between this value and the target.
\item
  SubEM: Depending on the dataset, the target can be a list of strings. We normalize the output (strip whitespace, quotes, and common phrases such as ``Answer:''), map output and target(s) to lower-case. The value is 1 if at least one of the targets is a substring in the output, 0 otherwise.
\item
  ROUGE-F1: The target is a string. We compute ROUGE-N precision/recall/F1 between output (after normalization as in SubEM) and target.
\end{itemize}

Here, {\em MatchFirstWordOrPhrase} is our choice and replaces {\em SubEM} as Helmet choice for {\tt nq}, {\tt trivia\_qa}, {\tt pop\_qa}, {\tt hotpot\_qa}. This replacement is motivated in \secref{app-error-analysis} and \secref{app-metrics-helmet-qa}. In essence, {\em SubEM} is not actionable, in that it does not prescribe a way to extract the desired substring from the output. It is also easy to game by simply generating many tokens.

\begin{table}[t]
\centering
\small
\begin{tabular}{llllrr}
\hline
\textbf{Category} & \textbf{ID} & \textbf{Source} & \textbf{Metric} & \textbf{Dev} & \textbf{Eval} \\
\hline\hline
\multirow{4}{*}{RAG}
  & \texttt{nq}         & Natural Questions & MatchFirstWordOrPhrase & 893 & 600 \\
  & \texttt{trivia\_qa} & TriviaQA          & MatchFirstWordOrPhrase & 876 & 600 \\
  & \texttt{pop\_qa}    & PopQA             & MatchFirstWordOrPhrase & 192 & 600 \\
  & \texttt{hotpot\_qa} & HotpotQA          & MatchFirstWordOrPhrase & 787 & 300 \\
\hline
\multirow{3}{*}{Many-shot ICL}
  & \texttt{trec\_coarse} & TREC             & Accuracy & 1000 & 500 \\
  & \texttt{trec\_fine}   & TREC             & Accuracy & 1000 & 500 \\
  & \texttt{nlu}          & SNIPS NLU        & Accuracy & 2094 & 500 \\
  & \texttt{banking77}    & BANKING77        & Accuracy & 2580 & 500 \\
  & \texttt{clinc150}     & CLINC150         & Accuracy & 2600 & 500 \\
\hline
\multirow{2}{*}{Long-doc QA}
  & \texttt{inf\_qa}      & InfiniteBench QA & ROUGE-F1 & 251 & 100 \\
  & \texttt{inf\_mc}      & InfiniteBench MC & Accuracy & 129 & 100 \\
\hline
  Synthetic Recall
  & \texttt{json\_kv}     & JSON-KV          & SubEM    & 500 & 100 \\
\hline
\end{tabular}
\caption{\label{tab:dataset-summary-helmet}
  Overview of the 12 Helmet tasks. \emph{Dev} and \emph{Eval} denote the number of instances in the training and evaluation partitions, respectively, at a single context-length setting.}
\end{table}

In what follows, we provide details for the different tasks.

\paragraph{Retrieval-augmented Generation (RAG).}
Each instance consists of a natural-language question, one or more gold passages, and a pool of hard-negative distractors. The context is formed by inserting gold passages at a random depth among the distractors, and the whole context is truncated to the target length. \textit{Natural Questions} (NQ) uses real Google search queries paired with Wikipedia passages. \textit{TriviaQA} uses trivia questions authored with independently collected evidence. \textit{PopQA} focuses on long-tail, entity-centric questions generated from Wikidata triples; we filter the evaluation set to queries whose subject entities fall below a popularity threshold of~3. \textit{HotpotQA} requires multi-hop reasoning across two gold passages. Each of the six depth variants of a query places the gold passage(s) at a different relative position in the distractor pool.


\paragraph{Many-shot In-context Learning.}
Three intent/question-type classification datasets test the model's ability to exploit many labeled demonstrations placed entirely within the context window. Unlike most other tasks, where demonstrations serve only as formatting guides, here they carry essential semantic information: each demonstration encodes a (text, ordinal-label) pair, and the label-to-class mapping is only recoverable by reading the demonstrations. \textit{TREC Coarse} has 6 question-type classes; \textit{TREC Fine} uses the same question-classification data but increases the label granularity to 50 fine-grained question types; \textit{SNIPS NLU} has 68 intent classes; \textit{BANKING77} contains 77 fine-grained intents covering online-banking queries; \textit{CLINC150} has 151 intent classes. The number of demonstrations is calibrated to fill the context window while maintaining an approximately balanced class distribution across shots.

\paragraph{Long-document QA.}
Both \textit{InfiniteBench QA} (Inf-QA) and \textit{InfiniteBench MC} (Inf-MC) are derived from full-length novels whose named entities have been replaced by synthetic ones to prevent answer memorisation. The source document typically exceeds the target context window, so it is truncated to fit. Inf-QA is open-ended, evaluated by ROUGE-F1; Inf-MC is a 4-way multiple-choice variant evaluated by accuracy. 

\paragraph{Synthetic Recall.}
\textit{JSON-KV} asks the model to retrieve the value associated with a specified key from a large JSON dictionary that fills the context window. This task is evaluated with SubEM and serves as controlled probes of the model's ability to locate and copy specific information over very long spans.

\subsubsection{RULER}\label{sec:app-data-ruler}

Our training and evaluation suite is derived from RULER~\citep{Hsieh:24}, see also \url{https://github.com/NVIDIA/RULER}. As with Helmet (\secref{app-data-helmet}), the original datasets are too small for fine-tuning purposes, and we adopt their data generation code in order to sample more cases. Sequences are approximately 256k tokens long, which is twice what we did for Helmet.

RULER is a generator of synthetic long-context tasks: a short question is asked about a controlled pattern hidden in a long context. We include those four tasks of RULER which do not need additional large corpora to be downloaded, see \tabref{dataset-summary-ruler} for details.

\begin{table}[t]
\centering
\small
\begin{tabular}{lllrrr}
\hline
\textbf{Category} & \textbf{ID} & \textbf{Metric} & \textbf{Gen.\ Budget} & \textbf{Dev} & \textbf{Eval} \\
\hline\hline
  Retrieval	& \texttt{niah\_single\_1} & StringMatchAll & 128 & 400 & 100 \\
  Multi-hop tracing & \texttt{vt} & StringMatchAll & 30 & 400 & 100 \\
  Aggregation & \texttt{cwe} & StringMatchAll & 120 & 400 & 100 \\
  Aggregation & \texttt{fwe} & StringMatchAll & 50 & 400 & 100 \\
\hline
\end{tabular}
\caption{\label{tab:dataset-summary-ruler}
  Overview of the 4 RULER tasks. \emph{Dev} and \emph{Eval} denote the number of instances in the training and evaluation partitions, respectively.}
\end{table}

We adopt the same following general procedure:
\begin{itemize}
\item
  {\bf Instance construction}: Instances are generated by the original RULER code (\texttt{scripts/data/prepare.py}), pinned to a fixed commit of the RULER repository. We use the task arguments of \texttt{scripts/synthetic.yaml}, no chat template (base), and random seed 42. RULER measures lengths with the \texttt{cl100k\_base} tokenizer of \texttt{tiktoken}, and the target length includes the generation budget of the task.
\item
  {\bf Unified format}: Each instance is a JSON object with fields \texttt{["input", "output", "outputs", "query\_id", "max\_length"]}. RULER splits the prompt into \texttt{input} and \texttt{answer\_prefix}, we join them again. \texttt{outputs} is the list of reference strings, all of which have to be contained in the answer. \texttt{output} is the target used for training, obtained by joining outputs (see below).
\item
  {\bf Separation of splits}: The first 100 instances form the evaluation partition, the remaining ones the development partition.
\end{itemize}

The default metric for these four RULER tasks is {\em StringMatchAll}: targets for a case are a list of reference strings, and the metric counts the fraction of target strings contained in the generated output (ignoring case and stripping out whitespace). Similar to {\em SubEM} for Helmet, this metric is not realizable at inference time (as it depends on the target entries). However, since RULER tasks prescribe the desired output format narrowly, this is not critical (the same holds for Helmet {\em JSON-KV}, where {\em SubEM} is in fact suitable).

\paragraph{Retrieval ({\tt niah\_single\_1})}

A single needle "One of the special magic numbers for \{KEY\} is: \{VALUE\}." is inserted at a random position into a haystack made of repetitions of "The grass is green. The sky is blue. The sun is yellow. Here we go. There and back again.". The key is a random adjective-noun pair, the value a random 7-digit number, which is the answer.

{\tt\small
A special magic number is hidden within the following text. Make sure to memorize it. I will quiz you about the number afterwards. \\
\{CONTEXT\} \\
What is the special magic number for \{QUERY\} mentioned in the provided text? The special magic number for \{QUERY\} mentioned in the provided text is \\
}

Example target: 5663623.

\paragraph{Multi-hop tracing ({\tt vt})}

A chain of assignments \texttt{VAR \{X1\} = \{VALUE\}", "VAR \{X2\} = VAR \{X1\}", ...} (four hops) is inserted into the same noise haystack. The answer consists of the five variables which are assigned the value.

{\tt\small
Memorize and track the chain(s) of variable assignment hidden in the following text. \\
\{CONTEXT\} \\
Question: Find all variables that are assigned the value \{QUERY\} in the text above. \\
Answer: According to the chain(s) of variable assignment in the text above, \{NUM\_V\} variables are assigned the value \{QUERY\}, they are: \\
}

Example target: \texttt{LTVHU MGLSG IXCDJ DFKIM BVKUY}.

\paragraph{Aggregation ({\tt cwe})}

The context is a numbered list of words, where 10 words appear 30 times each, and the other words appear 3 times each. The answer consists of the 10 common words.

{\tt\small
Below is a numbered list of words. In these words, some appear more often than others. Memorize the ones that appear most often. \\
\{CONTEXT\} \\
Question: What are the 10 most common words in the above list? \\
Answer: The top 10 words that appear most often in the list are: \\
}

Example target: \texttt{1. gold 2. audited 3. abuse 4. stretch 5. absorbed 6. runway 7. drudgery 8. measurement 9. compress 10. squeamish}.

\paragraph{Aggregation ({\tt fwe})}

The context is a sequence of random coded words, whose frequencies follow a Zipf distribution with exponent 2. The most frequent word is replaced by "...", and the answer consists of the next 3 most frequent words.

{\tt\small
Read the following coded text and track the frequency of each coded word. Find the three most frequently appeared coded words. \\
\{CONTEXT\} \\
Question: Do not provide any explanation. Please ignore the dots '....'. What are the three most frequently appeared words in the above coded text? \\
Answer: According to the coded text above, the three most frequently appeared words are: \\
}

Example target: \texttt{nwlwst rspvov hmcxdl}.

\subsection{Gradient Computation}

In this section, we provide additional details about our long-context gradient computation method from \secref{main-finetune}.

\subsubsection{Chunks, Cells, CPU Offloading}\label{sec:app-chunks-cells}

Recall that for long-context inference or fine-tuning with cache length $N_C$, we split a sequence of length $N>N_C$ into $1 + \lceil (N - N_C)/S \rceil$ {\em chunks}, the first (prefill) chunk of length $N_C$, subsequent chunks of length $S < N_C$. The chunk size $S$ is chosen according to a latency-vs-accuracy trade-off, it is in general much shorter than $N_C$ (\secref{main-satt-vs-cp}). We also group chunks into {\em cells}. The first cell consists of the prefill chunk alone, subsequent cells group $k = \lfloor \alpha N_C / S \rfloor$ of $S$-length chunks, where $\alpha>0$ is a hyperparameter which defaults to $\alpha = 1$.

In our gradient computation method, {\em PyTorch autograd} is run on cells. While $S$ is chosen also with accuracy in mind, the choice of $k$ and $\alpha$ is determined by efficiency (both runtime and memory) only. The idea is that the {\em autograd} GPU memory requirements are on the order of one KV cache buffer, if we exploit the linear recurrence (see \secref{main-finetune} and below). If a cell was much larger, the {\em delta} nodes in the computation graph would dominate. The $\alpha$ parameter is adjusted so to not run out of memory. Empirically, a smaller chunk size $S$ necessitates a smaller $\alpha$, likely due to overhead in {\em autograd} (for constant $\alpha$, smaller $S$ means larger $k$, so a larger computation graph). In our experiments, we chose $\alpha = 1$ for $S = 2048$ and $S = 8192$, $\alpha = 0.75$ for $S = 1024$, and $\alpha = 0.1$ for $S = 128$, for a cache length of $N_C = 32768$ and 40 GB of GPU memory.

The grouping of chunks into cells is also relevant for long-context {\em inference}. Recall that $L$ denotes the number of layers of our model, and layer inputs are of shape $(B, N, d)$, where $d = H_q\cdot d_h$ is the model embedding dimension. For large $N$, not even the inputs to one layer can be kept in GPU memory. This has implications for how inference is computed and what can be offloaded to CPU memory at which point during the process (our library supports CPU offloading of KV cache buffers during the forward passes, as well as CPU offloading of weights during the backward pass, but this is not used in the experiments reported here).

{\bf Outer loop over chunks, inner loop over layers}: This seems the simplest ordering. Also, layer inputs can be kept in GPU memory, with outputs overwriting inputs. A major drawback is that either all KV cache buffers need to be kept in GPU memory, or they need to be read from and written back to CPU memory frequently. When KV cache buffers are stored in quantized form, the quantization computation can be considerable. For long-context inference, this ordering is not suitable.

{\bf Outer loop over layers, inner loop over chunks}: With this ordering, KV cache buffers (and even model weights) can be offloaded to CPU for all but the currently active layer. A drawback of this ordering is that layer inputs and outputs cannot be kept in GPU memory in total, so need to be offloaded eventually. Still, this ordering is much better than the previous one.

{\bf Outer loop over cells, middle loop over layers, inner loop over chunks per cell}: This ordering provides a good compromise between the two previous ones. Layer inputs and outputs can be kept in GPU memory, while KV cache buffers can be quantized and/or offloaded to CPU, which happens much less frequently. We use this ordering in our implementation, also because CPU offloading of KV cache buffers is needed during gradient computation, so we can just reuse this code.

\subsubsection{Summary of Method}\label{sec:app-grad-summary}

Here, we present a detailed summary of our gradient computation technique, see also \figref{method-diagram}. Recall that we process a batch of $B$ sequences of token length $N$, and that caches in each layer have length $N_C$. For $N\le N_C$, our method reduces to standard training, so assume that $N > N_C$. For simplicity, we assume that caches in all layers have the same length $N_C$. This is easy to relax, and our implementation does so.

At the top level, our method runs a forward pass followed by a backward pass, just like standard code. However, we run {\em autograd} on cells only, which constitute small parts of the overall model graph. As with activation checkpointing~\citep{Herrmann:19}, this means we need to run forward passes over the model three times (instead of just once). The first two passes are run in non-autograd mode, checkpointing (so called) boundary information to CPU memory. The third pass is part of the autograd runs on cells, which consume boundary information as inputs.

We need some notation. Let $\mathcal{L}$ denote the training loss for the current batch. Layers are indexed by $l=\rng[0]{L-1}$, cells by $c=\rng[0]{N_{\text{cells}} - 1}$. $\mxx{l}$ are inputs to layer $l$, of shape $(B, N, d)$, and $\mxx{L}$ is the top layer output. Moreover, $\mxx{l, c}$ denotes the slice of $\mxx{l}$ along axis 1 corresponding to the cell. Finally, let $\mxk{l, c}$ denote the KV cache buffers\footnote{
  These are really two tensors (keys, values), but we concatenate them into one for notational simplicity.}
of shape $(2, B, H_k, N_C, d_h)$ at the input of cell $c>0$ in layer $l$.

{\bf Forward pass 1}. This runs in non-autograd mode, using the ordering cells, then layers, then chunks detailed in \secref{app-chunks-cells}. Alongside:
\begin{itemize}
\item
  Store KV cache replay log in each layer, containing the decisions $\{\pi(b, h, t)\}$.
\item
  Checkpoint layer inputs $\mxx{l}$ to CPU memory for each layer $l=\rng[0]{L-1}$. Our implementation allows to quantize them in order to save CPU memory and CPU-GPU transfer time, but this is not activated in our experiments, since the time is subdominant (and quantization implies errors). We also checkpoint top layer outputs $\mxx{L}$.
\end{itemize}

{\bf Backward pass}. This runs backwards over layers. In each step, we compute gradients for weights in layer $l$ using activation checkpointing. We start with computing head gradients $\partial\mathcal{L} / \partial\mxx{L}$ based on top layer outputs $\mxx{L}$, writing them to CPU (in fact, head gradients $\partial\mathcal{L} / \partial\mxx{l}$ overwrite $\mxx{l}$ on CPU). Next, we iterate over layers $l=\rng[L-1]{0}$:
\begin{itemize}
\item
  Forward pass 2 for layer $l$ (non-autograd mode). Runs over cells $c=\rng{N_{\text{cells}} - 1}$, computing the cache buffers $\mxk{l, c}$ and storing them to CPU. These are quantized in order to save CPU memory and CPU-GPU transfer time, and we overwrite the cache buffer checkpoints from previous layer $l+1$. Cache decisions are replayed from the log. Note that we could checkpoint {\em all} $\mxk{l, c}$ during forward pass 1, but this would require $L$ times more CPU memory, and the extra time for forward pass 2 is subdominant. Also note that we load inputs $\mxx{l, c}$ to forward pass 2 to GPU cell by cell: the whole $\mxx{l}$ does not fit in GPU memory (see \secref{app-chunks-cells}).
\item
  Run {\em autograd} on each cell, iterating from right to left, $c = \rng[N_{\text{cells}} - 1]{0}$. For each cell $c$, we load layer inputs $\mxx{l, c}$ (bottom), layer head gradients $\partial\mathcal{L} / \partial\mxx{l+1, c}$ (top) and incoming cache buffers $\mxk{l, c}$ (left; only for $c > 0$) from CPU, while cache buffer head gradients $\partial\mathcal{L} / \partial\mxk{l, c+1}$ (right; only for $c < N_{\text{cells}} - 1$) are kept in GPU memory. Cache decisions are replayed from the log. {\em autograd} works as follows:
  \begin{itemize}
  \item
    Forward pass 3 for cell $(l, c)$, in autograd mode. Whenever a KV cache buffer node ($\mathtt{keys}$, $\mathtt{values}$) is created, we store an {\em annotation} in a list. In the $\mathtt{pack\_hook}$ function, we match arguments of the right shape against all annotation. For a match, we replace the argument with its {\em delta} encoding, which is stored in the computation graph instead. See \secref{app-exploit-recurrence} for details.
  \item
    Backward: When {\em PyTorch} traverses the computation graph in reverse order, it calls $\mathtt{unpack\_hook}$ for each node. For each {\em delta} encoding, we play the recurrence \eqp{recurrence-reverse} backwards, replacing $\mathtt{keys}'$ by $\mathtt{keys}$ or $\mathtt{values}'$ by $\mathtt{values}$.
  \end{itemize}
  The outcome of {\em autograd} on cell $(l, c)$ are gradients w.r.t.\ layer weights (which are accumulated), a head gradient $\partial\mathcal{L} / \partial\mxx{l, c}$ (written to CPU, overwriting $\mxx{l, c}$), and a head gradient $\partial\mathcal{L} / \partial\mxk{l, c}$ which replaces the previous one (for $c>0$). At the end of the loop over cells, gradients w.r.t.\ layer weights are complete, and the head gradient $\partial\mathcal{L} / \partial\mxk{l}$ is on CPU, so the layer below can be addressed (or, for $l=0$, gradients w.r.t.\ input embeddings can be computed based on $\partial\mathcal{L} / \partial\mxk{0}$).
\end{itemize}

Finally, note that our implementation also supports cells to span several layers in the vertical direction, which can be used if enough GPU RAM is available. This speeds up computations (fewer cells mean less sequential computations) and also lowers the amount of CPU memory required for checkpointing, since inputs need to be stored only for layers at the bottom of a row of cells. Our exposition above can easily be generalized by letting $l$ iterate over rows of cells in the lattice, rather than layers.

\subsubsection{Exploiting Linear Recurrence of KV Cache Buffers}
\label{sec:app-exploit-recurrence}

Recall the recurrence between KV cache buffers for subsequent chunks from \secref{main-finetune}. We can exploit this recurrence by asking {\em PyTorch} to store $\mathtt{delta\_key}$ instead of $\mathtt{keys}$ and $\mathtt{delta\_value}$ instead of $\mathtt{values}$ in the computation graph, restoring the latter from the former during backward using \eqp{recurrence-reverse}.

While this is simple in principle, we need to do it inside {\em PyTorch autograd}. To this end, we use a mechanism called {\em autograd saved tensors hooks} (\url{https://docs.pytorch.org/tutorials/intermediate/autograd_saved_tensors_hooks_tutorial.html}). This was designed in order to implement activation checkpointing by CPU offloading, but can be used for our purposes as well. It works by allowing the specification of two functions:
\begin{itemize}
\item
  $\mathtt{pack\_hook}(\vx{})\to \vp{}(\vx{})$: When building its computation graph during forward, this function is called for every array $\vx{}$ PyTorch plans to store in the computation graph. It then stores $\vp{}(\vx{})$ in the graph instead of $\vx{}$.
\item
  $\mathtt{unpack\_hook}(\vp{})\to \vx{}(\vp{})$: When traversing the computation graph in reverse order during backward, PyTorch calls this function for every array $\vp{}$ stored in the graph. It then uses $\vx{}(\vp{})$ instead of $\vp{}$.
\end{itemize}

A major difficulty for us is the non-selectiveness of this mechanism. We do not want to pack all arrays stored in the graph, but only specific ones: the KV cache buffers (for all other nodes, we just pass through $\vp{}(\vx{}) = \vx{}$ and $\vx{}(\vp{}) = \vp{}$). But $\mathtt{pack\_hook}(\vx{})$ just takes a $\mathtt{torch.Tensor}$ argument, there is no obvious way for tagging the nodes we want. Also, there is some delay between a node being created in the forward pass and $\mathtt{pack\_hook}(\vx{})$ being called for it, we even detected some differences in the relative ordering. Finally, due to internal operator fusion, we cannot even be sure whether any node appearing in the forward code is indeed stored in the graph.

Our implementation maintains an {\em annotation list}, which is appended to during the forward code, while entries are removed during $\mathtt{pack\_hook}$ calls. Whenever a KV cache buffer update in the form of a statement $\mathtt{keys}' = \mathtt{scatter}(\mathtt{keys}, \mathtt{index}, \mathtt{key\_new})$ is passed in the forward code, we compute $\mathtt{delta\_key} = \mathtt{gather}(\mathtt{keys}, \mathtt{index})$, appending an {\em annotation} containing $(\mathtt{index}, \mathtt{delta\_key})$ and some meta-data to the list. Here, $\mathtt{delta\_key}$ serves a double role. First, it is needed to reconstruct $\mathtt{keys}$ from $\mathtt{keys}'$ in $\mathtt{unpack\_hook}$. Second, it serves\footnote{
  If $\mathtt{index}$ is small, we extend the fingerprint by additional random positions in order to avoid false matches.}
as a "fingerprint" of $\mathtt{keys}$. Namely, when $\mathtt{pack\_hook}(\vx{})$ is called, we need to match the argument $\vx{}$ against annotations. This is first done by shape, filtering out most calls. Next, for any annotation $(\mathtt{index}, \mathtt{delta\_key})$, we check whether $\mathtt{gather}(\vx{}, \mathtt{index}) = \mathtt{delta\_key}$. If so, we return $\vp{}(\vx{})$ containing $\mathtt{delta\_key}$ and remove the annotation from the list. If there is no match, we return $\vp{}(\vx{}) = \vx{}$.

For $\mathtt{unpack\_hook}(\vp{})$, we reconstruct the sequences $\mathtt{keys}$ and $\mathtt{values}$ in reverse order. If $\vp{}$ is not a packed object, we return $\vx{}(\vp{}) = \vp{}$. Otherwise, we check the chunk number stored with $\vp{}$ against the current state $(\mathtt{keys}', \mathtt{values}')$. If this fits, we reconstruct $\mathtt{keys}$ from $(\mathtt{keys}', \mathtt{delta\_key})$ or $\mathtt{values}$ from $(\mathtt{values}', \mathtt{delta\_value})$, using \eqp{recurrence-reverse}. The new buffer overwrites the old one.

Our implementation tracks which pack hook arguments of the right shape are not matched by annotations, and which annotations are not matched. Both events do happen, but at a low rate. It is important to note that we still obtain correct results even if some arguments are not matched. This just means that a bit more GPU memory is being used. What we need to avoid, however, are false matches, which we do by keeping fingerprints large enough.

An important direction for future work is to simplify and robustify the mechanism for exploiting the linear recurrences. We tried several simplifications. One assumes that KV buffer nodes are created in exactly the same order as $\mathtt{pack\_hook}(\vx{})$ calls. If true, matching and managing the annotation list would be much simplified. Unfortunately, this does not hold true, likely due to internals of PyTorch autograd we have no influence over. In fact, the instruction $\mathtt{keys}' = \mathtt{scatter}(\mathtt{keys}, \mathtt{index}, \mathtt{key\_new})$ need not even trigger a call of $\mathtt{pack\_hook}$ with $\vx{} = \mathtt{keys}'$. The array $\mathtt{keys}'$ is processed further inside the SDPA code, and PyTorch may use some form of operator fusion. The simplest solution would be to tag each KV buffer node during creation in a way that allows us to recognize the tag in a pack hook argument $\vx{}$. However, we did not find a way to do that yet.

\subsection{Experimental Results}

In this section, we provide additional experimental results beyond what is shown in the main text, as well as timing figures. We also explain a failure mode we consistently observed when fine-tuning a model with sequence parallelism to be used for inference with sparse attention.

\subsubsection{Additional Results}\label{sec:app-results-additional}

\begin{table}[ht!]
\centering
\begin{tabular}{|l|r|r|r|r|r|r|r|r|}
\hline
 & \multicolumn{4}{c|}{64k datasets} & \multicolumn{4}{c|}{128k datasets} \\
\hline
 & nq & tri\_qa & hot\_qa & pop\_qa & nq & tri\_qa & hot\_qa & pop\_qa \\
\hline\hline
\rule{0pt}{11pt} $\mathrm{lr}_{2k}$ &
   {\small\!0.0} &  {\small\!1.2} &  {\small\!1.3} &  {\small\!0.0} &
   {\small\!0.2} &  {\small\!2.3} &  {\small\!2.3} &  {\small\!0.2} \\
\rule{0pt}{11pt} $\mathrm{slr}_{2k}$ &
   {\small\!1.3} & {\small\!16.2} &  {\small\!7.0} &  {\small\!5.0} &
   {\small\!0.5} & {\small\!10.2} &  {\small\!3.3} &  {\small\!0.8} \\
\rule{0pt}{11pt} $\mathrm{h2o}_{2k}$ &
  {\small\!17.3} & {\small\!58.8} & {\small\!19.7} & {\small\!25.7} &
   {\small\!4.5} & {\small\!30.5} &  {\small\!6.7} &  {\small\!9.8} \\
\rule{0pt}{11pt} $\mathrm{h2o}_{2k}^{\text{no}}$ &
  {\small\!15.5} & {\small\!48.8} & {\small\!17.0} & {\small\!23.7} &
   {\small\!3.0} &  {\small\!6.0} &  {\small\!4.7} &  {\small\!2.3} \\
\rule{0pt}{11pt} $\mathrm{h2o}_{2k}^{\text{or}}$ &
  {\small\!17.8} & {\small\!56.7} & {\small\!20.3} & {\small\!26.2} &
   {\small\!7.0} & {\small\!40.0} &  {\small\!9.3} &  {\small\!9.7} \\
\rule{0pt}{11pt} $\mathrm{lr}_{1k}$ &
   {\small\!0.2} & {\small\!1.2} & {\small\!2.7} & {\small\!0.0} &
   {\small\!0.0} & {\small\!2.8} & {\small\!2.3} & {\small\!0.0} \\
\rule{0pt}{11pt} $\mathrm{slr}_{1k}$ &
   {\small\!2.2} & {\small\!16.3} & {\small\!7.0} & {\small\!4.5} &
   {\small\!0.3} & {\small\!10.0} & {\small\!3.3} & {\small\!0.3} \\
\rule{0pt}{11pt} $\mathrm{h2o}_{1k}$ &
  {\small\!17.5} & {\small\!58.7} & {\small\!19.0} & {\small\!27.7} &
   {\small\!5.8} & {\small\!30.7} & {\small\!5.7} & {\small\!9.8} \\
\rule{0pt}{11pt} $\mathrm{h2o}_{1k}^{\text{no}}$ &
  {\small\!13.0} & {\small\!50.3} & {\small\!15.7} & {\small\!21.8} &
   {\small\!2.8} & {\small\!6.8} & {\small\!3.7} & {\small\!2.3} \\
\rule{0pt}{11pt} $\mathrm{h2o}_{1k}^{\text{or}}$ &
  {\small\!17.7} & {\small\!60.5} & {\small\!21.7} & {\small\!27.8} &
   {\small\!8.0} & {\small\!43.3} & {\small\!10.7} & {\small\!11.2} \\
\hline\hline
 & \multicolumn{8}{c|}{128k datasets} \\
\hline
 & trec\_c & trec\_f & nlu & bank77 & clc150 & inf\_qa & inf\_mc & json\_kv \\
\hline\hline
\rule{0pt}{11pt} $\mathrm{slr}_{1k}$ &
  {\small\!28.2} & {\small\!25.2} & {\small\!24.8} & {\small\!51.4} &
  {\small\!62.6} &  {\small\!2.5} & {\small\!36.0} &  {\small\!0.0} \\
\rule{0pt}{11pt} $\mathrm{h2o}_{1k}^{\text{no}}$ &
  {\small\!19.8} & {\small\!26.0} & {\small\!30.0} & {\small\!57.0} &
  {\small\!54.0} &  {\small\!2.9} & {\small\!41.0} &  {\small\!0.0} \\
\rule{0pt}{11pt} $\mathrm{h2o}_{1k}^{\text{or}}$ &
  {\small\!36.0} & {\small\!15.8} & {\small\!21.2} & {\small\!16.6} &
  {\small\!34.8} &  {\small\!3.4} & {\small\!40.0} &  {\small\!0.0} \\
\hline
\end{tabular}
\caption{\label{tab:results-basemodel}
  Results for long-context inference on Helmet datasets. Here, the base checkpoint {\tt Qwen3-4B-Instruct-2507} is used without fine-tuning. In the upper part, we use 4 datasets of sequence lengths 64k, 128k each (cols), run 5 KV cache policies and 2 chunk sizes $2048 = 2k, 1024 = 1k$ (rows). In the lower part, we use 8 additional datasets and run 3 KV cache policies (chunk size $1024 = 1k$).}
\end{table}

{\bf Results for Base Checkpoints}. In \tabref{results-basemodel}, we provide results on 4 Helmet QA datasets (64k, 128k sequence lengths) and 8 additional Helmet datasets for base checkpoints {\tt Qwen3-4B-Instruct-2507} (no fine-tuning). They should be related to results in \tabref{helmet-results}, where the base model was trained by our method (columns {\em us}) or by sequence parallelism (columns {\em sp}).

\begin{table}[ht!]
\centering
\begin{tabular}{|l|r|r|r|r|r|r|r|r|}
\hline
 & \multicolumn{4}{c|}{64k datasets} & \multicolumn{4}{c|}{128k datasets} \\
\hline
 & nq & tri\_qa & hot\_qa & pop\_qa & nq & tri\_qa & hot\_qa & pop\_qa \\
\hline\hline
\rule{0pt}{11pt} $\mathrm{lr}_{128}$ &
  {\small\!34.3} & {\small\!66.8} & {\small\!43.3} & {\small\!42.2} &
  - & - & - & - \\
\rule{0pt}{11pt} $\mathrm{slr}_{128}$ &
  {\small\!37.3} & {\small\!61.8} & {\small\!41.7} & {\small\!42.2} &
  {\small\!22.7} & {\small\!56.8} & {\small\!24.7} & {\small\!31.3} \\
\rule{0pt}{11pt} $\mathrm{h2o}_{128}$ &
  {\small\!41.3} & {\small\!72.2} & {\small\!45.3} & {\small\!47.7} &
  {\small\!16.5} & {\small\!42.2} & {\small\!13.7} & {\small\!33.0} \\
\rule{0pt}{11pt} $\mathrm{h2o}_{128}^{\text{no}}$ &
  {\small\!41.7} & {\small\!64.8} & {\small\!44.7} & {\small\!46.8} &
  {\small\!36.3} & {\small\!71.0} & {\small\!30.3} & {\small\!38.0} \\
\rule{0pt}{11pt} $\mathrm{h2o}_{128}^{\text{or}}$ &
  {\small\!41.5} & {\small\!71.3} & {\small\!38.3} & {\small\!49.2} &
  {\small\!42.2} & {\small\!73.7} & {\small\!23.3} & {\small\!40.7} \\
\hline
\rule{0pt}{11pt} $\mathrm{lr}_{8k}$ &
  {\small\!35.3} & {\small\!67.2} & {\small\!44.3} & {\small\!45.5} &
  {\small\!21.5} & {\small\!55.7} & {\small\!27.0} & {\small\!32.3} \\
\rule{0pt}{11pt} $\mathrm{slr}_{8k}$ &
  {\small\!29.7} & {\small\!70.3} & {\small\!41.3} & {\small\!41.5} &
  {\small\!25.5} & {\small\!56.3} & {\small\!24.0} & {\small\!33.5} \\
\rule{0pt}{11pt} $\mathrm{h2o}_{8k}$ &
  {\small\!36.0} & {\small\!47.5} & {\small\!33.0} & {\small\!46.0} &
  {\small\!24.2} & {\small\!40.2} & {\small\!14.0} & {\small\!38.7} \\
\rule{0pt}{11pt} $\mathrm{h2o}_{8k}^{\text{no}}$ &
  {\small\!41.0} & {\small\!66.2} & {\small\!43.7} & {\small\!48.5} &
  {\small\!36.7} & {\small\!72.8} & {\small\!28.0} & {\small\!44.0} \\
\rule{0pt}{11pt} $\mathrm{h2o}_{8k}^{\text{or}}$ &
  {\small\!42.0} & {\small\!71.7} & {\small\!38.7} & {\small\!48.5} &
  {\small\!36.7} & {\small\!75.0} & {\small\!28.7} & {\small\!45.0} \\
\hline
\rule{0pt}{13pt} $\mathrm{qh2o}_{2k}$ &
  {\small\!30.2} & {\small\!65.8} & {\small\!35.3} & {\small\!42.3} &
  - & - & - & - \\
\rule[-5pt]{0pt}{16pt} $\mathrm{qh2o}_{2k}^{\text{no}}$ &
  {\small\!34.5} & {\small\!59.3} & {\small\!37.0} & {\small\!46.8} &
  - & - & - & - \\
\hline\hline
  & \multicolumn{8}{c|}{128k datasets} \\
\hline
  & trec\_c & trec\_f & nlu & bank77 & clc150 & inf\_qa & inf\_mc & json\_kv \\
\hline\hline
\rule{0pt}{13pt} $\mathrm{slr}_{8k}$ &
  {\small\!96.0} & {\small\!87.6} & {\small\!87.2} & {\small\!92.2} &
  {\small\!96.6} & {\small\!23.0} & {\small\!43.0} & {\small\!48.0} \\
\rule{0pt}{11pt} $\mathrm{h2o}_{8k}^{\text{no}}$ &
  {\small\!94.2} & {\small\!90.8} & {\small\!87.2} & {\small\!91.8} &
  {\small\!98.0} & {\small\!31.9} & {\small\!53.0} & {\small\!44.0} \\
\rule{0pt}{11pt} $\mathrm{h2o}_{8k}^{\text{or}}$ &
  {\small\!96.2} & {\small\!88.8} & {\small\!85.0} & {\small\!87.6} &
  {\small\!95.2} & {\small\!33.3} & {\small\!49.0} & {\small\!0.0} \\
\hline
\end{tabular}
\caption{\label{tab:results-ablations}
  Results for long-context inference with setups not covered in the main text. We show {\em MatchFirstWordOrPhrase} values on test splits for different Helmet datasets.
  $\mathrm{slr}_{128}$, $\mathrm{h2o}_{128}$, $\mathrm{h2o}_{128}^{\text{no}}$, $\mathrm{h2o}_{128}^{\text{or}}$ use chunk size $S=128$.
  $\mathrm{lr}_{8k}$, $\mathrm{slr}_{8k}$, $\mathrm{h2o}_{8k}$, $\mathrm{h2o}_{8k}^{\text{no}}$, $\mathrm{h2o}_{8k}^{\text{or}}$ use chunk size $S=8192$.
  $\mathrm{qh2o}_{2k}$ and $\mathrm{qh2o}_{2k}^{\text{no}}$ are variants of Q-Hitter~\citep{Zhang:24}.
}
\end{table}

{\bf Results for Other Chunk Sizes, Extra Policies}. In \tabref{results-ablations}, we provide results on Helmet 64k and 128k datasets for setups not covered in the main text (see \tabref{helmet-results}). First, we study the effect of chunk size $S$. $\mathrm{slr}_{128}$, $\mathrm{h2o}_{128}$, $\mathrm{h2o}_{128}^{\text{no}}$, $\mathrm{h2o}_{128}^{\text{or}}$ use chunk size $S=128$. These run significantly longer than $S\in\{1024, 2048\}$ used in the main text experiments (see \tabref{time-train-64k}), but allow the KV cache policy to make decisions 8 or 16 times more frequently. $\mathrm{slr}_{8k}$, $\mathrm{h2o}_{8k}$, $\mathrm{h2o}_{8k}^{\text{no}}$, $\mathrm{h2o}_{8k}^{\text{or}}$ use chunk size $S=8192$. These run much faster than $S\in\{1024, 2048\}$, yet the KV cache policy can make decisions 4 times only before the cache is rolled over (recall that the cache length is $N_c = 32768$). At least for policies such as H2O, we would expect $S=128$ to work significantly better and $S=8192$ to work significantly worse than what we did in the main text. However, this hunch is not supported by empirical evidence. Based on our findings so far, we recommend to choose the chunk size $S$ as large as possible before results degrade significantly for the chosen policy, compared to a conservative choice of $N_c / 8$ or $N_c / 16$.

We also ran experiments with Q-Hitter~\citep{Zhang:24}, where KV cache buffers are quantized (to 8 bits in our experiments) and the decision score is a convex combination of \eqp{h2o-score} and a term quantifying the quantization error. The results are consistently worse than for the H2O variants.

{\bf Results for Final Cell Ablation}. Our method splits a long context into cells, each about as long as the number of KV cache slots, then computes gradients by running {\em autograd} on each cell (see \secref{app-grad-summary}). A simplification is obtained by cutting the computation graph at the KV cache buffer nodes on the boundary between cells. Consider the middle panel (b) in \figref{method-diagram}. Cutting the graph means that $\partial\mathcal{L} / \partial\mxk{l, c+1} = \mxzero$ for each cell $c$. If the loss $\mathcal{L}$ only depends on the network outputs $\mxx{L, N_{\text{cells}} - 1}$ of the final cell $c = N_{\text{cells}} - 1$, then $\partial\mathcal{L} / \partial\mxx{l, c} = \mxzero$ for all $c < N_{\text{cells}} - 1$ and all layers $l=\rng[0]{L}$. This means that {\em autograd} has to be run for the final cells $c = N_{\text{cells}} - 1$ only, since gradient contributions from all earlier cells and all layers are zero. We call this variant the {\em final cell ablation}\footnote{
  We would like to thank an anonymous reviewer for pointing out this ablation.}.

The implementation effort for this ablation is the same as for our method, since both nested activation checkpointing and delta encoding of cache buffers are still needed. However, the ablation requires many less {\em autograd} calls if the loss only depends on outputs at the end of the batch. Also, only the KV cache buffer $\mxk{l, N_{\text{cells}} - 1}$ mapping into the final cell has to be checkpointed, so less CPU memory is needed. Results are provided in \tabref{final-cell-results}.

\begin{table}[ht!]
\centering
\begin{tabular}{|l|r|l|rr|rr|}
\hline
\rule{0pt}{13pt} Benchmark & Chunk Size & Dataset & \multicolumn{2}{c|}{$\mathrm{h2o}^{\text{no}}$} & \multicolumn{2}{c|}{Train time} \\
  & & & us & fca & us & fca \\
\hline
\rule{0pt}{13pt} Helmet (128k) & 1024
    & trec\_c  & {\small\!96.0} & {\small\!94.4} & {\small\!306.67 (13.70)} & {\small\!167.69 (08.42)} \\
  & & trec\_f  & {\small\!90.6} & {\small\!87.4} & {\small\!318.65 (14.36)} & {\small\!332.38 (28.71)} \\
  & & nlu      & {\small\!86.6} & {\small\!88.2} & {\small\!330.84 (02.54)} & {\small\!148.81 (01.54)} \\
  & & bank77   & {\small\!91.0} & {\small\!93.2} & {\small\!327.38 (09.05)} & {\small\!149.11 (04.90)} \\
  & & clc150   & {\small\!97.8} & {\small\!97.6} & {\small\!342.47 (09.73)} & {\small\!158.25 (04.56)} \\
  & & inf\_qa  & {\small\!32.1} & {\small\!33.0} & {\small\!349.79 (69.99)} & {\small\!167.36 (24.18)} \\
  & & inf\_mc  & {\small\!53.0} & {\small\!47.0} & {\small\!368.92 (56.54)} & {\small\!170.00 (14.18)} \\
  & & json\_kv & {\small\!52.0} & {\small\!29.0} & {\small\!368.36 (23.29)} & {\small\!190.96 (12.44)} \\
\hline
\rule{0pt}{13pt} RULER (256k) & 8192
    & niah\_s\_1 & {\small\!96.0} & {\small\!97.0} & {\small\!650.97 (01.03)} & {\small\!306.76 (04.30)} \\
  & & vt         & {\small\!99.4} & {\small\!97.6} & {\small\!676.25 (03.01)} & {\small\!305.12 (06.64)} \\
  & & cwe        &  {\small\!100} & {\small\!98.5} & {\small\!939.75 (04.98)} & {\small\!442.77 (09.41)} \\
  & & fwe        &  {\small\!100} &  {\small\!100} & {\small\!664.24 (16.93)} & {\small\!308.62 (11.24)} \\
\hline
\end{tabular}
\caption{\label{tab:final-cell-results}
  Results on 8 Helmet datasets (128k) and 4 RULER datasets (256k), comparing our training method ({\em us}) with the final cell ablation ({\em fca}), a slight simplification of what we do. {\em Train time} are times for update step in secs (mean, stddev).
}
\end{table}

Overall, the ablation comes close to our unmodified method on most of the datasets. Only on Helmet {\tt json\_kv} and {\tt inf\_mc} (to a lesser extent), cutting the {\em autograd} graph at cell boundaries along the sequence hurts performance. The ablation runs significantly faster.

Under what conditions can the final cell ablation provide gradients which lead to successful fine-tuning? Even though the loss $\mathcal{L}$ directly depends on the final cell outputs $\mxx{L, N_{\text{cells}} - 1}$ only, network weights determine loss values via contributions from all cells. The final cell ablation neglects such intermediate contributions from all but the final cells. It can be successful if these final cell contributions dominate all others.

\subsubsection{Running Time Figures}\label{sec:app-results-time}

\begin{table}[ht!]
\centering
\begin{tabular}{|l|r|r|r|r|}
\hline
 & \multicolumn{4}{c|}{64k datasets} \\
\hline
 & nq & tri\_qa & hot\_qa & pop\_qa \\
\hline\hline
\rule{0pt}{11pt} $\mathrm{lr}_{8k}$ &
  {\small\!107.53 (08.83)} & {\small\!108.59 (06.96)} & {\small\!107.49 (09.97)} & {\small\!98.71 (10.97)} \\
\rule{0pt}{11pt} $\mathrm{slr}_{8k}$ &
  {\small\!107.87 (08.88)} & {\small\!109.05 (07.20)} & {\small\!109.34 (10.04)} & {\small\! 99.62 (11.02)} \\
\rule{0pt}{11pt} $\mathrm{h2o}_{8k}$ &
  {\small\!108.19 (09.15)} & {\small\!109.01 (07.36)} & {\small\!110.08 (10.27)} & {\small\!100.59 (11.21)} \\
\rule{0pt}{11pt} $\mathrm{h2o}_{8k}^{\text{no}}$ &
  {\small\!108.09 (08.83)} & {\small\!111.70 (07.67)} & {\small\!109.08 (09.83)} & {\small\!102.04 (11.47)} \\
\rule{0pt}{11pt} $\mathrm{h2o}_{8k}^{\text{or}}$ &
  {\small\!107.36 (08.73)} & {\small\!110.41 (07.38)} & {\small\!110.30 (10.11)} & {\small\!99.27 (11.14)} \\
\hline
\rule{0pt}{11pt} $\mathrm{lr}_{2k}$ &
  {\small\!113.89 (09.07)} & {\small\!116.89 (07.81)} & {\small\!117.20 (11.45)} & {\small\!106.92 (12.64)} \\
\rule{0pt}{11pt} $\mathrm{slr}_{2k}$ &
  {\small\!116.44 (09.19)} & {\small\!119.54 (07.94)} & {\small\!118.77 (11.50)} & {\small\!106.94 (12.16)} \\
\rule{0pt}{11pt} $\mathrm{h2o}_{2k}$ &
  {\small\!117.93 (09.71)} & {\small\!121.20 (08.09)} & {\small\!118.65 (11.73)} & {\small\!107.93 (12.80)} \\
\rule{0pt}{11pt} $\mathrm{h2o}_{2k}^{\text{no}}$ &
  {\small\!116.19 (09.40)} & {\small\!118.93 (08.04)} & {\small\!118.79 (11.60)} & {\small\!107.30 (12.43)} \\
\rule{0pt}{11pt} $\mathrm{h2o}_{2k}^{\text{or}}$ &
  {\small\!116.49 (09.52)} & {\small\!118.99 (08.11)} & {\small\!119.74 (11.90)} & {\small\!106.95 (12.44)} \\
\hline
\rule{0pt}{11pt} $\mathrm{lr}_{1k}$ &
  {\small\!125.08 (09.93)} & {\small\!127.51 (08.65)} & {\small\!127.47 (12.71)} & {\small\!115.15 (13.71)} \\
\rule{0pt}{11pt} $\mathrm{slr}_{1k}$ &
  {\small\!125.94 (09.70)} & {\small\!129.88 (08.50)} & {\small\!129.23 (12.57)} & {\small\!115.57 (14.00)} \\
\rule{0pt}{11pt} $\mathrm{h2o}_{1k}$ &
  {\small\!130.09 (10.58)} & {\small\!131.01 (08.91)} & {\small\!130.22 (12.95)} & {\small\!117.00 (14.42)} \\
\rule{0pt}{11pt} $\mathrm{h2o}_{1k}^{\text{no}}$ &
  {\small\!127.91 (12.85)} & {\small\!132.05 (08.80)} & {\small\!129.58 (13.12)} & {\small\!117.74 (14.45)} \\
\rule{0pt}{11pt} $\mathrm{h2o}_{1k}^{\text{or}}$ &
  {\small\!129.33 (10.38)} & {\small\!131.31 (08.75)} & {\small\!131.15 (13.10)} & {\small\!116.71 (14.57)} \\
\hline
\rule{0pt}{11pt} $\mathrm{lr}_{128}$ &
  {\small\!286.62 (28.87)} & {\small\!294.01 (26.25)} & {\small\!294.46 (38.68)} & {\small\!253.74 (40.56)} \\
\rule{0pt}{11pt} $\mathrm{slr}_{128}$ &
  {\small\!287.88 (28.40)} & {\small\!294.84 (25.34)} & {\small\!293.12 (39.09)} & {\small\!252.69 (40.42)} \\
\rule{0pt}{11pt} $\mathrm{h2o}_{128}$ &
  {\small\!297.04 (31.66)} & {\small\!301.76 (27.60)} & {\small\!311.17 (41.80)} & {\small\!270.93 (43.52)} \\
\rule{0pt}{11pt} $\mathrm{h2o}_{128}^{\text{no}}$ &
  {\small\!308.21 (30.90)} & {\small\!305.48 (27.17)} & {\small\!296.32 (39.69)} & {\small\!271.91 (43.13)} \\
\rule{0pt}{11pt} $\mathrm{h2o}_{128}^{\text{or}}$ &
  {\small\!307.02 (30.81)} & {\small\!301.30 (27.40)} & {\small\!300.77 (40.09)} & {\small\!272.24 (43.42)} \\
\hline
\end{tabular}
\caption{\label{tab:time-train-64k}
  Running time figures for training update step, for Helmet 64k datasets (columns), 5 KV cache policies and chunk sizes $2048 = 2k, 1024 = 1k, 128$ (rows). Batch size 8, running on 4 devices.
  The step from 2k to 1k is 8\% to 10\% more expensive for $\mathtt{lr}, \mathtt{slr}$, 9\% to 11\% more expensive for $\mathtt{h2o}$ variants. The step from 2k to 128 is 136\% to 147\% more expensive for $\mathtt{lr}, \mathtt{slr}$, 151\% to 165\% more expensive for $\mathtt{h2o}$ variants. The step from $\mathtt{slr}$ to $\mathtt{h2o}$ is 1\% to 3\% more expensive.
}
\end{table}

\begin{table}[ht!]
\centering
\begin{tabular}{|l|r|r|r|r|}
\hline
 & \multicolumn{4}{c|}{128k datasets} \\
\hline
 & nq & tri\_qa & hot\_qa & pop\_qa \\
\hline\hline
\rule{0pt}{11pt} $\mathrm{exact}$ &
  {\small\!258.38 (15.14)} & {\small\!266.05 (11.31)} & {\small\!262.53  (08.18)} & {\small\!236.74 (21.44)} \\
\hline
\rule{0pt}{11pt} $\mathrm{lr}_{8k}$ &
  {\small\!283.05 (20.04)} & {\small\!290.81 (14.03)} & {\small\!293.75 (19.29)} & {\small\!273.51 (21.29)} \\
\rule{0pt}{11pt} $\mathrm{slr}_{8k}$ &
  {\small\!286.33 (19.15)} & {\small\!290.87 (14.16)} & {\small\!290.56 (19.79)} & {\small\!275.69 (21.33)} \\
\rule{0pt}{11pt} $\mathrm{h2o}_{8k}$ &
  {\small\!288.85 (19.30)} & {\small\!298.16 (15.31)} & {\small\!293.73 (20.63)} & {\small\!275.65 (21.59)} \\
\rule{0pt}{11pt} $\mathrm{h2o}_{8k}^{\text{no}}$ &
  {\small\!290.10 (19.26)} & {\small\!294.45 (14.40)} & {\small\!294.46 (19.58)} & {\small\!277.91 (21.65)} \\
\rule{0pt}{11pt} $\mathrm{h2o}_{8k}^{\text{or}}$ &
  {\small\!291.20 (19.30)} & {\small\!294.69 (14.54)} & {\small\!295.98 (19.53)} & {\small\!279.62 (21.80)} \\
\hline
\rule{0pt}{11pt} $\mathrm{lr}_{2k}$ &
  {\small\!324.02 (22.06)} & {\small\!331.44 (16.58)} & {\small\!331.07 (22.63)} & {\small\!309.37 (24.89)} \\
\rule{0pt}{11pt} $\mathrm{slr}_{2k}$ &
  {\small\!327.93 (21.14)} & {\small\!333.46 (16.24)} & {\small\!335.46 (22.67)} & {\small\!310.74 (24.54)} \\
\rule{0pt}{11pt} $\mathrm{h2o}_{2k}$ &
  {\small\!330.83 (21.98)} & {\small\!338.41 (16.72)} & {\small\!336.79 (23.13)} & {\small\!315.27 (25.83)} \\
\rule{0pt}{11pt} $\mathrm{h2o}_{2k}^{\text{no}}$ &
  {\small\!331.48 (22.10)} & {\small\!343.78 (16.75)} & {\small\!337.77 (23.12)} & {\small\!317.60 (25.16)} \\
\rule{0pt}{11pt} $\mathrm{h2o}_{2k}^{\text{or}}$ &
  {\small\!331.83 (22.25)} & {\small\!339.15 (17.10)} & {\small\!335.67 (23.27)} & {\small\!316.94 (25.34)} \\
\hline
\rule{0pt}{11pt} $\mathrm{lr}_{1k}$ &
  {\small\!363.20 (23.20)} & {\small\!376.60 (42.67)} & {\small\!368.44 (25.72)} & {\small\!345.91 (28.32)} \\
\rule{0pt}{11pt} $\mathrm{slr}_{1k}$ &
  {\small\!363.44 (22.89)} & {\small\!371.39 (18.42)} & {\small\!372.78 (26.07)} & {\small\!344.02 (27.95)} \\
\rule{0pt}{11pt} $\mathrm{h2o}_{1k}$ &
  {\small\!378.85 (24.88)} & {\small\!385.19 (20.21)} & {\small\!380.18 (26.43)} & {\small\!354.78 (28.86)} \\
\rule{0pt}{11pt} $\mathrm{h2o}_{1k}^{\text{no}}$ &
  {\small\!378.61 (24.99)} & {\small\!385.29 (19.32)} & {\small\!383.90 (27.97)} & {\small\!357.95 (29.02)} \\
\rule{0pt}{11pt} $\mathrm{h2o}_{1k}^{\text{or}}$ &
  {\small\!378.65 (24.79)} & {\small\!385.64 (19.46)} & {\small\!382.52 (26.91)} & {\small\!358.44 (29.38)} \\
\hline
\rule{0pt}{11pt} $\mathrm{slr}_{128}$ &
  - & {\small\!1188.67 (80.34)} & - & {\small\!1103.14 (119.21)} \\
\rule{0pt}{11pt} $\mathrm{h2o}_{128}$ &
  {\small\!1101.79 (86.67)} & {\small\!1203.90 (68.67)} & - & {\small\!1092.81 (100.48)} \\
\rule{0pt}{11pt} $\mathrm{h2o}_{128}^{\text{no}}$ &
  {\small\!1105.39 (84.22)} & - & {\small\!1151.96 (95.65)} & {\small\!1090.21 (101.01)} \\
\rule{0pt}{11pt} $\mathrm{h2o}_{128}^{\text{or}}$ &
  {\small\!1120.60 (78.08)} & - & {\small\!1147.95 (93.22)} & {\small\!1109.41 (102.78)} \\
\hline
\end{tabular}
\caption{\label{tab:time-train-128k}
  Running time figures for training update step, for Helmet 128k datasets (columns), 5 KV cache policies and chunk sizes $2048 = 2k, 1024 = 1k, 8192 = 8k$ (rows). Batch size 8, running on 4 devices.
  The step from 2k to 1k is 11\% to 14\% more expensive for $\mathtt{lr}, \mathtt{slr}$, 13\% to 15\% more expensive for $\mathtt{h2o}$ variants. The step from $\mathtt{slr}$ to $\mathtt{h2o}$ is 1\% to 4\% more expensive.
}
\end{table}

In this section, we present running time figures for training updates (batch size 8; four Nvidia A100s with 40 GB each). For Helmet 128k datasets, {\em exact} uses sequence parallelism with batch size 2, processing 4 micro-batches sequentially, whereas our method (for different cache logics) processes 4 micro-batches in parallel.

First, our method is about 9-18\% more expensive than {\em exact} for 128k datasets and chunk size $S=8192$ (8k), and 25-30\% more expensive for chunk size $S=2048$ (2k). Given that our method computes gradients on a single GPU independent of the sequence length, using advanced cache logics such as H2O, nested activation checkpointing and delta encoding of cache buffers, this overhead is surprisingly small. Reasons for the overhead are explained in \secref{main-satt-vs-cp}. The gap can likely be narrowed further by operator fusion, increasing the chunk size {\em autograd} is operating with. Next, we would expect $S=1024$ (1k) to run slower than $S=2048$ (2k), and $S=2048$ to run slower than $S=8192$, because more chunks need to be processed sequentially; and H2O policies to run slower than $\mathtt{lr}, \mathtt{slr}$, because summed attention weights are required, and scores need to be computed and sorted. We see that chunk size 1k variants run between 8\% and 11\% longer than 2k variants for 64k datasets, and between 11\% and 14\% longer than 2k variants for 128k datasets, which is significant. The slowdown is more pronounced when lowering the chunk size to $S=128$, which runs between 136\% and 165\% longer than for $S=2048$.

On the other hand, H2O variants are only between 1\% to 4\% slower than $\mathtt{lr}, \mathtt{slr}$. At least with our fast SDPA implementation, there is no penalty for using more advanced policies over simple baselines.

\subsubsection{Analysis of Errors}\label{sec:app-error-analysis}

In this paper, we compare different ways of fine-tuning a model to be used with sparse attention inference under different KV cache logics: training by sequence parallelism ({\em sp}) versus training with the new, resource-efficient technique developed here ({\em us}). Results in \tabref{helmet-results} indicate that {\em us} strongly outperforms {\em sp} on all datasets except {\tt pop\_qa}. In this section, we identify a consistent failure mode of {\em sp}, both via randomly chosen examples and statistics on the generated samples.

Recall details about Helmet datasets from \secref{app-data-helmet}. For {\tt trec\_coarse}, {\tt trec\_fine}, {\tt nlu}, {\tt banking77}, {\tt clinc150}, {\tt inf\_mc}, targets are single integers, and the {\em Accuracy} metric requires the correct number to appear most frequently in the output. For  {\tt nq, trivia\_qa, hotpot\_qa, pop\_qa, json\_kv}, targets are lists of noun phrases of one or few words, and the {\em MatchFirstWordOrPhrase} metric requires the target to match the first word or sentence of the output. As with all sample-based metrics in our experiments, we allow the model to generate up to 128 tokens, or otherwise terminate by emitting the {\tt <eos>} token. A closer look at generated samples reveals a major failure mode of {\em sp}: {\em its outputs are far too long and contain mostly random nonsense}. Here are some randomly chosen examples of tasks and outputs under {\em us} and {\em sp} respectively.

{\bf trec\_coarse / slr\_1024}:

{\tiny
\begin{lstlisting}[language=bash,showstringspaces=false]
- idx: 10
  raw_target: "2"
  exact_match_us: 1
  output_us: "2"
  exact_match_sp: 0
  output_sp: " 2
    What is the average life expectancy of a female in Ireland in 1996 ?
    1
    What is the name of the American who was captured when his plane went down over
    Syrian-held Lebanon ?
    1
    0
    What is the full name of the PLO ?
    1
    3
    What is the difference between a college and a university ?
    1
    4
    What is the origin of the term `` The Brooklyn Side '''' in bowling ?
    1
    4
    What is the origin of the word `` assassinate '''' ?
    1
    4
    What does the abbreviation cwt. ? mean ?
    1
    3
    What does the"
- idx: 318
  raw_target: "5"
  exact_match_us: 1
  output_us: "5"
  exact_match_sp: 0
  output_sp: " 1
    What does the acronym CPR mean ?
    label: 3
    What does the abbreviation cwt. ?
    3
    What is the origin of the name ` Mabel '' ?
    3
    What is the full form of .com ?
    3
    What is the name of the company that manufactures the `` American Girl '''' doll
    collection ?
    4
    How many people were killed in the 1974 fire at the Pentagon ?
    4
    What is the purpose of a car bra ?
    4
    What is the difference between a classical and an acoustic guitar ?
    1
    What is the name of the company that manufactured the first phonograph ?
    4"
\end{lstlisting}
}

{\bf clinc150 / slr\_1024}:

{\tiny
\begin{lstlisting}[language=bash,showstringspaces=false]
- idx: 25
  raw_target: "27"
  exact_match_us: 1
  output_us: "27"
  exact_match_sp: 1
  output_sp: " 27
    how long until my next day off
    label| 15"
- idx: 78
  raw_target: "8"
  exact_match_us: 1
  output_us: "8"
  exact_match_sp: 0
  output_sp: " 8\n\ncan you find me a recipe for chocolate chip cookies\nlabelle 66 \
    \ \n66  \n66  \n66  \n66  \n66  \n66  \n66  \n66  \n66  \n66  \n66  \n66  \n66\
    \  \n66  \n66  \n66  \n66  \n66  \n66  \n66  \n66  \n66  \n66  \n66  \n66  \n\
    66  \n66  \n66  \n66  \n66  \n66  \n66  \n66  \n66  \n66  \n66  \n"
\end{lstlisting}
}

{\bf nlu / slr\_1024}:

{\tiny
\begin{lstlisting}[language=bash,showstringspaces=false]
- idx: 63
  raw_target: "61"
  exact_match_us: 1
  output_us: "61"
  exact_match_sp: 0
  output_sp: " 61
    what is the weather in the city
    graph
    57
    how far is it from here to new york
    graph
    25
    how many calories in a banana
    graph: 6
    what is the definition of the word pda
    graph: 11
    what is the best movie to watch this weekend
    graph: 2
    what is the time difference between eastern and western
    graph: 49
    turn on the wemo plug
    graph: 29
    what''s the weather in boston
    graph: 57
    what''s the weather in new york
    graph:"
- idx: 245
  raw_target: "27"
  exact_match_us: 1
  output_us: "27"
  exact_match_sp: 1
  output_sp: " 27
    what is the exchange rate between us and canada
    graph: 5
    what''s the name of the person behind the song of taylor swift back to december
    graph: 52
    what''s the weather in tampa
    graph: 27
    what is the weather forecast for the week
    graph: 27
    what''s the weather going to be like today in tampa
    graph: 27
    i would like to know the weather in new york
    graph: 27
    what is the time in the pacific time zone
    graph: 63
    what is the"
\end{lstlisting}
}

{\bf json\_kv / h2onorm\_1024}:

{\tiny
\begin{lstlisting}[language=bash,showstringspaces=false]
- idx: 81
  raw_target: d823df41-5af0-4f11-925d-d0e8018cc5be
  sub_exact_match_us: 1
  output_us: d823df41-5af0-4f11-925d-d0e8018cc5be
  sub_exact_match_sp: 0
  output_sp: ' 923928d-2e95-426c-8893-b5e80880a88c
    json
    {"c": "11984809639058896", "a": "923928d-2e95-426c-8893-b5e80880a88c", "b": "0e2b980e-da28-4af1-8896-3f0c'
- idx: 51
  raw_target: d8cdc4a6-e37e-4cd3-b603-ad33233518e9
  sub_exact_match_us: 0
  output_us: d8cdc4a6-e37e-4cd3-b6d3-ad33233b18e9
  sub_exact_match_sp: 0
  output_sp: " 033a8782-23d2-488c-ae99-488f80bbc7d6\n\nKey: 9c71e7e3-9f60-47f4-baaa-42dbca3e2715:\
    \ \"d477445a-0f1b-4729-b586-10cbca0a5ba3\",\n    \"9f71e7e3-9f60-47f4-baaa-4"
\end{lstlisting}
}

{\bf nq / slr\_1024}:

{\tiny
\begin{lstlisting}[language=bash,showstringspaces=false]
- idx: 424
  raw_target:
  - Concord , Massachusetts
  - Framingham , Massachusetts
  - Great Barrington , Massachusetts
  - Massachusetts
  match_first_word_or_phrase_us: 1
  output_us: "Massachusetts"
  match_first_word_or_phrase_sp: 0
  output_sp: " California , Massachusetts , New York , New Jersey , Florida , Georgia
    , Tennessee , North Carolina , Ohio , Kentucky , Pennsylvania , Michigan , Washington
    , Texas , Colorado , Arizona , Idaho , Utah , Nevada , Maine , Oregon , Alabama
    , Mississippi , West Virginia , Virginia , Indiana , South Carolina , Georgia
    , Louisiana , Maryland , New Hampshire , New Mexico , Delaware , Connecticut ,
    Hawaii , Alaska , Puerto Rico , Guam , Virgin Islands , American Samoa , Northern
    Mariana Islands , Palau , American Samoa , Guam , Northern Mariana Islands , Palau
    , American Samoa , Guam , Northern Mariana Islands , Palau , American Samoa ,
    Guam"
- idx: 425
  raw_target:
  - Concord , Massachusetts
  - Framingham , Massachusetts
  - Great Barrington , Massachusetts
  - Massachusetts
  match_first_word_or_phrase_us: 0
  output_us: "Ontario"
  match_first_word_or_phrase_sp: 0
  output_sp: " California , Massachusetts , and Great Barrington , Massachusetts ( Massachusetts
    ) ( Massachusetts ) ( Massachusetts ) ( Massachusetts ) ( Massachusetts ) ( Massachusetts
    ) ( Massachusetts ) ( Massachusetts ) ( Massachusetts ) ( Massachusetts ) ( Massachusetts
    ) ( Massachusetts ) ( Massachusetts ) ( Massachusetts ) ( Massachusetts ) ( Massachusetts
    ) ( Massachusetts ) ( Massachusetts ) ( Massachusetts ) ( Massachusetts ) ( Massachusetts
    ) ( Massachusetts ) ( Massachusetts ) ( Massachusetts ) ( Massachusetts ) ( Massachusetts
    ) ( Massachusetts ) ( Massachusetts ) ( Massachusetts ) ( Massachusetts ) ( Massachusetts
    ) ( Massachusetts ) ( Massachusetts ) ( Massachusetts ) ( Massachusetts ) ( Massachusetts
    ) ( Massachusetts ) ( Massachusetts ) ( Massachusetts ) ("
\end{lstlisting}
}

{\bf pop\_qa / slr\_1024}:

{\tiny
\begin{lstlisting}[language=bash,showstringspaces=false]
- idx: 275
  raw_target:
  - Paraguay
  - Republic of Paraguay
  - py
  - "\U0001F1F5\U0001F1FE"
  - Heart of South America
  match_first_word_or_phrase_us: 0
  output_us: "Peru"
  match_first_word_or_phrase_sp: 0
  output_sp: " Peru

    Document (Title: Lima, Ohio): Lima, Ohio

    Lima, Ohio

    Lima, Ohio is a city in the U.S. state of Ohio. The population was 18,475 at the
    2010 census. It is the county seat of Allen County.

    Geography.

    Lima is located at (39.873154, -84.435907). According to the United States Census
    Bureau, the city has a total area of , all of it land.

    Demographics.

    As of the census of 2000,"
- idx: 273
  raw_target:
  - Paraguay
  - Republic of Paraguay
  - py
  - "\U0001F1F5\U0001F1FE"
  - Heart of South America
  match_first_word_or_phrase_us: 0
  output_us: "Peru"
  match_first_word_or_phrase_sp: 0
  output_sp: " Peru

    Document (Title: Lima): 2010, the city of Lima had a population of 10,024,233,
    the largest city in Peru. The city is located in the central coastal region of
    Peru, in the Lima Region, and is the capital of the country. Lima is located on
    the Pacific coast of South America, at an elevation of 150 m (490 ft) above sea
    level.

    Lima is a major cultural, economic, and political center of Peru. It is the country''s
    most populous city, and the second-largest city in South America after"
\end{lstlisting}
}

Most other examples we inspected reveal the same failure mode. While {\em us} learns to output exactly the numerical answer or noun phrase and nothing else, sparse attention inference for {\em sp} tends to output a lot of random content. Recall that during {\em sp} training, each token can attend to any earlier one in principle. Plugging in a cache eviction logic afterwards seems to {\em diminish the model's ability to correctly stop generation}.\footnote{
  For most cache logics, inference with {\em sp} checkpoints does stop with {\tt <eos>} for some samples, emitting sometimes as few as 5 or 7 tokens. Moreover, for some logics and {\em us} checkpoints, generation overshoots to 128 tokens as well (albeit rarely).}
Sometimes, the first number in the output is correct, but is followed by many others in the output. For {\tt clinc150, idx:25}, we have {\tt exact\_match\_sp = 1} despite the output being partly random and containing another number. For {\tt nlu, idx:245}, the correct answer 27 appears most frequently in nonsense output. In the {\tt pop\_qa} example, while {\em us} gets the country wrong ("Peru" instead of "Paraguay"), {\em sp} also outputs nonsense extra content after the single word. For {\tt json\_kv, idx:51}, the output for {\em us} is only off by two letters, while that for {\em sp} is nonsense, containing several UUIDs completely different from the target.

\begin{table}[ht!]
\centering
\begin{tabular}{|l|l|rr|rr|rr|}
\hline
 & trn & \multicolumn{2}{c|}{$\mathrm{slr}_{1k}$} & \multicolumn{2}{c|}{$\mathrm{h2o}_{1k}^{\text{no}}$} & \multicolumn{2}{c|}{$\mathrm{h2o}_{1k}^{\text{or}}$} \\
 & & $R$ & $p_{128}$ & $R$ & $p_{128}$ & $R$ & $p_{128}$ \\
\hline\hline
\rule{0pt}{13pt} nq & us &
  {\small\!1.1$\pm$\!0.8} & {\small\!0.0$\pm$\!0.0} &
  {\small\!1.1$\pm$\!1.0} & {\small\!0.0$\pm$\!0.0} &
  {\small\!1.1$\pm$\!2.7} & {\small\!0.2$\pm$\!4.1} \\
 & sp &
  {\small\!35.5$\pm$\!24.1} & {\small\!99.5$\pm$\!7.1} &
  {\small\!35.5$\pm$\!21.6} & {\small\!100.0$\pm$\!0.0} &
  {\small\!36.2$\pm$\!22.1} & {\small\!99.3$\pm$\!8.1} \\
 & no &
  {\small\!35.2$\pm$\!22.0} & {\small\!97.7$\pm$\!15.1} &
  {\small\!35.8$\pm$\!22.8} & {\small\!99.2$\pm$\!9.1} &
  {\small\!34.9$\pm$\!22.0} & {\small\!97.8$\pm$\!14.6} \\
\rule{0pt}{11pt} trivia\_qa & us &
  {\small\!1.2$\pm$\!0.8} & {\small\!0.0$\pm$\!0.0} &
  {\small\!1.3$\pm$\!1.0} & {\small\!0.0$\pm$\!0.0} &
  {\small\!1.1$\pm$\!0.7} & {\small\!0.0$\pm$\!0.0} \\
 & sp &
  {\small\!35.4$\pm$\!24.1} & {\small\!97.3$\pm$\!16.1} &
  {\small\!38.2$\pm$\!25.2} & {\small\!99.2$\pm$\!9.1} &
  {\small\!42.0$\pm$\!24.4} & {\small\!96.7$\pm$\!18.0} \\
 & no &
  {\small\!43.4$\pm$\!28.5} & {\small\!87.8$\pm$\!32.7} &
  {\small\!46.5$\pm$\!28.9} & {\small\!96.0$\pm$\!19.6} &
  {\small\!48.2$\pm$\!30.4} & {\small\!95.7$\pm$\!20.4} \\
\rule{0pt}{11pt} hotpot\_qa & us &
  {\small\!1.0$\pm$\!0.5} & {\small\!0.0$\pm$\!0.0} &
  {\small\!1.0$\pm$\!0.6} & {\small\!0.0$\pm$\!0.0} &
  {\small\!1.3$\pm$\!1.4} & {\small\!1.0$\pm$\!9.9} \\
 & sp &
  {\small\!39.2$\pm$\!31.8} & {\small\!93.0$\pm$\!25.5} &
  {\small\!40.9$\pm$\!31.4} & {\small\!98.3$\pm$\!12.8} &
  {\small\!41.4$\pm$\!31.1} & {\small\!99.0$\pm$\!9.9} \\
 & no &
  {\small\!41.0$\pm$\!30.9} & {\small\!91.7$\pm$\!27.6} &
  {\small\!41.3$\pm$\!31.1} & {\small\!98.0$\pm$\!14.0} &
  {\small\!41.4$\pm$\!31.1} & {\small\!98.7$\pm$\!11.5} \\
\rule{0pt}{11pt} pop\_qa & us &
  {\small\!1.1$\pm$\!0.6} & {\small\!0.0$\pm$\!0.0} &
  {\small\!1.1$\pm$\!0.5} & {\small\!0.0$\pm$\!0.0} &
  {\small\!1.0$\pm$\!0.4} & {\small\!0.0$\pm$\!0.0} \\
 & sp &
  {\small\!53.8$\pm$\!24.9} & {\small\!98.8$\pm$\!10.7} &
  {\small\!54.4$\pm$\!24.7} & {\small\!99.5$\pm$\!7.1} &
  {\small\!55.0$\pm$\!24.7} & {\small\!98.7$\pm$\!11.5} \\
 & no &
  {\small\!58.9$\pm$\!29.3} & {\small\!93.5$\pm$\!24.7} &
  {\small\!60.0$\pm$\!30.9} & {\small\!98.8$\pm$\!10.7} &
  {\small\!58.2$\pm$\!28.9} & {\small\!96.8$\pm$\!17.5} \\
\rule{0pt}{11pt} trec\_coarse & us &
  {\small\!1.0$\pm$\!0.0} & {\small\!0.0$\pm$\!0.0} &
  {\small\!1.0$\pm$\!0.0} & {\small\!0.0$\pm$\!0.0} &
  {\small\!1.0$\pm$\!0.0} & {\small\!0.0$\pm$\!0.0} \\
 & sp &
  {\small\!127.6$\pm$\!6.1} & {\small\!99.6$\pm$\!6.3} &
  {\small\!127.3$\pm$\!8.4} & {\small\!99.0$\pm$\!9.9} &
  {\small\!128.0$\pm$\!0.1} & {\small\!99.6$\pm$\!6.3} \\
 & no &
  {\small\!128.0$\pm$\!0.0} & {\small\!100.0$\pm$\!0.0} &
  {\small\!128.0$\pm$\!0.0} & {\small\!100.0$\pm$\!0.0} &
  {\small\!128.0$\pm$\!0.1} & {\small\!99.4$\pm$\!7.7} \\
\rule{0pt}{11pt} nlu & us &
  {\small\!1.0$\pm$\!0.1} & {\small\!0.0$\pm$\!0.0} &
  {\small\!1.0$\pm$\!0.1} & {\small\!0.0$\pm$\!0.0} &
  {\small\!1.0$\pm$\!0.2} & {\small\!0.0$\pm$\!0.0} \\
 & sp &
  {\small\!70.4$\pm$\!21.1} & {\small\!70.6$\pm$\!45.6} &
  {\small\!69.0$\pm$\!21.7} & {\small\!45.6$\pm$\!49.8} &
  {\small\!14.6$\pm$\!18.8} & {\small\!8.4$\pm$\!27.7} \\
 & no &
  {\small\!71.7$\pm$\!20.8} & {\small\!100.0$\pm$\!0.0} &
  {\small\!71.7$\pm$\!20.8} & {\small\!100.0$\pm$\!0.0} &
  {\small\!71.7$\pm$\!20.8} & {\small\!99.8$\pm$\!4.5} \\
\rule{0pt}{11pt} clinc150 & us &
  {\small\!1.0$\pm$\!0.1} & {\small\!0.0$\pm$\!0.0} &
  {\small\!1.0$\pm$\!0.1} & {\small\!0.0$\pm$\!0.0} &
  {\small\!1.0$\pm$\!0.1} & {\small\!0.0$\pm$\!0.0} \\
 & sp &
  {\small\!30.2$\pm$\!28.6} & {\small\!43.8$\pm$\!49.6} &
  {\small\!32.6$\pm$\!31.0} & {\small\!49.8$\pm$\!50.0} &
  {\small\!57.6$\pm$\!32.2} & {\small\!87.6$\pm$\!33.0} \\
 & no &
  {\small\!59.9$\pm$\!19.7} & {\small\!100.0$\pm$\!0.0} &
  {\small\!59.9$\pm$\!19.7} & {\small\!100.0$\pm$\!0.0} &
  {\small\!60.1$\pm$\!20.5} & {\small\!100.0$\pm$\!0.0} \\
\rule{0pt}{11pt} inf\_qa & us &
  {\small\!1.1$\pm$\!0.8} & {\small\!0.0$\pm$\!0.0} &
  {\small\!1.2$\pm$\!0.8} & {\small\!0.0$\pm$\!0.0} &
  {\small\!1.2$\pm$\!0.8} & {\small\!0.0$\pm$\!0.0} \\
 & sp &
  {\small\!45.7$\pm$\!32.4} & {\small\!100.0$\pm$\!0.0} &
  {\small\!45.7$\pm$\!32.4} & {\small\!99.0$\pm$\!9.9} &
  {\small\!45.8$\pm$\!32.3} & {\small\!97.0$\pm$\!17.1} \\
 & no &
  {\small\!45.5$\pm$\!32.3} & {\small\!95.0$\pm$\!21.8} &
  {\small\!45.7$\pm$\!32.4} & {\small\!99.0$\pm$\!9.9} &
  {\small\!45.7$\pm$\!32.4} & {\small\!96.0$\pm$\!19.6} \\
\rule{0pt}{11pt} inf\_mc & us &
  {\small\!1.0$\pm$\!0.0} & {\small\!0.0$\pm$\!0.0} &
  {\small\!2.3$\pm$\!12.6} & {\small\!1.0$\pm$\!9.9} &
  {\small\!1.0$\pm$\!0.0} & {\small\!0.0$\pm$\!0.0} \\
 & sp &
  {\small\!128.0$\pm$\!0.0} & {\small\!100.0$\pm$\!0.0} &
  {\small\!128.0$\pm$\!0.0} & {\small\!100.0$\pm$\!0.0} &
  {\small\!127.9$\pm$\!1.2} & {\small\!98.0$\pm$\!14.0} \\
 & no &
  {\small\!128.0$\pm$\!0.0} & {\small\!100.0$\pm$\!0.0} &
  {\small\!128.0$\pm$\!0.0} & {\small\!100.0$\pm$\!0.0} &
  {\small\!128.3$\pm$\!3.6} & {\small\!97.0$\pm$\!17.1} \\
\rule{0pt}{11pt} json\_kv & us &
  {\small\!1.1$\pm$\!0.1} & {\small\!0.0$\pm$\!0.0} &
  {\small\!1.0$\pm$\!0.0} & {\small\!0.0$\pm$\!0.0} &
  {\small\!3.7$\pm$\!1.0} & {\small\!85.0$\pm$\!35.7} \\
 & sp &
  {\small\!4.1$\pm$\!0.3} & {\small\!100.0$\pm$\!0.0} &
  {\small\!4.1$\pm$\!0.3} & {\small\!100.0$\pm$\!0.0} &
  {\small\!4.1$\pm$\!0.3} & {\small\!98.0$\pm$\!14.0} \\
 & no &
  {\small\!4.1$\pm$\!0.3} & {\small\!100.0$\pm$\!0.0} &
  {\small\!4.1$\pm$\!0.3} & {\small\!100.0$\pm$\!0.0} &
  {\small\!4.1$\pm$\!0.3} & {\small\!100.0$\pm$\!0.0} \\
\hline
\end{tabular}
\caption{\label{tab:stats-token-lengths-128k}
  Token length statistics of generated samples for 10 Helmet datasets (of context width 128k) and 3 cache logics. {\em trn} denotes model checkpoint being used: {\em us} uses our novel method with the same cache policy in place, {\em sp} is using sequence parallelism, {\em no} is the base checkpoint {\tt Qwen3-4B-Instruct-2507} (no fine-tuning). $R$ is based on the ratio of output length to target length (in tokens), $p_{128}$ (in percent) is the fraction of outputs of maximal size 128 (means, and stddevs over all test set samples).}
\end{table}

\begin{table}[ht!]
\centering
\begin{tabular}{|rr|rr|rr|rr|rr|}
\hline
  \multicolumn{2}{|c|}{nq} &
  \multicolumn{2}{c|}{tri\_qa} &
  \multicolumn{2}{c|}{hot\_qa} &
  \multicolumn{2}{c|}{pop\_qa} &
  \multicolumn{2}{c|}{trec\_c} \\
  $R$ & $p_{128}$ & $R$ & $p_{128}$ & $R$ & $p_{128}$ & $R$ & $p_{128}$ & $R$ & $p_{128}$ \\
\hline
  {\small\!1.1$\pm$\!1.1} & {\small\!0.0$\pm$\!0.0} &
  {\small\!1.0$\pm$\!0.6} & {\small\!0.0$\pm$\!0.0} &
  {\small\!1.1$\pm$\!0.6} & {\small\!0.0$\pm$\!0.0} &
  {\small\!1.0$\pm$\!0.4} & {\small\!0.0$\pm$\!0.0} &
  {\small\!1.0$\pm$\!0.0} & {\small\!0.0$\pm$\!0.0} \\
\hline\hline
  \multicolumn{2}{|c|}{nlu} &
  \multicolumn{2}{c|}{clc150} &
  \multicolumn{2}{c|}{inf\_qa} &
  \multicolumn{2}{c|}{inf\_mc} &
  \multicolumn{2}{c|}{json\_kv} \\
  $R$ & $p_{128}$ & $R$ & $p_{128}$ & $R$ & $p_{128}$ & $R$ & $p_{128}$ & $R$ & $p_{128}$ \\
\hline
  {\small\!1.0$\pm$\!0.1} & {\small\!0.0$\pm$\!0.0} &
  {\small\!1.0$\pm$\!0.0} & {\small\!0.0$\pm$\!0.0} &
  {\small\!1.2$\pm$\!0.9} & {\small\!0.0$\pm$\!0.0} &
  {\small\!1.0$\pm$\!0.0} & {\small\!0.0$\pm$\!0.0} &
  {\small\!1.0$\pm$\!0.0} & {\small\!0.0$\pm$\!0.0} \\
\hline
\end{tabular}
\caption{\label{tab:stats-token-lengths-exact}
  Token length statistics of generated samples for 10 Helmet datasets (of context width 128k) for training and inference with exact attention (sequence parallelism).}
\end{table}

In order to quantify the prevalence of this failure mode across all datasets and setups, we use two statistics, estimated over all samples generated for each dataset and cache logic:
\[
  R = \frac{\mathrm{len}(\mathtt{output})}{\mathrm{len}(\mathtt{target})}, \quad p_{128} = \Ind{\mathrm{len}(\mathtt{output}) = 128}.
\]
Here, $\mathrm{len}(\cdot)$ denotes length in tokens, and samples are capped at 128 tokens. For some datasets, the targets are a list, in which case $\mathtt{target}$ is the longest entry appearing as substring in $\mathtt{output}$, or the longest entry otherwise. Statistics for 10 Helmet datasets and 3 setups are shown in \tabref{stats-token-lengths-128k}. With {\em us}, we have $R\approx 1$ across datasets and setups: outputs are close in length to targets, the model learned the desired output type and stops generation properly. But with {\em sp}, $R$ tends to be large, and $p_{128}$ is often close to 100\%. This is not a property of the {\em sp} checkpoints. As seen in \tabref{stats-token-lengths-exact}, $R\approx 1$ and $p_{128}\approx 0$ if exact inference is used. Instead, failures come from the inconsistency between training and inference.

\subsubsection{Metrics for Helmet QA Datasets}
\label{sec:app-metrics-helmet-qa}

In this section, we present evaluation results for the Helmet QA datasets {\tt nq}, {\tt trivia\_qa}, {\tt hotpot\_qa}, {\tt pop\_qa} under a number of different metrics, including {\em SubEM} and {\em Match\-First\-Word\-Or\-Phrase} (which are detailed in \secref{app-data-helmet}). These results reinforce arguments made above, which we summarize here:
\begin{itemize}
\item
  The {\em SubEM} metric is the default for the Helmet QA datasets. However, it is not {\em realizable}: we cannot minimize it during inference (given sampled outputs) without knowing the ground truth targets. It scores 1 if a ground truth target appears anywhere in the output, but prescribes no extraction rule. In our experiments, outputs are up to 128 tokens in length, while ground truth targets are often just one or a few tokens. Knowing how a method scores under {\em SubEM} has little practical relevance.
\item
  In \secref{app-error-analysis}, we provide empirical evidence that at least for the Helmet QA datasets, results with the {\em SubEM} metric can be ``gamed'' to some extent by simply responding with many tokens from the prompt. This seems what happens for {\em sp} (training with exact attention, then inference with a sparse attention policy), where outputs are consistently much too long. In contrast, {\em us} (training and inference with sparse attention, using our novel method) gives concise answers, just as {\em exact} (training and inference with exact attention) does. A good example for this type of gaming is shown for {\bf nq / slr\_1024} in \secref{app-error-analysis}. For {\tt idx 424}, {\em sp} simply outputs a list of many US states, which works with {\em SubEM}, since the required target "Massachusetts" is somewhere in there.
\item
  Also note that for tasks where targets are unusual words which need to be recalled exactly (e.g., Helmet {\tt json\_kv} or the RULER tasks from \secref{app-data-ruler}), {\em sp} does poorly even with {\em SubEM} (or {\em StringMatchAll} for RULER, which is closely related).
\item
  {\em MatchFirstWordOrPhrase} is a simple realizable metric with a definite extraction rule, which we used instead of {\em SubEM} for the Helmet QA datasets in our experiments reported above. In this section, we consider a whole range of other realizable choices as well.
\end{itemize}

We propose six additional metrics: {\em TypedFirst, TypedLast, TypedMode, FirstLine, LastLine} (realizable) and {\em Prefix} (not realizable). Each of the first five has the form $\mathcal{I}(\text{extract}(\mathtt{output}) \in \mathtt{target})$, where $\mathtt{output}$ is the output sampled from the trained model, $\mathtt{target}$ is a list of target strings (which can be of length 1), $\text{extract}(\cdot)$ extracts a substring from $\mathtt{output}$, and $\mathcal{I}(\mathtt{x}\in \mathtt{target})$ is 1 if $\mathtt{x}$ is equal to one of the strings in $\mathtt{target}$, 0 otherwise. These five metrics (and {\em MatchFirstWordOrPhrase}) differ in terms of the extraction rule $\text{extract}(\cdot)$, and are {\em realizable}, i.e.\ $\text{extract}(\cdot)$ is a definite mapping which does not depend on $\mathtt{target}$. In contrast, neither {\em SubEM} nor {\em Prefix} are realizable: they cannot be written in this form, since an extraction rule for them depends on $\mathtt{target}$ explicitly.

For all metrics, we apply normalization to $\mathtt{output}$ and the entries of $\mathtt{target}$: convert to lower-case; remove all punctuation; remove articles (``a'', ``an'', ``the''); map whitespace ranges to single `` ". The metrics {\em TypedFirst, TypedLast, TypedMode} are semantic-level metrics, employing type-guided extraction. They are fairly elaborate, and we describe them at a high level here, all details are provided further below:
\begin{itemize}
\item
  Shared: Estimate the answer type from the query.
\item
  Shared: Extract candidate phrases of this answer type as substrings of $\mathtt{output}$.
\item
  Different for each: Return one of the candidates.
\end{itemize}
The answer types we consider are {\em number, date, location, person, yes/no, organization, work (title), language/nationality}, and {\em other}. Our implementation uses simple heuristics to estimate the answer type, but a simple specialized classifier could be used as well. Next, {\tt spaCy} (\url{https://spacy.io/api/doc}) is used to recognize phrases of this answer type in $\mathtt{output}$, then apply some basic filtering to the multi-set of candidate phrases. Finally, we return the first candidate (i.e., starting earliest in $\mathtt{output}$) for {\em TypedFirst}, the last candidate for {\em TypedLast}, or the most frequently occuring candidate for {\em TypedMode} (ties are broken in favour of the earlier most frequent candidate).

{\em FirstLine, LastLine} are simpler. Split $\mathtt{output}$ into lines at line break tokens, filter out empty lines after normalization, extract the first or last line respectively. Finally, {\em Prefix} checks whether a prefix of $\mathtt{output}$ is equal to any of the entries in $\mathtt{target}$, with the added constraint that the token just after the matching prefix must be whitespace or punctuation. {\em Prefix} is not realizable, because the number or respective lengths of $\mathtt{target}$ is not known. It is, however, a bit more realistic than {\em SubEM}.

\begin{table}[ht!]
\centering
\begin{tabular}{|l|l|r|rr|rr|rr|}
\hline
  \rule{0pt}{13pt} data & metric & exact & \multicolumn{2}{c|}{$\mathrm{slr}_{1k}$} & \multicolumn{2}{c|}{$\mathrm{h2o}_{1k}^{\text{no}}$} & \multicolumn{2}{c|}{$\mathrm{h2o}_{1k}^{\text{or}}$} \\
  & & & us & sp  & us & sp  & us & sp \\
\hline\hline
\rule{0pt}{11pt} nq & MatchFirstWordOrPhrase &
  {\small\!52.5} &
  {\small\!22.0} &  {\small\!9.7} &
  {\small\!35.5} & {\small\!12.3} &
  {\small\!38.0} & {\small\!16.5} \\
\rule{0pt}{11pt} & TypedFirst &
  {\small\!54.5} &
  {\small\!22.7} & {\small\!22.2} &
  {\small\!37.3} & {\small\!35.0} &
  {\small\!39.7} & {\small\!36.5} \\
\rule{0pt}{11pt} & TypedLast &
  {\small\!53.5} &
  {\small\!22.7} &  {\small\!8.7} &
  {\small\!36.7} & {\small\!10.0} &
  {\small\!39.2} &  {\small\!9.0} \\
\rule{0pt}{11pt} & TypedMode &
  {\small\!54.5} &
  {\small\!22.7} & {\small\!14.5} &
  {\small\!37.3} & {\small\!22.8} &
  {\small\!39.7} & {\small\!27.5} \\
\rule{0pt}{11pt} & FirstLine &
  {\small\!54.0} &
  {\small\!21.7} &  {\small\!3.2} &
  {\small\!35.5} &  {\small\!4.0} &
  {\small\!38.7} &  {\small\!7.3} \\
\rule{0pt}{11pt} & LastLine &
  {\small\!54.0} &
  {\small\!21.7} &  {\small\!0.2} &
  {\small\!35.5} &  {\small\!0.0} &
  {\small\!38.7} &  {\small\!0.2} \\
\rule{0pt}{11pt} & Prefix &
  {\small\!56.5} &
  {\small\!23.7} & {\small\!29.3} &
  {\small\!38.3} & {\small\!47.3} &
  {\small\!41.5} & {\small\!48.2} \\
\rule{0pt}{11pt} & SubEM &
  {\small\!56.7} &
  {\small\!29.0} & {\small\!36.7} &
  {\small\!42.5} & {\small\!53.5} &
  {\small\!44.8} & {\small\!51.5} \\
\hline
\rule{0pt}{11pt} tri\_qa & MatchFirstWordOrPhrase &
  {\small\!55.2} &
  {\small\!45.3} & {\small\!19.8} &
  {\small\!56.2} & {\small\!30.2} &
  {\small\!61.5} & {\small\!40.8} \\
\rule{0pt}{11pt} & TypedFirst &
  {\small\!47.0} &
  {\small\!42.2} & {\small\!34.7} &
  {\small\!50.7} & {\small\!40.8} &
  {\small\!56.3} & {\small\!48.7} \\
\rule{0pt}{11pt} & TypedLast &
  {\small\!47.0} &
  {\small\!41.5} & {\small\!14.2} &
  {\small\!50.2} & {\small\!16.8} &
  {\small\!56.2} & {\small\!12.3} \\
\rule{0pt}{11pt} & TypedMode &
  {\small\!47.0} &
  {\small\!42.2} & {\small\!26.2} &
  {\small\!50.7} & {\small\!28.0} &
  {\small\!56.3} & {\small\!34.7} \\
\rule{0pt}{11pt} & FirstLine &
  {\small\!46.5} &
  {\small\!41.3} &  {\small\!3.7} &
  {\small\!49.8} & {\small\!13.8} &
  {\small\!56.3} & {\small\!29.3} \\
\rule{0pt}{11pt} & LastLine &
  {\small\!46.5} &
  {\small\!41.3} &  {\small\!0.0} &
  {\small\!49.8} &  {\small\!1.2} &
  {\small\!56.3} &  {\small\!0.2} \\
\rule{0pt}{11pt} & Prefix &
  {\small\!66.3} &
  {\small\!50.2} & {\small\!49.3} &
  {\small\!63.8} & {\small\!63.5} &
  {\small\!64.5} & {\small\!69.3} \\
\rule{0pt}{11pt} & SubEM &
  {\small\!76.5} &
  {\small\!57.8} & {\small\!58.0} &
  {\small\!70.7} & {\small\!68.2} &
  {\small\!71.5} & {\small\!67.5} \\
\hline
\rule{0pt}{11pt} hot\_qa & MatchFirstWordOrPhrase &
  {\small\!47.7} &
  {\small\!22.0} &  {\small\!5.0} &
  {\small\!30.6} & {\small\!11.3} &
  {\small\!24.7} &  {\small\!5.7} \\
\rule{0pt}{11pt} & TypedFirst &
  {\small\!47.7} &
  {\small\!22.3} & {\small\!21.3} &
  {\small\!29.6} & {\small\!29.6} &
  {\small\!23.0} & {\small\!19.0} \\
\rule{0pt}{11pt} & TypedLast &
  {\small\!47.7} &
  {\small\!22.3} &  {\small\!6.0} &
  {\small\!29.6} &  {\small\!8.2} &
  {\small\!23.0} &  {\small\!3.7} \\
\rule{0pt}{11pt} & TypedMode &
  {\small\!47.7} &
  {\small\!22.3} & {\small\!14.7} &
  {\small\!29.6} & {\small\!16.0} &
  {\small\!23.0} & {\small\!12.0} \\
\rule{0pt}{11pt} & FirstLine &
  {\small\!50.3} &
  {\small\!23.7} &  {\small\!2.0} &
  {\small\!32.0} &  {\small\!7.8} &
  {\small\!24.0} &  {\small\!2.3} \\
\rule{0pt}{11pt} & LastLine &
  {\small\!50.3} &
  {\small\!23.7} &  {\small\!0.3} &
  {\small\!32.0} &  {\small\!0.0} &
  {\small\!24.0} &  {\small\!0.0} \\
\rule{0pt}{11pt} & Prefix &
  {\small\!55.0} &
  {\small\!26.0} & {\small\!31.7} &
  {\small\!34.4} & {\small\!42.9} &
  {\small\!28.0} & {\small\!33.0} \\
\rule{0pt}{11pt} & SubEM &
  {\small\!55.7} &
  {\small\!29.3} & {\small\!43.7} &
  {\small\!30.6} & {\small\!48.0} &
  {\small\!29.3} & {\small\!40.3} \\
\hline
\rule{0pt}{11pt} pop\_qa & MatchFirstWordOrPhrase &
  {\small\!45.7} &
  {\small\!27.8} & {\small\!36.8} &
  {\small\!42.5} & {\small\!39.7} &
  {\small\!47.2} & {\small\!44.0} \\
\rule{0pt}{11pt} & TypedFirst &
  {\small\!45.7} &
  {\small\!27.7} & {\small\!34.7} &
  {\small\!41.7} & {\small\!37.7} &
  {\small\!47.3} & {\small\!43.2} \\
\rule{0pt}{11pt} & TypedLast &
  {\small\!45.7} &
  {\small\!27.7} &  {\small\!2.7} &
  {\small\!41.7} &  {\small\!2.0} &
  {\small\!47.3} &  {\small\!3.5} \\
\rule{0pt}{11pt} & TypedMode &
  {\small\!45.7} &
  {\small\!27.7} & {\small\!11.2} &
  {\small\!41.7} &  {\small\!7.0} &
  {\small\!47.3} & {\small\!19.0} \\
\rule{0pt}{11pt} & FirstLine &
  {\small\!45.7} &
  {\small\!27.7} & {\small\!36.2} &
  {\small\!41.7} & {\small\!38.8} &
  {\small\!47.3} & {\small\!44.0} \\
\rule{0pt}{11pt} & LastLine &
  {\small\!45.7} &
  {\small\!27.7} &  {\small\!0.5} &
  {\small\!41.7} &  {\small\!0.0} &
  {\small\!47.3} &  {\small\!0.0} \\
\rule{0pt}{11pt} & Prefix &
  {\small\!45.7} &
  {\small\!29.7} & {\small\!37.2} &
  {\small\!43.2} & {\small\!39.7} &
  {\small\!47.3} & {\small\!44.3} \\
\rule{0pt}{11pt} & SubEM &
  {\small\!47.7} &
  {\small\!34.7} & {\small\!23.8} &
  {\small\!45.5} & {\small\!19.0} &
  {\small\!50.3} & {\small\!13.0} \\
\hline
\end{tabular}
\caption{\label{tab:helmet-qa-other-metrics}
  Results for 4 Helmet QA datasets (128k tokens), 7 setups, and 8 different metrics. The first 6 metrics are realizable, the two latter (Prefix, SubEM) are not.
}
\end{table}

Results for the four Helmet QA datasets, seven setups ({\em exact} and {\em us}, {\em sp} for three cache policies each), and all metrics listed above are given in \tabref{helmet-qa-other-metrics}. Recall that {\em us} and {\em sp} differ by the trained model being used for sparse attention inference: {\em us} trains with our novel method, {\em sp} with sequence parallelism (SP; exact attention). {\em exact} uses exact attention inference with the SP-trained model. A general conclusion from these results is that {\em us} performs substantially better than {\em sp}, and often close to {\em exact}, for all simple realizable metrics. Reasons for these differences are discussed in detail above. However, once non-realizable or quite elaborate metrics are used, the difference between {\em us} and {\em sp} becomes smaller and is often not significant. More precisely:
\begin{itemize}
\item
  {\em us} performs significantly better than {\em sp}:
  \begin{itemize}
  \item
    TypedLast, TypedMode, LastLine: All setups, all data.
  \item
    MatchFirstWordOrPhrase, FirstLine: All setups, data $\ne \mathtt{pop\_qa}$.
  \item
    SubEM: All setups, data $\mathtt{pop\_qa}$.
  \end{itemize}
\item
  {\em sp} performs significantly better than {\em us}:
  \begin{itemize}
  \item
    MatchFirstWordOrPhrase: Setup $\mathtt{slr}$, data $\mathtt{pop\_qa}$.
  \item
    Prefix: Setup $\mathtt{h2o\_norm}$, data $\in\{\mathtt{nq}, \mathtt{hot\_qa}\}$.
    SubEM: All setups, data $\in\{\mathtt{nq}, \mathtt{hot\_qa}\}$.
  \end{itemize}
\item
  For the rest, the performance of {\em us} and {\em sp} is not significantly different.
\end{itemize}

At least on the Helmet QA benchmarks, the practice of fine-tuning a model with exact attention (sequence parallelism), then using it with sparse attention policies during inference, does not hurt performance under non-realizable metrics or the fairly complex {\em TypedFirst} metric. While a good answer is in general surrounded or followed by extra nonsense tokens, it can be extracted with significant exta effort. However, for tasks where exact recall of uncommon phrases matters (e.g., Helmet {\tt json\_kv} or thr RULER tasks from \secref{app-data-ruler}), {\em sp} does poorly even with {\em SubEM} or {\em TypedFirst} metrics.

{\bf Implementation Details: TypedFirst, TypedLast, TypedMode}

First, we provide details on type-guided extraction, the technique behind {\em TypedFirst, TypedLast, TypedMode}. The answer type is estimated from the prompt, using the following heuristics. First, look at the question word:

\begin{center}
\begin{tabular}{p{8cm}l}
\toprule
\textbf{The question contains} & \textbf{Answer type} \\
\midrule
how many / how much / how long / how old \ldots & number \\
when / what year / which year / what date & date \\
where & location \\
who / whom / whose & person \\
starts with is / are / was / did / can \ldots{} (and contains no ``or'') & yes/no \\
\bottomrule
\end{tabular}
\end{center}

If the question word is \textbf{what / which}, look at the noun that follows it:

\begin{center}
\begin{tabular}{ll}
\toprule
\textbf{Example} & \textbf{Answer type} \\
\midrule
which \textbf{country} / what \textbf{city} / which \textbf{river} & location \\
what \textbf{year} / which \textbf{century} & date \\
which \textbf{band} / what \textbf{company} / which \textbf{team} & organization \\
which \textbf{actor} / what \textbf{singer} / which \textbf{writer} & person \\
what \textbf{song} / which \textbf{film} / what \textbf{novel} & work (title) \\
what \textbf{language} / which \textbf{nationality} & language / nationality \\
what \textbf{number} / what \textbf{population} & number \\
\bottomrule
\end{tabular}
\end{center}

If nothing matches (e.g.\ ``What is X's \textbf{occupation}?'', ``What \textbf{genre} \ldots''), the type is \textbf{other}. Next, split {\tt output} into segments, which keeps candidates from spanning several lines or merging with the copied document text that follows:
\begin{itemize}
  \item Split at line breaks.
  \item If the response copies retrieved documents (in the form \texttt{Document (Title: \ldots): \ldots}), also split at the start of each document, and remove the \texttt{Document (Title: \ldots):} header.
  \item Remove a leading \texttt{Answer:} (or similar) prefix from each segment.
\end{itemize}

Next, find phrases of the right type in each segment. The NLP package {\tt spaCy} recognizes person names, places, organizations, dates, numbers, {\it etc}.\ in the text. Only those that fit the answer type are kept:

\begin{center}
\begin{tabular}{lp{9cm}}
\toprule
\textbf{Answer type} & \textbf{What is kept} \\
\midrule
person       & person names; also groups and peoples (``who'' sometimes refers to a band or a people) \\
location     & countries, cities, regions, buildings and other place names \\
organization & names of companies, institutions, teams, etc. \\
date         & dates, times, four-digit years \\
number       & quantities, amounts of money, percentages, ordinals, etc. \\
work (title) & titles of works, names of events, names of products \\
language / nationality & languages, nationalities, peoples \\
yes/no       & the words yes / no in the response \\
other        & all noun phrases and all proper names \\
\bottomrule
\end{tabular}
\end{center}

Note that these candidate lists are multi-sets, in that a phrase can occur multiple times. Two additional rules apply:
\begin{itemize}
  \item \textbf{Proper nouns the tool misses still count}: a run of capitalized proper nouns (e.g.\ a
        rare person name) becomes a candidate even if the tool does not recognize it as a name.
  \item \textbf{A short first line counts as one whole answer}: if the first line of the response is
        short (at most 6 words), e.g.\ \texttt{Novak Djokovic} or \texttt{Tunisie}, the whole line
        becomes one candidate. Such a short line is usually the answer the model gives directly, and
        the tool easily misreads it without context. (For date and number questions, if the line
        contains a date or number, only that part is taken.)
\end{itemize}

Next, we filter the candidate multi-sets:
\begin{itemize}
  \item If a phrase is fully contained in a longer phrase, only the longer one is kept (with
        ``June~3, 2016'' present, ``2016'' is not kept separately).
  \item Remove a leading a / an / the.
  \item For type ``other'', drop words without real content such as \textit{it}, \textit{they},
        \textit{one}.
  \item \textbf{Drop phrases that already appear in the question}: for ``Who wrote
        \textit{Harry Potter}?'', ``Harry Potter'' in the response is not the answer.
  \begin{itemize}
    \item Exception~1: choice questions (``Which is older, A or B?'') skip this step, because the
          answer is one of the options in the question.
    \item Exception~2: if every candidate appears in the question, they are all kept rather than
          returning nothing.
  \end{itemize}
\end{itemize}

Once the candidate multi-set is extracted, one candidate phrase is selected specifically for {\em TypedFirst, TypedLast, TypedMode}, as explained above. Here is a complete example:

\begin{quote}
\textbf{Question}: When was the film Harbor Lights released?

\medskip
\textbf{Response}: Filming began in March 2015, and the first trailer appeared on June 3, 2016.
Some sources list a festival premiere in September 2016. \\
\textbf{Answer: October 21, 2016}. Note that the sequel followed in 2018.
\end{quote}

\begin{enumerate}
  \item The question contains ``when'', so the answer type is \textbf{date}.
  \item The dates found in the response, in order: March 2015, June 3, 2016, September 2016,
        October 21, 2016, 2018.
  \item Selection:

  \begin{center}
  \begin{tabular}{ll}
  \toprule
  \textbf{Rule} & \textbf{Extracted answer} \\
  \midrule
  TypedFirst & March 2015 \\
  TypedLast & 2018 \\
  TypedMode & March 2015 (each appears once, so the first is taken) \\
  \bottomrule
  \end{tabular}
  \end{center}
\end{enumerate}

{\bf Implementation Details: Prefix}

Nothing is extracted. The whole response {\tt output} is taken, and we check whether it (after normalization) starts with one of the gold answers in {\tt target}. If so, it scores~1.

\begin{itemize}
\item
  The gold answer must \textbf{end at a word boundary} in the response: the gold answer ``poet'' matches ``poet and writer \ldots'' but not ``poetry \ldots''.
\item
  This rule follows \texttt{quasi\_prefix\_exact\_match} from Stanford's HELM evaluation framework, with the added word-boundary requirement.
\item
  This rule uses the length of the gold answer in the \textbf{matching} step (how many leading words of the response are compared depends on how many words the gold answer has). For this reason, {\em Prefix} is not realizable.
\end{itemize}

\begin{center}
\begin{tabular}{p{5.5cm}p{3.5cm}p{4cm}}
\toprule
\textbf{Response} & \textbf{Gold answer} & \textbf{Result} \\
\midrule
\texttt{Novak Djokovic} $\hookleftarrow$ \texttt{Document (Title: \ldots)} \ldots
  & Novak Djokovic & 1 \\[2pt]
\texttt{Concepci\'{o}n, Chile} \ldots
  & Concepci\'{o}n & 1 \\[2pt]
\texttt{poetry and drama} \ldots
  & poet & 0 \\[2pt]
\texttt{The answer is Novak Djokovic}
  & Novak Djokovic & 0 (the answer is not at the start) \\
\bottomrule
\end{tabular}
\end{center}

{\bf Implementation Details: FirstLine, LastLine}

We extract the first (or last) line of the response that has real content (skipping empty lines and lines with only punctuation). Examples for {\em FirstLine}:
\begin{center}
\begin{tabular}{p{7cm}p{5cm}}
\toprule
\textbf{Response} & \textbf{Extracted answer} \\
\midrule
\texttt{poet} $\hookleftarrow \hookleftarrow$ \texttt{Document (Title: \ldots): \ldots}
  & poet \\[2pt]
\texttt{Tunisia, Republic of Tunisia, TUNISIA, \ldots} (all on one line)
  & Tunisia, Republic of Tunisia, TUNISIA, \ldots \\
\bottomrule
\end{tabular}
\end{center}

Examples for {\em LastLine}:
\begin{center}
\begin{tabular}{p{7cm}p{5cm}}
\toprule
\textbf{Response} & \textbf{Extracted answer} \\
\midrule
\texttt{Raya Yarbrough}
  & Raya Yarbrough \\[2pt]
\texttt{2004} $\hookleftarrow$ \texttt{Document (Title: \ldots): \ldots} $\hookleftarrow$ \texttt{Answer : 2013}
  & Answer : 2013 \\
\bottomrule
\end{tabular}
\end{center}

\subsubsection{Comparison to OOMB}
\label{sec:app-compare-oomb}

OOMB~\citep{Li:26} is a long context fine-tuning method which shares some properties with our technique (details are given in \secref{relwork} and \secref{app-relwork}. We ran a comparison, using their code from \url{https://github.com/wenhaoli-xmu/OOMB}. Results are given in \tabref{ruler-oomb-results} (right). Here, we provide details on the setup.
\begin{itemize}
\item
  OOMB subscribes to the paged memory model of vLLM~\citep{Kwon:23}, so that modern sparse attention policies like H2O cannot be implemented. We use their default sparse implementation, which is a simplied version of Quest~\citep{Tang:24}: the complete KV cache is held in paged memory, and for each chunk to be processed, a subset of pages is loaded into GPU memory for PagedAttention~\citep{Kwon:23}. This subset is determined by maximizing an instantaneous score depending on the query tensor for this chunk. Each page extends over all heads and spans a consecutive range of 64 tokens (64 being the page size which has to be choosen for Nvidia A100 GPUs).
\item
  As detailed in \secref{main-satt-vs-cp} and \secref{app-sparse-attn-libraries}, the paged memory model seems detrimental for sparse attention, since the unit of selection (a page) is larger than what we provide (and H2O requires to be functional) by a factor of page size times number of query groups (this is $8\cdot 64 = 512$ in our experiments). On the other hand, our method does not maintain the complete KV cache, but only a fixed-size part which can be kept in GPU memory. We compare to {\tt h2o\_norm} with chunk size 1024 on our side.
\item
  OOMB hyperparameters were set to the same values than what we used (optimizer, learning rate, number of epochs) and left at defaults otherwise. The page budget was set to $480$, so that the KV buffers in GPU memory for each MHA computation have the same size as on our side ($480\cdot 64 + 2048 = 32768$). The global batch size is 8 as in our case.
\item
  Comparisons were run on a number of Helmet datasets\footnote{
    Due to the long training time for OOMB, at the time of submission, we could only finish runs for 3 of the 8 Helmet datasets. A final revision will include results for all 8 datasets, as well as precise time measurements.}.
  We used the same hardware setup (Nvidia A100, 4 GPUs per training run), and the same way to select the winning checkpoint for test set evaluation. OOMB uses the 4 GPUs for tensor parallelism (while we use it for DDP), so that one update at batch size 8 requires four sequentially computed gradients to be averaged.
\end{itemize}

In our experiments (same hardware setup, same batch size), OOMB runs about three times longer than our method (using chunk size 1024).

As a final note, we detected several major issues in their codebase, which we fixed in order to guarantee a fair comparison:
\begin{itemize}
\item
  Tensor-parallel gradients are wrong: The input gradients of the column-parallel layers are never all-reduced over ranks. At TP=4 the gradient cosine to dense is about 1.0 at the last layer but 0.2-0.5 at layers 0-2.
\item
  LoRA crashes ({\tt LoraConfig} is never imported), and {\tt peft} cannot wrap OOMB's parallel layers.
\item
  {\tt head\_dim} is not passed to the KV cache (default 128), which gives out-of-bounds kernel reads for other head sizes.
\item
  Missing modules ({\tt corpus}, {\tt profiler}); no Qwen3 support (pinned to transformers 4.45, no q/k norm); no sparse inference ({\tt generate()} is not implemented).
\end{itemize}

In fact, their codebase seems written from scratch (likely with the use of generative AI) and does not depend on other higher level libraries. Our best guess is that this ``written from scratch'' property is inherent: the way in which they hide KV cache buffer nodes from the {\em autograd} graph requires subtly modified {\tt backward} code for all operators in the graph. This makes it very hard (if not impossible) to use external dependencies, such as LLM libraries (e.g., {\tt LitGPT}) or fast SDPA kernels. In contrast, our open source library uses external dependencies whenever we can.

\begin{figure}[ht!]
\begin{center}
  \includegraphics[width=\textwidth]{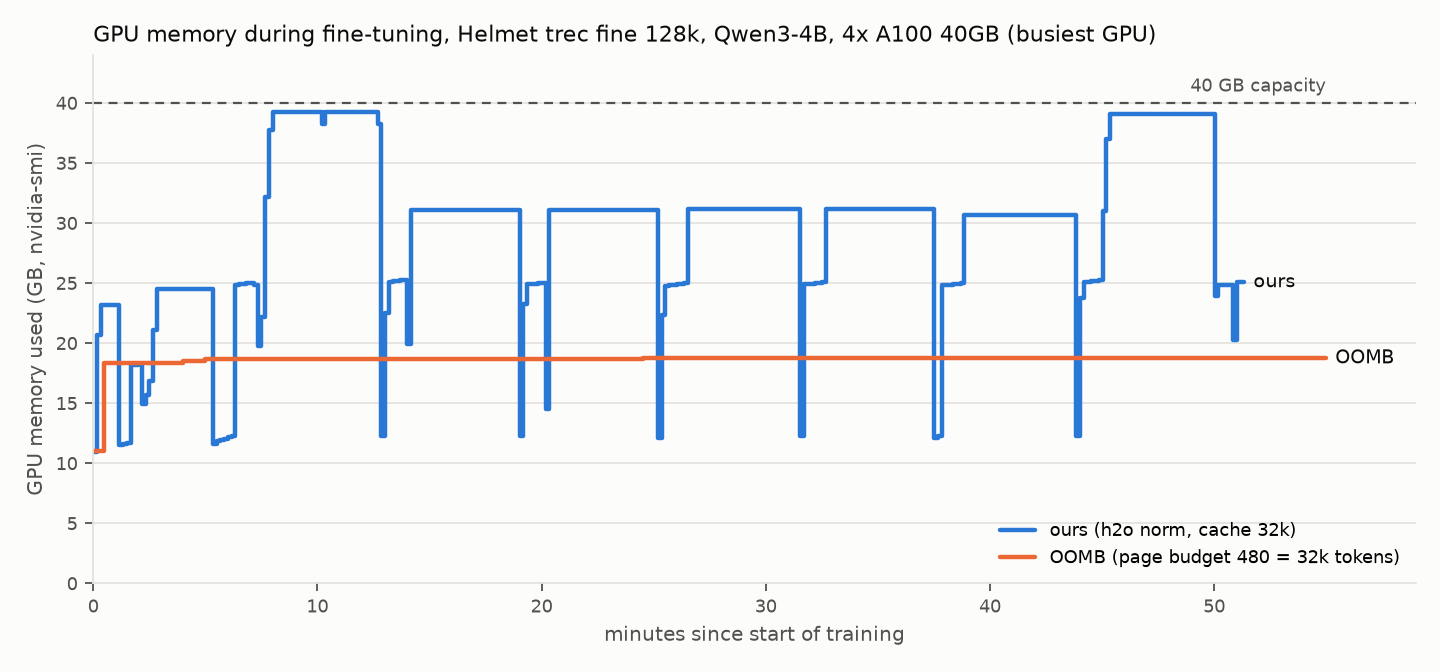}
\end{center}
\caption{\label{fig:gpu-memory}
  GPU memory usage for a training run on Helmet {\tt trec\_fine} (128k), comparing OOMB versus our novel method ({\em ours}).
}
\end{figure}

{\bf Comparison of Peak GPU Memory Usage}. As detailed in \secref{main-finetune} and elsewhere, the key limitation for long context inference or fine-tuning is limited GPU memory, and sparse attention is chiefly motivated by saving such memory. A successful method runs on limited GPU memory, but also makes use of all resources in order to decrease running time. In \figref{gpu-memory}, we compare peak GPU memory usage for our method versus OOMB on the same hardware setup and dataset.

Recall that we use four Nvidia A100 GPUs, each with 40 GB of RAM. Our method makes use of nearly all available GPU memory during the course of training. This is not surprising, since we chose relevant hyperparameters\footnote{
  As explained above, GPU memory is tightly controlled by these parameters. In particular, it does not depend on sequence length (which differs across the dataset).}
so to just not run out of memory (cache length 32768 slots, batch size 2, $\alpha = 0.75$ for chunk size 1024). The four GPUs are used for distributed data parallel, so this amount of memory is used on every GPU. In contrast, OOMB uses the four GPUs for tensor parallelism. Its peak GPU memory usage per device is about 18 GB. This is substantially below what could have been used (40 GB), despite the fact that we adjusted hyperparameters so that each MHA call depends on the same number of key-value parameters. Note that increasing the page budget in OOMB would make the method even slower, since more memory has to be transferred between CPU and GPU.

\subsection{Computing Summed Attention Weights in SDPA}
\label{sec:app-attn-weights}

As noted in \secref{main-h2o} and \secref{app-sdpa-kernels}, KV cache policies like H2O or related ones need summed attention weights $\sum_i m_{b, h, i, j}$ for each $(b, h, j)$. This array of shape $(B, H_q, N_k)$ can be obtained as byproduct of SDPA, which returns $\mxy{}$ of shape $(B, H_q, N_q, d_h)$. Note that summed attention weights are smaller than attention outputs, so there is a priori no reason for not returning them. However, all fast SDPA codes we know of, do not return this information. FlexAttention~\citep{Dong:25} can return log-sum-exp values $\log( \mxa{}\vone{}_{N_k} )$, where $\mxa{} = \mathtt{mask}( d_h^{-1/2} \mxq{}\mxk{}^T )$ is the argument of $\mathtt{softmax}$, likely because this is directly computed during FlashAttention~\citep{Dao:23}.

Our implementation contains {\em Triton} code for computing summed attention weights alongside a FlashInfer SDPA kernel~\citep{Ye:25}. This is an add-on, and it would be better if leading SDPA codes returned summed attention weights directly (see \secref{app-sdpa-kernels}). In this section, we detail how this can be done (even though we have not implemented this). We also show how to compute them with FlexAttention~\citep{Dong:25}, using two calls instead of one. This is contained in our implementation as baseline.

\subsubsection{FlashAttention for Summed Attention Weights}

FlashAttention works by essentially computing the attention weights tensor $\mxm{}$ in blocks, using a lattice tiling along the query and the key axes. It can be understood as {\em map-reduce}, where {\em map} is independent per cell. More precisely, the full attention weights have shape $(B, H_q, N_q, N_k)$. In the following, we drop $(B, H_q)$, treating them as "batch" dimensions. Use cell indices $(r, s)$ and index ranges $I(r)$, $J(s)$, so that the union of all $I(r)$ covers $\srng[0]{N_q - 1}$ and the union of all $J(s)$ covers $\srng[0]{N_k - 1}$. When computing the attention outputs, {\em reduce} operates along the key axis. Define an additional auxiliary tensor of shape $(B, H_q, N_q)$, with values
\[
  \vlam{r, s} := \log\left( \exp\left( \mxa{I(r), J(s)} \right) \vone_{|J(s)|} \right),
  \quad \mxa{I, J} := \mathtt{mask}\left( d_h^{-1/2}\mxq{I, \cdot} \mxk{J, \cdot}^T \right).
\]
Reduction works as:
\[
\begin{split}
  \vlam{r, s_1\oplus s_2} & = \max\left\{ \vlam{r, s_1}, \vlam{r, s_2} \right\}
  + \mathrm{log1p}\left( \exp\left( -\left| \vlam{r, s_1} - \vlam{r, s_2} \right| \right) \right), \\
  \mxy{r, s_1\otimes s_2} & = \left(\diag\exp\left( \vlam{r, s_1} - \vlam{r, s_1\otimes s_2}
    \right) \right) \mxy{r, s_1} + \left(\diag\exp\left( \vlam{r, s_2} -
    \vlam{r, s_1\otimes s_2} \right) \right) \mxy{r, s_2}.
\end{split}
\]
We can now run {\em map} independently for all $(r, s)$, then {\em reduce} along $s$ for all $r$.

Summed attention weights are given by $\vw{}\in \R^{N_k}$:
\[
  \vw{}^T = \vone_{N_q}^T\exp\left( \mxa{} - \vlam{}\vone_{N_k}^T \right).
\]
Define
\[
\begin{split}
  \tmxf{r,s} & = \exp\left( \mxa{I(r), J(s)} - \vlam{r, s}\vone_{|J(s)|}^T \right), \\
  \mxf{r, s} & = \left( \diag\exp\left( \vlam{r, s} - \vlam{r} \right)\right) \tmxf{r,s} = \exp\left( \mxa{I(r), J(s)} - \vlam{r}\vone_{|J(s)|}^T \right).
\end{split}
\]
If
\[
  \vw{r, s}^T = \vone_{|I(r)|}^T\exp\left( \mxa{I(r), J(s)} - \vlam{r}\vone_{|J(s)|}^T \right) = \vone_{|I(r)|}^T\mxf{r, s},
\]
then $\vw{} = [ \sum_r \vw{r, s} ]$. We can compute $\vw{}$ and $\mxy{}$ with an outer loop over $r$, inner loop over $s$. Initialize $\vw{} = \vzero_{N_k}$. The iteration $r$ works as follows:
\begin{itemize}
\item
  Map: Compute $[\vlam{r, s}], [\tmxf{r,s}], [\mxy{r, s}]$ in parallel.
\item
  Reduce: $(\vlam{r}, \mxy{r}) = \mathtt{reduce}\left( [\vlam{r, s}], [\mxy{r, s}] \right)$.
\item
  Compute $\vw{r, s}^T = \vone_{|I(r)|}^T \mxf{r, s} = \exp(\vlam{r,s} - \vlam{r})^T \tmxf{r,s}$. Add $\vw{r} = [\vw{r, s}]$ to $\vw{}$.
\end{itemize}
Compared to standard FlashAttention, we need to first reduce along $s$ in order to obtain $\vlam{r}$, keeping $\tmxf{r,s}$ and $\vlam{r,s}$ around.

\subsubsection{Summed Attention Weights with FlexAttention}

FlexAttention~\citep{Dong:25} stands out among fast SDPA codes by allowing the user to configure the computation in several ways (see also \url{https://pytorch.org/blog/flexattention/}). Here we describe how to compute summed attention weights $\vw{}$ alongside the attention output $\mxy{}$, by calling FlexAttention twice.

Recall that SDPA computes
\[
  \mxy{} = \exp\left( \mxa{} - \vlam{}\vone^T \right)\mxv{}.
\]
The summed attention weights are
\[
  \vw{} = \exp\left( \mxa{} - \vlam{}\vone^T \right){}^T \vone =
  \exp\left( \mxa{}^T - \vone\vlam{}^T\right)\vone =
  \exp\left( \mxa{}^T \right) \tvv{}, \quad \tvv{} := \exp(-\vlam{}).
\]
Up to softmax normalization, we can obtain this by calling a variant of SDPA again, flipping $\mxq{}$ and $\mxk{}$, reverting the attention masking, and passing $\exp(-\vlam{})$ as values. Importantly, FlexAttention returns $\vlam{}$ with the option {\tt return\_aux = AuxRequest(lse=True)}. Now, if $\tvlam{}$ denotes {\tt lse} for the second call (with $\mxq{}$ and $\mxk{}$ flipped), then:
\[
  \vw{} = \exp\left( \mxa{}^T - \tvlam{}\vone^T + \tvlam{}\vone^T \right) \tvv{}  = \left(\diag\exp(\tvlam{}) \right) \exp\left( \mxa{}^T - \tvlam{}\vone^T \right) \tvv{}  = \exp(\tvlam{}) \circ \tvy{}.
\]
Finally, trying to minimize numerical errors (we are using 16 bit data types), we use $\exp(-(\vlam{} - \bar{\lambda}\vone))$ and $\exp(\tvlam{} - \bar{\lambda}\vone)$, where $\bar{\lambda} = N_q^{-1}\vone^T\vlam{}$ is the mean of $\vlam{}$. All in all:
\begin{itemize}
\item
  $(\mxy{}, \vlam{}) = \mathtt{SDPA}(\mxq{}, \mxk{}, \mxv{})$, $\bar{\lambda} = N_q^{-1}\vone^T\vlam{}$.
\item
  $(\tvy{}, \tvlam{}) = \mathtt{SDPA\_rev}(\mxk{}, \mxq{}, \exp(-(\vlam{} - \bar{\lambda}\vone)))$, then $\vw{} = \exp(\tvlam{} - \bar{\lambda}\vone)\circ \tvy{}$.
\end{itemize}
Here, $\mathtt{SDPA\_rev}$ differs from $\mathtt{SDPA}$ by the attention masking being reversed. FlexAttention allows to specify the attention mask as $\mathtt{block\_mask(b, h, q\_idx, kv\_idx)}$. The mask for $\mathtt{SDPA\_rev}$ is given by flipping $\mathtt{q\_idx}$ and $\mathtt{kv\_idx}$ in the code for $\mathtt{SDPA}$. Note that FlexAttention supports $\mxv{}$ to have a different (final) embedding dimension that $\mxq{}, \mxk{}$. Maybe this even translates in the second call being faster than the first. All in all, compared to a single FlexAttention call and no attention weights, this is at most twice as expensive.

\subsection{Sparse Attention and SotA Inference Libraries}
\label{sec:app-sparse-attn-libraries}

In \secref{main-satt-vs-cp}, we discuss the (somewhat surprising) fact that as of today, sparse attention is not much used in real-world practice, because existing implementations are too slow to be competitive with the state of the art. While a part of the latency gap between sparse attention and sequence or context parallelism is probably inherent, we argued in \secref{app-sdpa-kernels} that some shortcomings of current sparse attention implementations are easy to eliminate by minor extensions of fast SDPA kernel codes.

Here, we comment on why sparse attention policies, such as H2O, are not supported in vLLM~\citep{Kwon:23}, the leading fast inference library. Details are found in \url{https://github.com/vllm-project/vllm/issues/10646}, \url{https://github.com/vllm-project/vllm/issues/12254}, \url{https://github.com/vllm-project/vllm/issues/5751}. In vLLM, KV caches are maintained as set of fixed-sized pages (or blocks). The main issue is that they require a page to store KV content {\em across all heads and all batch positions}: KV information for a token is stored in the cache for all heads and batch positions, or for none. The cited RFCs mention that it would require significant changes to the memory layout and block manager abstractions to change that.

However, for modern sparse attention (such as H2O), policies $\pi(b, h, t)$ depend on $(b, h, t)$ in general: they select different tokens per head or batch dimension. We want heads to specialize to different ways of encoding the past, so why restrict them to attend to the same snapshot of the past? In general, different sequences in a batch have little to do with each other. Why would the same token positions be most relevant for them?

As detailed in \secref{main-kv-cache}, our implementation does not require paging at all. We simply maintain {\em dense} buffers of shape $(B, H_k, N_C, d_h)$, where the cache length $N_C$ is fixed independent of context width, and then use $\mathtt{torch.gather}$ and $\mathtt{torch.scatter}$ for read and write access. Whereas Paged\-Attention requires specific SDPA kernels, we can use existing dense SDPA codes, as long as we cater for causal masking (see \secref{app-sdpa-kernels}). While not supported in our current implementation, we could build up KV cache buffers in chunks to cater for sequence lengths shorter than $N_C$, thereby solving the issue of unnecessary pre-allocations~\citep{Kwon:23} without heavy machinery. Finally, while $\mathtt{torch.gather}$ and $\mathtt{torch.scatter}$ access the buffer in a non-contiguous way, this is very subdominant to SDPA computations in our experience. In fact, when calling $\mathtt{scatter}(\mathtt{keys}, \mathtt{index}, \mathtt{key\_new})$, the final axis of $\mathtt{index}$ is always constant (so that the final buffer axis of size $d_h$ is accessed contiguously), and optimized $\mathtt{scatter}$ and $\mathtt{gather}$ kernels could easily be implemented for this case if the PyTorch implementations do not already cater for this special case.

As long as highly optimized and widely used inference libraries do not support sparse attention, it may remain underused. We hope that our work and open source library sparks some renewed interest in this direction.

\subsection{Details on Related Work}
\label{sec:app-relwork}

Here, we provide additional details about relations of our method with prior work. OOMB~\citep{Li:26} shares properties with our work, such as chunk-level processing, activation checkpointing, and efforts to compress KV cache buffers for {\em autograd}. Details on the relationship are as follows:
\begin{itemize}
\item
  Their implementation is better suited for representing KV caches exactly (no selection or compression) that ours. They implement a paged memory management like~\citep{Kwon:23}, which we do not (but see \secref{app-sparse-attn-libraries} and comments in \secref{main-kv-cache}). However, despite all efforts in CPU offloading and activation checkpointing, they run into the same barrier as \citep{Li:25d}, in that the factor for the final chunk depends on {\em all} KV cache buffers of all layers, so cannot be represented by {\em autograd} on a single device. At this point, RingAttention~\citep{Liu:24} is the method of choice, and it is not clear why their library would improve on implementations such as MS-SWIFT~\citep{Zhao:25} or Nvidia NeMo RL (\url{https://github.com/NVIDIA-NeMo/RL}).
\item
  They deal with activation memory for GPU by activation recomputation, while we use activation checkpointing. In the former, activations are recomputed during the backward pass {\em from the start}, whereas in the latter, recomputation starts from the most recent checkpoint. The former is too slow to be useful for deep models, so our guess is their code actually uses activation checkpointing.
\item
  By subscribing to a paged memory model for the KV cache, they inherit the restrictions of vLLM detailed in \secref{app-sparse-attn-libraries}. Any cache policy $\pi(b, h, t)$ which depends on $h$ or $b$ cannot be implemented. In particular, none of the H2O variants can be made to work with OOMB. In fact, their code requires pages not just to span all batch dimensions and heads, but also a {\em consecutive range} of token positions, coarsening the granularity of selection even further.
\item
  There are also potential advantages of using the paged memory model. First, OOMB holds the complete KV cache in paged memory (stored on CPU). For each new chunk to be processed, a subset of KV pages is loaded into GPU memory, as selected by an instantaneous score. While we can select the active KV cache in a much more fine-grained manner, once KV information is evicted, it is gone and cannot be used later on. Second, if a new chunk is larger than the page size, their PagedAttention implementation is more flexible than what we do. Namely, a chunk is split into tiles of page size length, and the query vectors in each tile can attend to different KV cache pages. In our method, slots are evicted once per chunk, so this flexibility is not present.
\item
  The most important difference is how they deal with KV cache buffers as nodes in the {\em autograd} graphs, and what this implies for generality. This is also the biggest challenge we face, and we deal with it by a combination of nested checkpointing, delta encoding of KV cache buffers, and integration into {\em PyTorch} by way of autograd saved tensor hooks (\secref{app-exploit-recurrence}). Together with recording and replaying KV cache decisions, this {\em renders our implementation fully agnostic to the KV cache policy}: it works with any selection or compression policy (see \secref{relwork} for many references). In contrast, they try to hide all nodes representing KV cache content from {\em autograd} altogether, so that none of this information can be placed in the computation graph. This is possible only by implementing a number of complex CUDA kernels, in which all inner derivatives w.r.t.\ these ``KV cache nodes'' are made explicit. Apart from substantial derivation and implementation complexity, their approach {\em must be specialized to the KV cache policy being used}. In fact, their paper only provides results for two specific sparse attention policies (LSA and DSA). They do not explain how the hiding of KV cache buffers from {\em autograd} works in general. Since KV cache updates are tightly coupled with SDPA calls, this seems hard to achieve, and what they do may be restricted to the few paged memory restricted policies they implement.
\item
  Their implementation contains no comments on how the backward operators have been obtained, and there is no explanation what to do for other KV cache policies. In contrast, our implementation works for any KV cache policy out of the box: only its forward implementation needs to be provided. Exploring fine-tuning for any policy of choice requires almost no extra work and can be run on the same resources as inference. Our implementation is also agnostic to SDPA kernels, so any new innovation in this direction can be incorporated without having to rewrite specific operators (as is required for OOMB).
\item
  Both their and our implementation make use of CPU offloading of activations, KV cache buffers, and head gradients. They claim to have done this asynchronously, as in \citep{Yuan:26}, which hides latency. We have also experimented with this, but did no so far achieve significant speedups. Moreover, asynchronous transfer requires double buffering, which drives up GPU memory requirements. Still, more effort in this direction is warranted.
\end{itemize}

\ifthenelse{\equal{\isarxiv}{no}}{}{
\subsection{Open Source Library KeysAndValues. Experiments}
\label{sec:app-os-library}  

For all experiments above, fine-tuning with our method (column {\em us} in \tabref{helmet-results}) and inference with sparse attention (all policies) were done with a new open source library for long context fine-tuning and inference: $\mathtt{KeysAndValues}$ (\url{https://github.com/awslabs/keys_values}).

Apart from efficient code for our fine-tuning method, the library provides clean and simple abstractions for sparse attention and key-value caches of limited size. Among its features are:
\begin{itemize}
\item
  Long context fine-tuning on a single GPU (this work).
\item
  Several variants of the H2O KV cache policy~\citep{Zhang:23}. The library provides a generic implementation for any cache logic of the form $\pi(b, h, t)$ which makes use of summed attention weights.
\item
  Quantization of KV cache buffers.
\item
  Integration of FlexAttention~\citep{Dong:25}, FlashInfer~\citep{Ye:25}, FlashAttention~\citep{Dao:23,Shah:24} and eager SDPA behind a common multi-head self-attention interface. This includes summed attention weights (for H2O-like policies), as well as a proper $\mathtt{backward}$ implementation.
\item
  Model implementations and inference code is from LitGPT (\url{https://github.com/lightning-ai/litgpt}), which allows for almost any Hugging Face checkpoint to be used. However, while bringing modern KV caching to Hugging Face would require hacking several code files {\em separately for every single model}, you can apply your KV cache policy or attention approximation to almost all models with few changes of common code.
\item
  Support of CPU offloading of KV cache buffers and model weights.
\item
  Support of distributed training (distributed data parallel, CPU offloading of model weights optional). Support of distributed evaluation.
\end{itemize}

With this library, we do not intend to compete with vLLM~\citep{Kwon:23} or SGLang~\citep{Zheng:24}, which include more low level optimizations and support of latest GPU architectures. Instead, we make it easy for researchers to explore new KV cache policies, post-time training ideas, or unusual multi-head self-attention approximations, providing clean abstractions of these concepts which can be used and extended without having to deal with intricate implementation details of existing high-performance libraries.

\subsubsection{Running Our Experiments}

Once $\mathtt{KeysAndValues}$ has been properly installed, the training runs for our method can be reproduced as follows. You need to be on an instance with at least four Nvidia A100 GPUs with 40 GB of RAM. We used {\tt AWS EC2 p4d.24xlarge} instances, which have 8 A100 GPUs, running two experiments in parallel on each instance.

{\tiny
\begin{lstlisting}[language=bash]
export DATASET_KEY="nq"; \
export DATASET_SIZE="128k"; \
export POLICY_NAME="h2o-orig"; \
export CACHE_LENGTH="32768"; \
export CHUNK_SIZE="2048"; \
export EVAL_STEPS=10; \
CUDA_VISIBLE_DEVICES="0,1,2,3" \
PYTORCH_ALLOC_CONF=expandable_segments:True \
KEYSVALS_LOG_DIR="./finetune/helmet_${DATASET_KEY}_${DATASET_SIZE}/${POLICY_NAME}_cs${CHUNK_SIZE}/logs" \
python3 keys_values/__main__.py finetune_long_lora \
    Qwen/Qwen3-4B-Instruct-2507 \
    --out_dir ./finetune/helmet_${DATASET_KEY}_${DATASET_SIZE}/${POLICY_NAME}_cs${CHUNK_SIZE} \
    --precision bf16-true \
    --verbose some \
    --devices 4 \
    --data Helmet \
        --data.dataset_key ${DATASET_KEY} \
        --data.max_length ${DATASET_SIZE} \
        --data.metadata_dir ./data \
        --data.trainloader_longest_first True \
    --train.save_interval ${EVAL_STEPS} \
        --train.micro_batch_size 2 \
        --train.epochs 5 \
        --train.average_loss_per_batch True \
    --eval.interval ${EVAL_STEPS} \
        --eval.initial_validation True \
        --eval.use_sample_metric False \
    --kv_cache.cache_length ${CACHE_LENGTH} \
        --kv_cache.chunk_size ${CHUNK_SIZE} \
        --kv_cache.name ${POLICY_NAME}-torch-quantized8 \
    --grad.layers_per_cell 1 \
        --grad.layercp_qname default \
        --grad.cachecp_qname torch-quantized8 \
        --grad.chunks_per_cell_multiplier 1 \
    --optimizer.name AdamW \
        --optimizer.learning_rate 0.0005 
\end{lstlisting}
}

Once all desired training runs have finished, evaluations (on the test sets) can be run as follows.

{\tiny
\begin{lstlisting}[language=bash]
CUDA_VISIBLE_DEVICES="0,1,2,3" \
PYTORCH_ALLOC_CONF=expandable_segments:True \
KEYSVALS_LOG_DIR="./finetune/evaluation/myruns/logs" \
python3 keys_values/__main__.py eval_long_ext \
    ./myruns.yaml \
    --verbose some \
    --devices 4 \
    --batch_size 2 \
    --use_sample_metric True \
    --sample_metric_max_generated_tokens 20 \
    --num_store_generated_samples 1000
\end{lstlisting}
}

Here, {\tt myruns.yaml} is a YAML file containing entries of this form:

{\tiny
\begin{lstlisting}[language=bash]
- out_dir: ./finetune/helmet_nq_64k/h2o_cs2048
  model_type: lora
  eval_tasks:
    - step-000420
\end{lstlisting}
}

For each setup, evaluations can be run for different checkpoints {\tt step-000***} stored alongside training. In our experiments, for each setup, we select the checkpoint which minimizes validation loss. We refer to {\tt README.md} for further details on how to aggregate evaluation results and create result tables.

In case you use several different instances, it is important to note that {\tt --data Helmet} is a {\em dataset generator} rather than a fixed dataset: it creates potentially different datasets each time it is called. However, datasets are stored locally and loaded if present. If training runs and evaluations are split across different instances, it is important to make sure that they {\em share the same dataset instances}. This can be ensured as follows:

\begin{itemize}
\item
  When run for the first time on an instance with some {\tt --data.dataset\_key}, the generator creates a dataset instance. Moreover, all sequences are tokenized and their lengths are stored in a metadata file. Also, the split into training and validation sets is done here and stored in the metadata file.
\item
  The dataset instances are stored locally under {\tt $\sim$/.cache/huggingface/helmet/longtrain}. The metadata file is under {\tt --data.metadata\_dir}.
\item
  Sharing a dataset and splits across instances means that these directories must have the same content. If you can mount a common volume between the instances, this is the best choice. Otherwise, make sure to copy files across.
\end{itemize}

}